\documentclass[acmtog, screen, nonacm]{acmart}
\makeatletter
\let\@authorsaddresses\@empty
\makeatother
\AtBeginDocument{%
  }

\begin{document}

\title{Detangled: A Framework for Creating, Editing, and Inferencing Feature Rich Hair Strands}


\author{Sarah Jobalia}
\email{sjobalia@cs.stanford.edu}
\orcid{0009-0001-2062-6161}
\affiliation{
  \institution{Stanford University}
  \city{Stanford}
  \state{California}
  \country{USA}
}

\author{Yitong Deng}
\email{yitongd@stanford.edu}
\orcid{0000-0001-9157-1519}
\affiliation{
  \institution{Stanford University}
  \city{Stanford}
  \state{California}
  \country{USA}
}

\author{Carolyn Smith}
\email{carsmith@cs.stanford.edu}
\orcid{0009-0008-6547-2232}
\affiliation{
  \institution{Stanford University}
  \city{Stanford}
  \state{California}
  \country{USA}
}

\author{Ronald Fedkiw}
\email{fedkiw@cs.stanford.edu}
\orcid{0009-0008-6853-2499}
\affiliation{
  \institution{Stanford University}
  \city{Stanford}
  \state{California}
  \country{USA}
}




\begin{abstract}
  We present a framework for understanding and generating feature rich hair strands. Drawing upon both scientific and cultural expertise, we define strand texture as the various distinctive patterns (curling, switchbacks, twist, etc.) that are formed by forces internal to a hair strand. We begin by proposing a novel five-dimensional parameter space, intended to be a bijection with naturally occurring hair strand textures. This encoding is both qualitatively accessible, allowing users to readily locate their own hair in the parameter space, and quantitatively precise, allowing the generation of individual strands from texture inputs. Importantly, strand texture should be independent from the overall strand direction. In order to disentangle strand texture from the overall strand direction, we identify centerline geometry and use it to map strands into a canonical space (a strand texture space). We construct centerlines using a novel method that cleanly distills complex hair grooms, separating the strand texture from the overall style (parameterized by style guides). We enable the creation of new strands conforming to our parametric description of texture via a generative artificial intelligence approach supervised by a separate neural network trained to label candidate strands according to our five-parameter description. The ability to create new strands conforming to any desired texture enables groom editing using either texture transfer or user-provided inputs. We demonstrate results on a variety of hair types.
\end{abstract}

\begin{teaserfigure}\label{fig:teaser}
  \includegraphics[width=\textwidth]{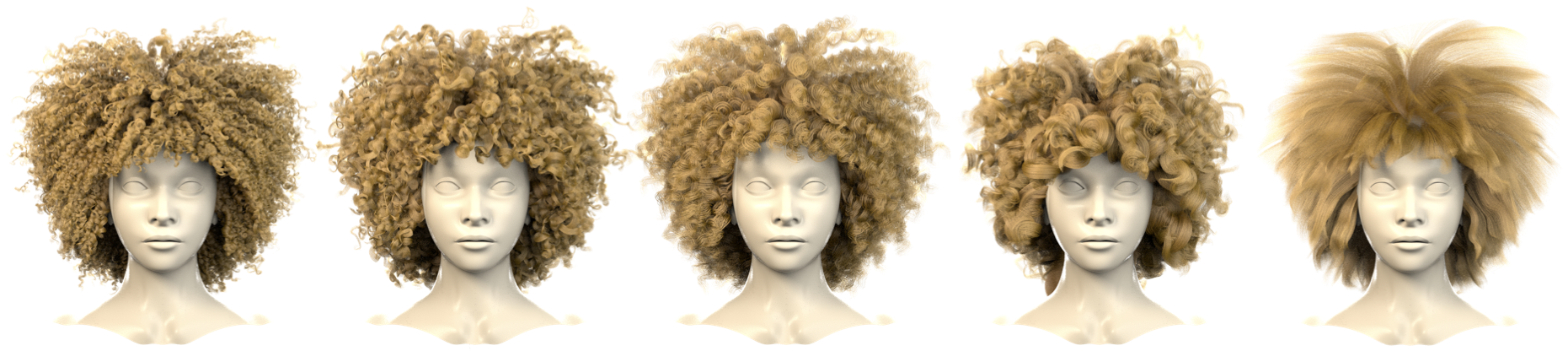}
  \caption{Using our pipeline, we transform an input groom (center) to reflect various target hair strand textures while maintaining the overall style. The four grooms surrounding the input groom are rendered outputs from our pipeline.}
  \Description{Center artist-created groom surrounded by four results from this pipeline.}
\end{teaserfigure}


\maketitle

\section{Introduction}\label{sec:introduction} 

Hair is central to how humans express themselves~\cite{culturalHistoryOfHair}. It not only reflects heritage (people, places, and communities) but also signifies the cultural landscapes people navigate and the personal choices they make within them. Dating back to as early as the Paleolithic era, descriptions and depictions of hair have been an essential part of storytelling~\cite{VenusDeWillendorf, venusOfBrassempouy}. In the modern digital world of film, television, social media, and gaming, hair representations are increasingly important. In fact, with U.S. consumers spending an average of eight hours per day engaging in digital media~\cite{mediaUseStats}, the characters they interact with have an outsized impact on how they perceive themselves and the world around them~\cite{blackHairPerception}.  Particularly for children, being exposed to depictions of hair that resemble their own builds a sense of relevance, validity, and empowerment~\cite{ImpactOfFilms}. 

 Certain hair types are much more rare on screen than in the population at large~\cite{kimRacistLegacy}. In particular, curly and/or afro-textured hair is drastically underrepresented compared to its almost 40\% prevalence worldwide (65\% in the U.S.)~\cite{IfBigNatural2015, bigHairVogue, blackWomenInFilm,TexturedTrends, BeyondtheCurl}. In order for digitally created content to better reflect the population, more robust hair generation and grooming tools are necessary. Current toolsets typically start from a straight hair default and lack the flexibility required for a non-expert to create and edit more diverse hairstyles. Even with recent advancements in data-driven approaches, curly and afro-textured hairstyles still remain challenging. Many state-of-the-art approaches either cite curls and/or complex hairstyles as failure cases or fail to demonstrate examples that would prove otherwise (see e.g.~\cite{hairGAN, hairStep, DynamicReconstruction, neuralHDHair, neuralHaircut, HAAR, Zhou2023}).

 In order to devise a more robust approach, we begin by re-evaluating what metrics best describe a head of hair. Many prior works aim to reconstruct hairstyles based on image and video data, emphasizing millimeter accuracy, single-strand accuracy, and other similar measures. However, these metrics of evaluation are outdated almost as soon as the photo is taken. \textit{Hairstyles}, as defined by specific strand placement, can be changed by factors as innocuous as a turn of the head or as common as a light breeze. Avoiding this ill-defined word altogether, we instead introduce the term \textbf{strand texture} to describe the intrinsic and semi-persistent properties of hair strands and the term \textbf{style guides} (which can be interpreted as centerlines or likened to guide hairs) to describe their dynamically changing extrinsic properties. Whereas style guides readily respond to external mechanical forces, strand texture maintains its integrity until chemical structure is altered (i.e.\ when perms, water interaction, sun exposure, and other chemical manipulations change protein bonds~\cite{follicleEnigma, macrofibrils}, lipid formations~\cite{strandLipids}, or other internal structures). In Section~\ref{sec:strand_texture}, we define strand texture via five fundamental parameters, offering the first attempt at providing comprehensive and quantifiable metrics for describing hair texture in computer graphics.

 In order to better orthogonalize the notions of strand texture and style guides, especially for data-driven approaches, we create a canonical space that removes as much of the style guide contribution as possible. Canonical spaces have recently proven to be quite useful for both NeRFs~\cite{NeRF} and Gaussian Splats~\cite{gaussiansplats}, where facial pose and expression parameters are used to transform information out of a neutral pose/expression in the canonical space (see e.g.~\cite{FlashAvatar, psavatar, GaussianHead, HifiGaussianHeadAvatar, hq3DAvatar}). In our canonical space, or texture space, the style guides are perfectly straight (albeit with varying lengths) and deviations from them are dictated by the strand texture. The one-to-one mapping from a canonical space style guide to a world space style guide informs the placement of textured hair strands on the head. 

Our canonical texture space fosters a myriad of opportunities for data-driven approaches to generate hair of all types. We demonstrate its utility by presenting a pipeline that can easily and accessibly alter an existing hair groom to contain a new hair type, which can be specified either by a set of texture parameters (radius, wavelength, and twist prevalence) or by a few reference hairs. This is accomplished using a diffusion model that outputs a differentiable strand representation in the canonical space. The generated candidate strands are supervised by both the input texture parameters and the lengths of the canonical space style guides, noting that length supervision facilitates the straightforward mapping of newly generated hairs from canonical space to the world space groom. To enable parameter supervision, we train a neural network to dynamically label texture parameters on candidate strands. This allows the diffusion model to continuously assess the texture of generated strands, aiming to adjust them to match the target texture (and length).

\subsection{Contributions}\label{sec:contributions}
\begin{itemize}
    \item  Based on our collaborations with hair stylists, scientists, and artists, we developed a universally applicable and novel five-dimensional parameter space to describe hair strand texture. 
    \item We introduce a novel method for calculating the hair strand centerlines in a manner that disentangles strand direction from our newly proposed strand texture. This facilitates orthogonalization of geometry and texture.
    \item Drawing on our strand texture parameterization and our canonical space representation, we present a generative model that can create diverse and realistic hair strands for any desired input texture.
    \item  To enable parameter supervision, we train a neural network to dynamically label texture parameters on candidate strands 
    \item We present compelling results for both hair texture transfer and generation.
\end{itemize}

\section{Related Work}\label{sec:related_work}

\textbf{Background:} Hair has been the subject of robust cultural~\cite{culturalHistoryOfHair}, artistic~\cite{VenusDeWillendorf}, and scientific~\cite{ Thibaut2007, mixedRaceStrands, Loussouarn2007} exploration. Due to the diverse range of hair types and styles present in commerce and the media, various taxonomies~\cite{matrix, typeFromImage, cloeteSystemsApproachHuman2019, Natural, Hairmony, B1988, HairCode, hairProperties} have emerged. Although earlier works focused on racial categorizations (see e.g.~\cite{Mettrie2007}), the diversity of hair types within similar ethno-geographic populations led to the creation of more nuanced classification. Andre Walker~\cite{Natural} popularized the notion of descriptive `buckets' that users could use to classify their hair type (see also~\cite{typeFromImage, matrix}). Within the scientific community, more technical classifications abound, including specific measurements for lipid content~\cite{lipidcontent, strandLipids}, breakage frequency~\cite{porosityEffect, Gray2000, hairDifferences}, follicle positioning~\cite{follicleEnigma, B1988}, etc. We use this scientific and cultural basis to inform our framework, distilling classifications that are significant from a computer graphics perspective in order to devise a taxonomy that is both interpretable and computationally tractable.

\textbf{2D Image Generation and Editing: } Although our work focuses on fully 3D modeling, it is worth briefly addressing the large body of work on portrait generation and editing. StyleGANs~\cite{stylegan} (derived from GANs~\cite{gan}, see also~\cite{Maluleke2022}) have been used to transfer hair from one image to another, e.g.~\cite{styleGANSalon, StyleYourHair, HairNet, xuPersonalizedHairstyleHair2024, HSSAN, ReStyle}. CLIP~\cite{clip} enables text-based image editing, see e.g.~\cite{HairCLIPv2, StyleCLIP}. Diffusion methods  (\hspace{1sp}\cite{diffusion}, see also~\cite{Karras2022, Ho2020}) are quite popular and have been used for image editing and generation, see e.g.~\cite{Palette}. A number of so-called 3D-aware methods (see e.g.~\cite{GeoWizard, VoxGRAF, pv3d}) have been utilized in order to facilitate image editing, see e.g.~\cite{textTo3DAware}. There are also a number of other exciting research directions aimed at identifying, generating, and modifying hair in images, see e.g.~\cite{Sapiens,blowingHair}. 

\textbf{Implicit 3D Geometry:} Gaussian Splatting~\cite{gaussiansplats} and NeRFs~\cite{NeRF} disentangle the camera view from the 3D model, even though they do not typically aim to explicitly reconstruct geometry or disentangle texture, shading, lighting, etc.. Their popularity stems from their ability to generate realistic images from novel views using either image or video data. Many authors have used these representations to facilitate full head/body and head-only reconstructions. Some of these methods produce notably good hair visualizations, see e.g.~\cite{hq3DAvatar, DeformableNerfs, FlashAvatar, HifiGaussianHeadAvatar, GaussianAvatar, GaussianHead, HeadGaS, HyperNeRF, insta, monoGaussianAvatar, PointAvatar, psavatar, splattingAvatar, Tri2plane}. In particular, see~\cite{HairNeRF,singleViewImplicit, 3DGH} for examples of hair style transfer from images to 3D representations. A number of authors leverage triangulated (or quadrilateral) surfaces to enhance the efficacy of their implicit representations, see e.g.~\cite{AG3D, LUCAS, CodecAvatarStudio, HumanRef, IntrinsicAvatar, MVHumanNet, Portrait4D, GaussianAvatars, PanoHead, GaussianHeads}.  Taking this approach one step further,~\cite{HVH, gaussianHair} build implicit representations on top of strand-based geometry. Notably, an implicit representation can enhance appearance, even when one already has accurate strand-based geometry. Unfortunately, these strand-based implicit representations currently struggle to capture the lighting and geometric details of curly and afro-textured hair in part because they rely on direction-based image gradients that assume low-curvature geometry.

\textbf{Strand-Based Representations: } Many works aim to reconstruct a full head of hair strands from input images and video. Some approach this this by retrieving sections of hair~\cite{Struct2Hair,databaseCombinations} or full grooms~\cite{Hairmony} from large datasets that best match the target image. Unfortunately, databases chronically underrepresent curly hairstyles. This leads to domain gaps in training data that make database selection methods ineffective for curly and afro-textured hair. In order to generate strands beyond those included in databases, 2D orientations can be inferred from the target image and used to predict a 3D orientation field used to grow strands~\cite{DynamicReconstruction, hairStep, neuralHDHair, hairGAN,GroomCap, MonoHair}. This can be aided by the use of high-resolution image capture~\cite{strandAccurateMVS,CT2Hair}. Unfortunately, relying on 2D orientations to capture hair geometry from an image is inadequate when curls with different directions intersect in a single pixel or strand gradients become too steep. This makes it difficult to reconstruct curly and afro-textured hair. Alternatively, a 2D latent texture can be used to directly predict strand geometry at various scalp locations~\cite{neuralHaircut,DiffLocks, NeuralStrands}. Although strand geometry can often be refined via differentiable rendering~\cite{gaussianHair,Zakharov2024, Dr.Hair, GAF}, this is challenging for curly and afro-textured hair, which often self-occludes. Additionally, existing rendering methods struggle to capture hair with high internal twisting. Text-driven hair generation has also recently become popular~\cite{HAAR,SimAvatar, Perm} (see also~\cite{Zhou2023}); however, these methods struggle with curly and afro-textured hair.

\textbf{Artist-Inspired Modeling and Editing: } When the priority is artistic control, 3D modeling packages such as Houdini~\cite{houdini}, Maya~\cite{maya}, and Blender~\cite{blender} are popular. Various works have aimed to improve the artist workflow by leveraging sculpted geometry~\cite{curvatureAnalysis}, sketches~\cite{zhangEnergyHairSketchBasedInteractive2022}, retargeting techniques~\cite{shapeAdaptation}, etc. Traditionally, sparse guide hairs have been used for modeling and editing under the assumption that they dictate the positions of the rendered hairs via some form of interpolation. More recently, the notion of densely-placed guide hairs (one centerline for each rendered wisp) has been proposed for curly hair~\cite{Perm, DCT}.  This forms the basis for our approach. Although we use~\cite{DCT} as an initial guess for our centerlines for the sake of expediency, we subsequently improve the centerlines to remove obvious deficiencies.~\cite{Perm} addresses the concept of disentanglement directly, identifying centerlines from PCA low frequencies; however, PCA makes the method data-dependent and non-intuitive for non-experts. In contrast, we take an artist-accessible approach, identifying texture features in a more interpretable fashion.~\cite{genGuides} has similar goals; however, they index too heavily on artist workflows, which lack scientifically-based tools for accurately representing highly-textured hair types.

\textbf{Simulation: }  Simulation techniques vary based on the model and parameterization (e.g.~\cite{Frenet, Bishop1975}) used to describe the hair. A variety of approaches (e.g.\ mass-spring systems~\cite{Selle2008}, Kirchoff Rods~\cite{Kirchhoff1876,Bergou2008}, etc.) have been used with varying success~\cite{Anjyo1992, Curly-Cue, MERCI, bertailsSuperhelicesPredictingDynamics2006}. Various works have aimed to improve stability~\cite{shiLiftedCurlsModel2023, ammosovGeneralizedMultiscaleFinite2024}, speed~\cite{Daviet2023, Hsu2024, Hsu2025,TowardsRealtime}, accuracy~\cite{ANIME-Rod, Lespagnol2024, hairwater, Tencers}, etc. See also~\cite{Hsu2023, Takahashi2025}. More recently, learning-based methods have gained popularity, see e.g.~\cite{multiResPhysics,Phase2vec, Romanya-Serrasolsas2025, Quaffure, augMassSpring, realtimeNeuralSimulation}.

\section{Strand Texture}\label{sec:strand_texture}
We define strand texture as the persistent strand shape that results from an amalgamation of internal biochemical processes. Quantitatively representing strand texture by a set of physical parameters enables  creation of a diverse set of realistic hair strands that can be used as data in our neural pipeline. Unlike many modeling tools that approximate highly textured hair with layers of waves and noise (or frizz)~\cite{houdini, blender}, we prioritize scientific underpinnnings in order to create realistic strands representative of the entire hair spectrum. 

Even if it were possible to visually evaluate the millions of protein bonds in the strand cortex, that level of detail would make modeling and simulation intractable; thus, we instead characterize the emergent properties of the internal structure of hair. Our definition is informed by social, scientific, and computational considerations: Socially, both the artists creating the grooms and the users interacting with them are familiar with the culturally-created terms society uses to discuss hair; therefore, our definition should be socially significant and easily understandable in order to facilitate both accessibility and editability. Scientifically, our definition should ensure that newly considered strands reflect the diverse range and physical limits of real-world hair strands. Computationally, our definition should describe the visible properties of the strand while being detailed enough to provide a constitutive model for numerical simulation.

\subsection{Parameterization}\label{sec:defining_texture_parameters} In order to ensure adherence to the aforementioned considerations, we iterated our parameter set through many conversations with hair stylists,  hair scientists, and other experts in the biosciences community. The set of parameters we use to define strand texture is as follows: \textit{thickness}, \textit{curl radius}, \textit{curl wavelength}, \textit{twist prevalence}, and \textit{porosity}. Although some of these parameters vary with location on the scalp and/or arc length along the strand, we use a single characteristic value for each parameter in order to mimic how hair is typically described by prioritizing the more visually prominent and often more highly-textured lower sections of strands. 

\subsubsection{Thickness} The cross-sectional size and shape of hair strands varies significantly across the Mammalia class, and it determines both form and function~\cite{Brunner1974}. On the human head, the cross-sectional shape of hair strands varies from circular to elliptical, and higher anisotropy increases the likelihood of bending. Although strand shape can be determined with the aid of a cheap microscope (100x or so magnification, costing around \$100), it is not something that indivuduals or stylists tend to refer to directly. Instead, anisotropy is better measured indirectly via observations of its emergent properties, such as curl radius and wavelength. The ratio of the minor to major axes ranges from $.55$ to $1$~\cite{fiberShape} and is roughly correlated with ethno-geographic variation~\cite{crossSectionRace}. In contrast to shape, the overall cross-sectional size is observable via the naked eye and is typically described via `thickness'. Strand diameter ranges from $20$ to $180 \mu m$~\cite{hairStructure}, and it influences the amount of space a hairstyle takes up as well as the weight of each strand. Hair thickness plays a central role in both hair maintenance and perception, tangibly influencing how one might describe a given head of hair. 

\subsubsection{Curl Radius and Wavelength} 
Curl radius and wavelength derive primarily from the internal forces that the protein bonds within the center of a hair strand apply to the strand geometry~\cite{nissimovHairCurvatureNatural2014}. Some protein bonds can break when water is present and reform in different locations when water is removed, allowing hair to `reset' when washed. Others can only be broken by various chemical treatments intended to temporarily alter hair type. Highly anisotropic strands flex more readily, creating proximities that facilitate a higher number of protein bonds. The unbalanced forces that these bonds apply cause the strand to bend and twist. The emergent behavior of these phenomena can be captured with parameters for curl radius, curl wavelength, and twist. Curl radius describes the curvature of the strand when projected onto a plane perpendicular to the strand's core helical (or screw) axis. This can be specified as the radius of the cylinder that tightly contains a single curl, and it is straightforward to measure with some guidance. Curl wavelength, sometimes referred to as tightness, can be measured as the axial distance between two consecutive curls. Various forces (gravity, contact, etc.) can cause curl radius and wavelength to vary along a strand. Notably, such forces can even alter hair type if they are present when hair dries or when chemical treatments are applied, since they can create and remove the proximities that facilitate protein bonding.\label{sec:simulationjustification}

\begin{figure}[ht]
  \centering
  \includegraphics[width=0.4\textwidth]{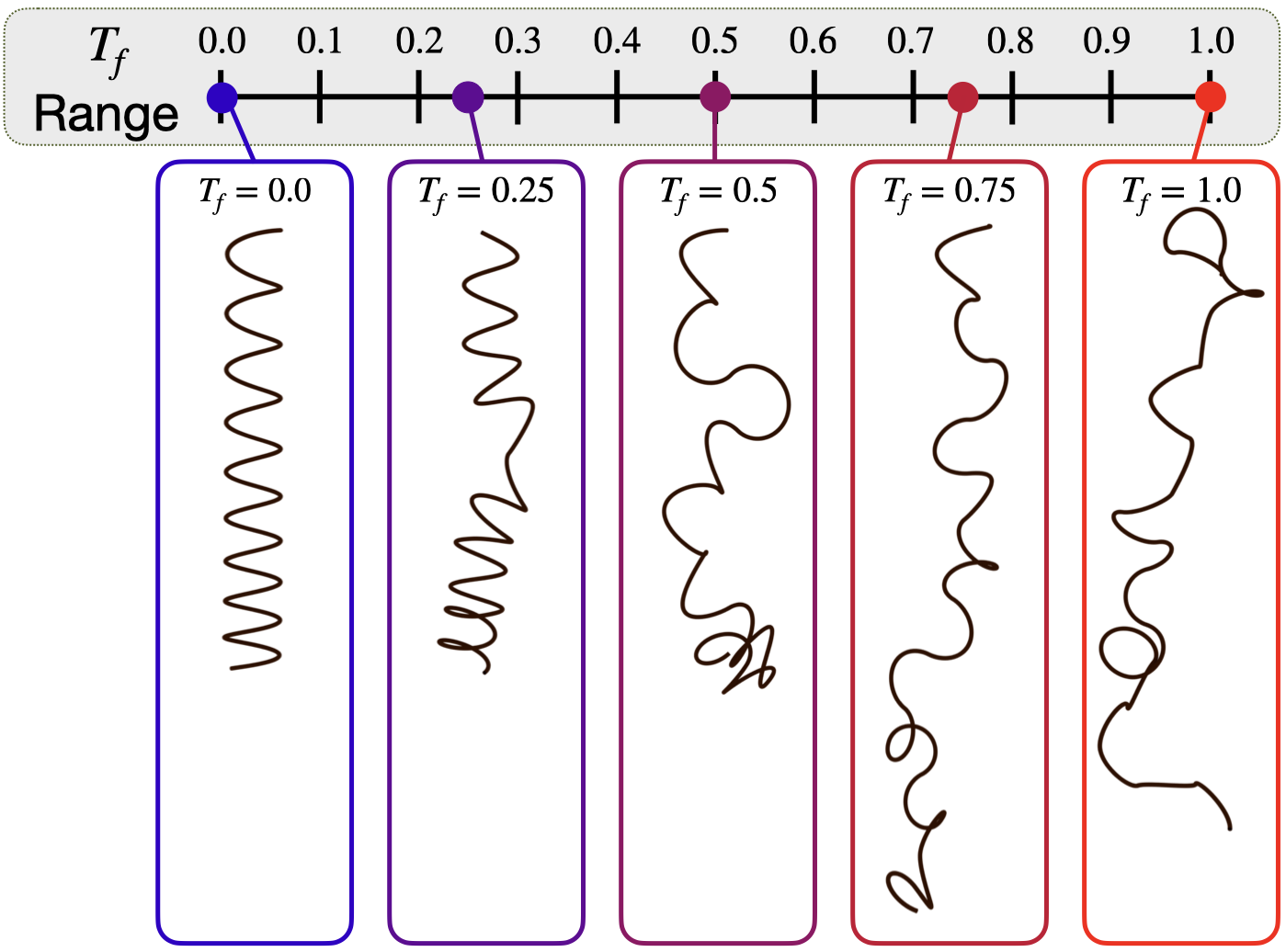}
  \caption{\footnotesize We offer a series of labeled visualizations to help users identify twist prevalence. After identifying hair type from these visualizations and accompanying descriptions, $T_f$ can be ascertained. A twist prevalence of 0 indicates that strands are helical with smoothly varying curvature, differing from each other only slightly. Low-medium twist prevalence strands ($T_f = .25$) exhibit rare deviations from the helix, often displaying `period skipping' before returning to a consistent repetitive structure. Medium twist prevalence strands ($T_f = .5$) spend about half of their length in helical repitition while exhibiting long interspersed periods of non-linear behavior. Medium-high twist prevalence strands ($T_f = .75$) have some visible helical curls but mostly exhibit non-helical formations. High twist prevalence strands curl heterogeneously with some sharp twists, lacking any visible helical segments. Note that it is straightforward to approximately interpolate $T_f$ values in between two labeled visualizations.}
  \Description{Visual guide showing different levels of twist prevalence in hair strands, demonstrating how users can identify their hair type from labeled reference photos to determine the appropriate twist prevalence parameter.}\label{fig:twist_prevalence}
\end{figure}

\subsubsection{Twist Prevalence} 
Twisting, like curling, is an emergent behavior caused by intra-strand protein bonding. It is distinct from curling in that it describes a rotation about the axial core of the strand itself. Unlike curls, which tend to be smoothly distributed along the strand, twist is typically localized to a discrete event with a random magnitude and direction~\cite{hairScans}; thus, we use the term twist prevalance to describe the average number of discrete twists per unit arclength. Twists cause a variety of non-periodic behaviors that are difficult to measure quantitatively. Therefore, we ask users to provide a qualitative description of their hair type by locating it in a series of labeled visualizations (see Figure~\ref{fig:twist_prevalence}). Many colloquial hair classifiers (such as~\cite{typeFromImage}) discuss twist, and terms that describe highly twisting strands (such as `kinky' or `afro-textured') are quite common; thus, twist prevalance will be familiar to many users, especially those for whom the parameter is most relevant.

\subsubsection{Porosity} Hair can absorb external molecules such as water, chemicals, etc.\ causing strands to swell, increasing their volume and weight~\cite{porosityEffect}. Notably, water can even be absorbed from or lost to the air depending on the humidity/aridity of the environment. The ability to absorb water differs from person to person and is dictated by hair strand porosity~\cite{hairProperties}. Since pores are formed by raised cuticles on the exterior of the strand, more porous hair also has larger surface friction. Moreover, porosity-influenced volume and weight influence hair behavior, and thus both modeling and simulation. See e.g.~\cite{hairwater, kimPorousModelsEnhanced2025, Fei2019}; although, to the best of our knowledge, prior works do not address increased friction or environmental attraction due to porosity. While porosity can change due to external factors such as heat or chemical interaction, each person's hair is typically predisposed to have a certain porosity. Porosity can be measured by comparing either cross-sectional area or weight after hydration or dehydration, although this can be difficult for non-experts; alternatively, a rough estimate can be ascertained by placing a strand into water to see if it sinks or floats~\cite{measurePorosity}.

\begin{figure}[ht]
  \centering
  \includegraphics[width=0.2\textwidth]{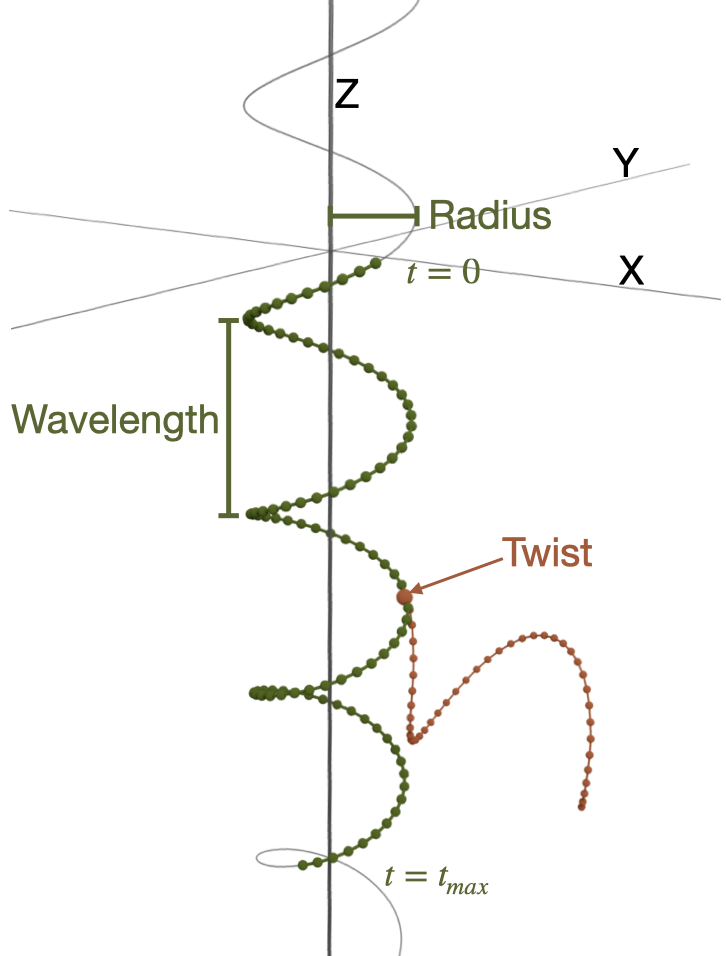}
  \caption{\footnotesize A helical segment illustrating curl radius, wavelength, and twist.}
  \Description{A 3D visualization of a helical hair segment showing the geometric parameters of curl radius, wavelength, and twist that define hair texture characteristics.}\label{fig:procedural_strand}
\end{figure}

\section{Strand Creation}\label{sec:constructing_strands}
In spite of the fact that our texture parameters are both well-motivated by the scientific community and accessible to artists and stylists, it is unfortunately not straightforward to construct hair strands from this parameterization. Although it is straightforward to procedurally generate helices from radius and wavelength, to subsequently add twisting events based on a prescribed prevalence, and to assign weight based on a combination of thickness, porosity, and humidity (or wetness), the visually interesting features that emerge when subjecting the strand to stress and strain via physics-based simulation cause the texture parameters of the resultant strand to differ significantly from those that were used to construct it. For example, gravity can pull hair downwards decreasing curl radius and increasing wavelength, twist angles can be pulled into alignment yielding single curls that appear stretched (a phenomenon referred to as period skipping~\cite{Curly-Cue}), etc.. These perturbations are typically more pronounced for higher porosity hair, which quickly gains mass when exposed to water and quickly loses mass when exposed to air. Additionally, thickness effects both stiffness (due to a higher number of protein bonds) and weight.

We embrace generative AI in order to generate post-simulation strands with the desired texture parameters directly, addressing the aforementioned issues. Supervision of the generative process requires the ability to describe (or label) observable or computable properties of the content at every step; specifically, we need to be able to assign our texture parameters to candidate strands. Labels are required not only for supervision but also for the dataset that will be used in training. We discuss dataset creation in the rest of this section and address labeling in the next section.

\subsection{Procedural Modeling}
\label{sec:strand_modeling}
Procedural modeling has had a significant impact on computer graphics, see e.g. L-Systems~\cite{Lsystems}, fractals~\cite{Fractals}, and Perlin noise \cite{perlinnoise}; thus, we begin the strand construction process procedurally. First, we initialize a simple helical structure:  \[
\vec{X}(t) = 
\left(
r(t) \cos\left(\frac{2\pi}{\lambda(t)} t\right),\ 
r(t) \sin\left(\frac{2\pi}{\lambda(t)}t\right),\ 
-t
\right)
\]
with a curl radius $r$ and wavelength $\lambda$, both of which can vary along the strand. A $t \in \left[0, t_{max}\right]$ subset of the helical structure is used with $t_{max}$ chosen to obtain the desired strand length. A discretized strand, $\mathbf{S}$, is computed by choosing  enough points to adequately resolve $\vec{X}$ and subsequently connecting them together with line segments. Since the maximum twist prevalence is about 1.5 twists per centimeter~\cite{hairScans} and $T_f \in [0, 1]$, we randomly choose $1.5 T_f |\mathbf{S}|$ interior points of the strand (ignoring endpoints) to receive twist. Here, the arc length $|\mathbf{S}|$ is computed as the sum of the lengths of all the segments of the strand. At each selected twist point, we rigidly rotate the entire downstream portion of the strand about the axis tangent to the prior segment. Reflecting biological patterns~\cite{hairScans}, the amount of twist is chosen randomly from $\left(-\pi, \pi\right)$. See Figure~\ref{fig:procedural_strand}. 

\begin{figure}[ht]
  \centering
  \includegraphics[width=0.3\textwidth]{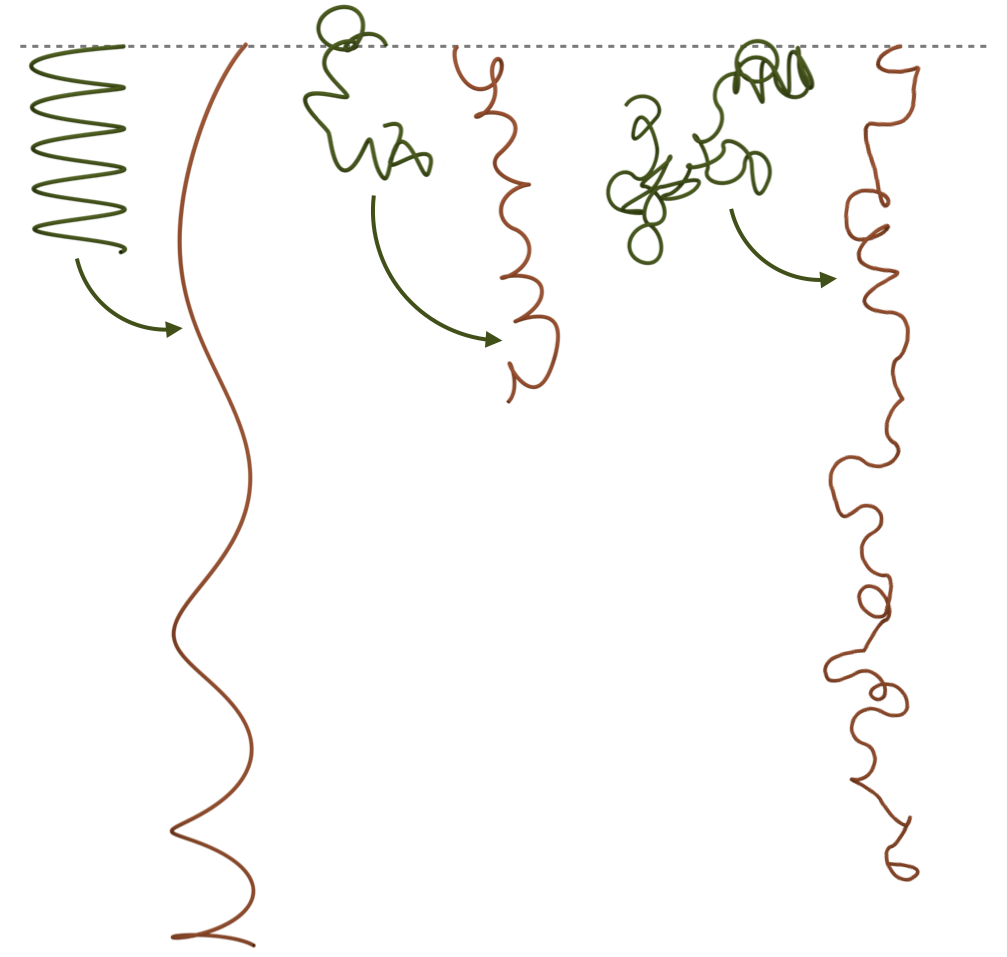}
  \caption{\footnotesize Three hair strands depicted both before (green) and after (red) physics-based simulation. The top end of the strand is fixed/constrained, while the rest of the rest of the strand responds to a downward gravitational force. (a) The increased mass acting upon the upper portion of the strand causes decreased radius and increased wavelenth. (b) When stretched, twist angles cause the helix to appear to be missing a curl . (c) Many non-linear behaviors result from the interactions of twist, curvature, and gravity. }
  \Description{Simulated and rendered curls highlighting the effects described in the caption.}
  \label{fig:simulation_effects}
\end{figure}

\subsection{Simulation }\label{sec:simulation}

In order to capture the myriad of visually interesting features caused by stress and strain, we subsequently simulate the procedurally modeled strands utilizing~\cite{Hsu2025}. See Figure~\ref{fig:simulation_effects}. Since the effects of stress and strain are baked into the strand structure (via chemical bonds) as it drys~\cite{chemicalBondsTypes}, it is important to simulate wet (not dry) hair. While this claim has not yet been embraced by the computer graphics literature, it is well-known in other communities: for example, this is why water is used to reset stubbornly out of place hairs and why removing a hair from the head and laying it out horizantally on the table does not modify its shape as much as one might expect. High porosity hair complicates this, since it not only wets faster when exposed to water but also dries faster when exposed to air. This means that the parts of a strand that are exposed to air can dry more quickly than those that are shielded from the air by other strands. For example, if the lower portion of the strand were to dry more quickly than the upper portion, the upper portion might recurl; otherwise, the upper portion would remain stretched out after drying. For lower porosity hair, which dries slowly and evenly, these variations are less notable. 

Higher porosity strands are assigned an increasing density from root to tip to reflect the gravitationally induced pooling of water at the bottom of the strand, whereas lower porosity strands are assigned a uniform density. This allows higher porosity strands, which can contain a lot of water, to be more stretched out due to gravity causing a high rate of parameter change along the strand. Thickness affects both the stiffness (due to a higher number of protein bonds) and the weight; thus, thicker hair can exhibit accentuated high porosity effects. The mass is computed from the density and thickness, and the quaternion rest state is initialized by sampling a material frame along each strand. Each strand geometry is simulated a number of times with varying choices of porosity, thickness, and stiffnesses in order to augment data. In addition, each parameter set is simulated to steady state twice: once beginning in the initial configuration and once beginning in a stretched out configuration (both results are kept only when the steady states differ). In order to capture the baked-in structure of dry hair, both the quaternion rest states and the masses are reset post-simulation before the steady state result is added to the dataset.

\subsection{Canonical Space \& Style Guides}\label{sec:canonical}

\begin{figure}[ht]
  \centering
  \includegraphics[width=0.47\textwidth]{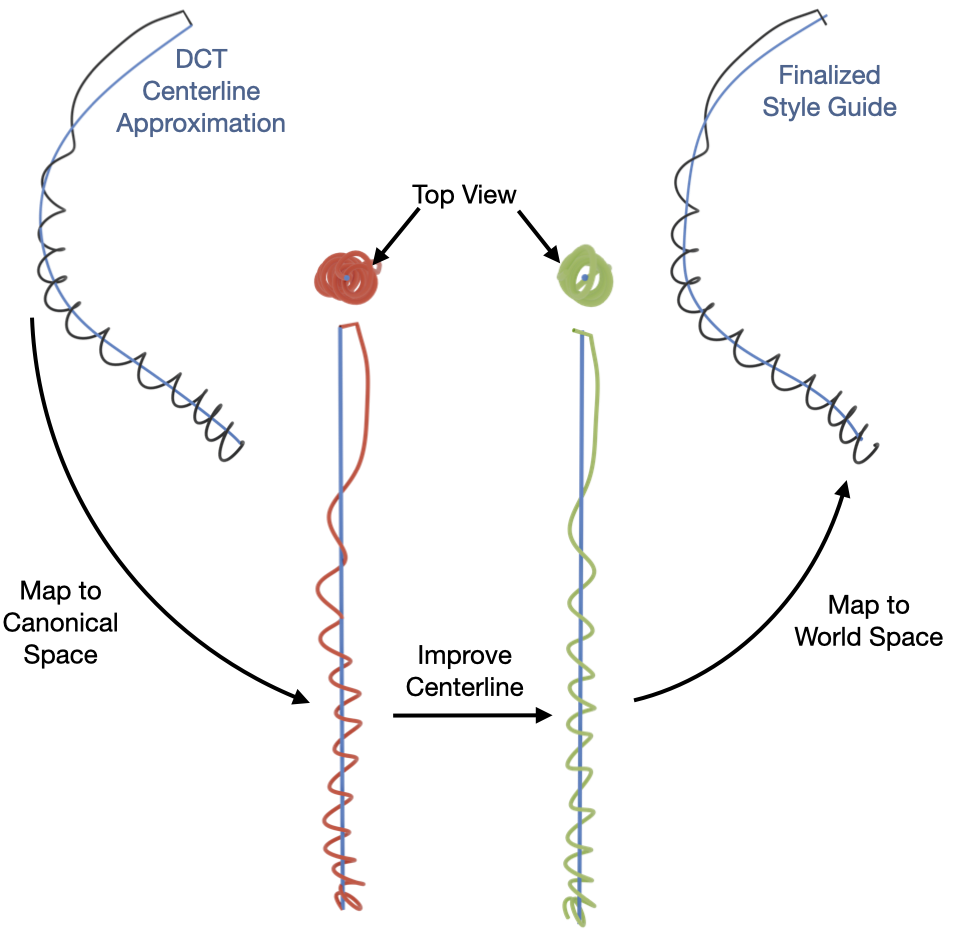}
  \caption{\footnotesize Top left: The DCT-based method of~\cite{DCT} is used to bootstrap a rough approximation to the centerline (shown in blue) of a discretized strand $\mathbf{S}$. Bottom left: After mapping to the canonical space, the errors in centerline prediction become more obvious (note the low frequency oscilations in the red strand). Bottom right: We improve the centerline approximation to better disentangle texture from geometry (note how the green strand no longer exhibits the low frequency oscilations of the red strand). Top right: The improved centerline is mapped back to world space in order to obtain a style guide (shown in blue).}
  \Description{A flowchart illustrating the process of mapping a strand to and from the canonical space, showing the initial strand and approximate centerline, the strand and approximate centerline in canonical space, the optimized centerline and strand in canonical space, and the strand and centerline mapped back to world space. These images are shown in counter clockwise order.}\label{fig:style_guide}
\end{figure}

Procedurally modeled and subsequently simulated strands, even with identical texture parameters, can look quite different when placed on the head. A strand within the bangs over the forehead might be cut short and brushed off to the side, a strand near the temple might be tucked behind the ear, and a strand atop the head might flow upwards briefly before gravity bends it downwards. In order to reduce variation, we disentangle strand texture from the dynamically changing strand geometry describing the geometry via style guides (similar in spirit to centerlines, guide hairs, etc). Motivated by NeRFs~\cite{DeformableNerfs, HyperNeRF, NeRF} and Gaussian Splatting~\cite{dynamicGaussians, GaussianAvatar, gaussiansplats}, we embrace the mapping of disparate data into a common canonical space in order to reduce variation, artificially augmenting the data to feature ratio. This also facilitates more robust regularization for the sake of neural modeling and generative artificial intelligence. Figure~\ref{fig:style_guide} summarizes how we compute a style guide $\mathbf{G}$ that disentangles geometry from texture for a discretized strand $\mathbf{S}$.
 
Given a discretized strand $\mathbf{S}$ with similarly (one to one) discretized centerline $\mathbf{C}$, we define a function $\hat{\mathbf{S}} = \phi(\mathbf{S};\mathbf{C})$ that maps $\mathbf{S}$ into canonical space as follows: First, a global translation (applied to both $\mathbf{S}$ and $\mathbf{C}$) is used to align the root endpoint of centerline $\mathbf{C}$ with the origin. Then, we compute the minimal rotation required to align each segment of $\mathbf{C}$ (from root to tip) with the negative z-axis. The rotation applied at particle $i$ of $\mathbf{C}$ rotates all particles $j > i$ of $\mathbf{C}$ and $j \geq i$ of $\mathbf{S}$, rotating all outboard segments rigidly (as is typical for articulated joints). As a result, all canonical space strands $\hat{\mathbf{S}}$ have centerlines coincident with the negative z-axis. Finally, we rediscretize the canonical space centerlines so that the particles on the new centerline $\hat{\mathbf{C}}$ have the same z-values as the corresponding particles on $\hat{\mathbf{S}}$.

As noted in Figure~\ref{fig:style_guide}, we use the frequency-based method from~\cite{DCT} to bootstrap a rough approximation $\mathbf{C}_{_{DCT}}$ to the centerline noting that $\mathbf{C}_{_{DCT}}$ will have the same number of vertices and connecting segments as $\mathbf{S}$ (otherwise, resampling would be required before $\phi$ could be applied). Unfortunately, $\hat{\mathbf{S}}_{_{DCT}} = \phi(\mathbf{S}; \mathbf{C}_{_{DCT}})$ has obvious deficiencies. The approach from~\cite{DCT} not only assumes that the endpoints of $\mathbf{S}$ and $\mathbf{C}_{_{DCT}}$ are coincident, but it also fails to adequately predict the centerline (as evidenced by the low frequency oscilations in $\hat{S}_{_{DCT}}$ in the figure). We improve upon their result by adjusting the x-values and y-values of $\hat{\mathbf{C}}_{_{DCT}}$, while holding the z-values fixed, aiming to find a $\hat{\mathbf{C}}_{_{OPT}}$ that better approximates the centerline. 
This is accomplished by minimizing 
\begin{align}
\frac{1}{2}\sum_i \left(||\vec{r_i}||_{_{2}} - \frac{1}{n}\sum_j||\vec{r_j}||_{_{2}} \right)^2  
\label{eq:dct_regularization}
\end{align}
where $||\vec{r_i}||_{_{2}}$ is the distance between vertex $i$ on $\hat{\mathbf{C}}_{_{OPT}}$  and the corresponding vertex on $\hat{\mathbf{S}}_{_{DCT}}$. Since equation~\ref{eq:dct_regularization} is trivially minimized when all the ${||\vec{r_i}||}_{_{2}}$ are equal to the mean, the key to good disentanglement lies in the choice of the regularization terms. Thus, we add regularization to minimize the difference in  ${||\vec{r_i}||}_{_{2}}$ between adjacent vertices, the standard deviation of the vertex displacements measured from $\hat{\mathbf{C}}_{_{DCT}}$ to $\hat{\mathbf{C}}_{_{OPT}}$, and the curvature of $\hat{\mathbf{C}}_{_{OPT}}$:
In order to keep the regularization terms robust near both the root and the tip, we extend the strands a bit in each direction before optimization. This is accomplished at each endpoint by duplicating and reflecting about 10\% of the vertices before aligning the endpoint material frames. The new $\hat{\mathbf{C}}_{_{OPT}}$ is aligned with the negative z-axis using $\phi(\hat{\mathbf{S}}_{_{DCT}};\hat{\mathbf{C}}_{_{OPT}})$ to obtain a new $\hat{\mathbf{S}}$ and $\hat{\mathbf{C}}$. 

Finally, we need to determine a world space style guide $\mathbf{G}$ that can be used to map our canonical space strand $\hat{\mathbf{S}}$ back into world space via the inverse function $\mathbf{S} = \phi^{-1}(\hat{\mathbf{S}};\mathbf{G})$. $\phi^{-1}$ uses the mapping of our improved centerline $\hat{\mathbf{C}}$, which lies on the negative-$z$ axis, to the world space style guide $\mathbf{G}$ in order to guide the mapping of our improved canonical space strand $\hat{\mathbf{S}}$ to the original world space $\mathbf{S}$. Similar to $\phi$, a global translation is used to align the root endpoints of $\hat{\mathbf{C}}$ and $\mathbf{G}$, root-to-tip rotations are used to make $\hat{\mathbf{C}}$ coincident with $\mathbf{G}$, and rotations applied at particle $i$ rotate all particles $j>i$ of $\hat{\mathbf{C}}$ and $j \geq i$ of $\hat{\mathbf{S}}$. Notably, defining $\mathbf{G} = \phi^{-1}(\hat{\mathbf{C}}_{OPT}; \mathbf{C}_{DCT})$ yields $\mathbf{S} = \phi^{-1}(\hat{\mathbf{S}};\mathbf{G})$ as desired; moreover, $\phi(\mathbf{S}; \mathbf{G})$ gives our improved $\hat{\mathbf{S}}$ and $\hat{\mathbf{C}}$ (as expected). Since $\mathbf{G}$ serves as a destination for $\hat{\mathbf{C}}$, $\mathbf{G}$ requires a 1:1 segment correspondence with $\hat{\mathbf{C}}$ and the segments of $\mathbf{G}$ need to be the same length as the corresponding segments of $\hat{\mathbf{C}}$. When $\mathbf{G}$ is defined via $\phi^{-1}(\hat{\mathbf{C}}_{OPT}; \mathbf{C}_{DCT})$ this is implicit, but moving $\hat{\mathbf{S}}$ to a different style guide might require resampling of either $\mathbf{G}$ or $\hat{\mathbf{C}}$.

\section{Labeling Texture Parameters}
\label{sec:differentiable_curves}

Via procedural modeling (Section~\ref{sec:strand_modeling}) and simulation (Section~\ref{sec:simulation}), we are able to create a large variety of realistic hair strands and map them to a canonical space (Section~\ref{sec:canonical}) where it is more straightforward to disentangle texture from geometry. Although it is just as easy to determine texture parameters for synthetically generated strands as it is for real-world strands, it is intractable to manually supervise every step of a generative AI process; thus, we devise an automatic method that assigns texture parameters to strands created synthetically by neural generation, artists, the process described in Section \ref{sec:constructing_strands}, or other means.

\subsection{Differentiable Curves}
\label{sec:ktv_representation}
In order to minimize the amount of training data required, we bootstrap the inference by first computing quantities that are more informative than raw vertex positions. In particular, we utilize a curvature-torsion-speed representation, which saliently distills information directly relevant to our texture parameters (curl radius, curl wavelength, twist prevalence). Given a parameterized curve $\vec{\gamma}(t)$, curvature, torsion, and speed are defined via 
\begin{align*}
\kappa(t) &= \frac{\left<\vec{T}'(t), \vec{N}(t)\right>}{||\vec{\gamma}'(t)||}  = \frac{||\vec{T}'(t)||}{||\vec{\gamma}'(t)||} \\
\tau(t) &= \frac{\left< \vec{N}'(t), \vec{B}(t)\right>}{||\vec{\gamma}'(t)||} \\
v(t) &= ||\vec{\gamma}'(t)|| 
\end{align*} where the tangent, normal, and binormal are defined via
\begin{align*}
\vec{T}(t) &= \frac{\vec{\gamma}'(t)}{||\vec{\gamma}'(t)||} \\
\vec{N}(t) &= \frac{\vec{\gamma}''(t) - \left< \vec{\gamma}''(t), \vec{T}(t) \right> \vec{T}(t)}{||\vec{\gamma}''(t) - \left< \vec{\gamma}''(t), \vec{T}(t) \right> \vec{T}(t)||} \\
\vec{B}(t) &= \vec{T}(t) \times \vec{N}(t)
\end{align*} respectively \cite{Frenet}. We select these Frenet-Serret equations over the commonly used Bishop Frame equations~\cite{Bishop1975} because they capture moments of twist (crucial for determining twist prevalence) rather than minimizing them.

As shown above, the curvature-torsion-speed representation relies on $\vec{\gamma}$, $\vec{\gamma}'$, $\vec{\gamma}''$, and $\vec{\gamma}'''$. Unfortunately, the inconsistent continuity and resolution in the procedurally generated or artist created raw vertex positions leads to extreme spikes in the (especially higher) derivatives limiting the ability of our neural model to extract high-quality information. In addition, normalizing the strand resolution (for training) requires resampling, which leads to aliasing artifacts that pollute the inputs to texture parameter inference. Since the undersampled high frequencies alias onto lower frequencies, it is impossible (even for a neural network) to remove the aliasing artifacts after sampling (as is well known in signal processing). Moreover, interpolating (especially higher) derivatives of $\vec{\gamma}$ spreads the noisy spikes to other regions. Thus, as is typical, we smooth the signals to remove potentially-aliasing higher frequency signals before resampling.

Fundamentally, the raw vertex positions are conceptualized as a smooth curve by an artist (or a numerical algorithm that penalizes variance). A piecewise cubic spline reasonably captures the clear, or intended, signal. We chose a piecewise cubic representation to get as much continuity as possible (it has $C^2$ continuity, unlike Catmull-Rom and Bezier splines~\cite{CatmullRom,bezier}) without the increased oscillatory noise that accompanies higher order interpolations. Given $n$ vertices and $4(n-1)$ cubic parameters in each dimension, the $2(n-1)$ interpolatory and $2(n-2)$ first and second derivative continuity constraints leave 2 undetermined parameters per dimension; thus, an additional derivative boundary condition on each end of the strand is determined by interpolating a cubic through the first/last four points. Letting $t$ parameterize the arc length along the piecewise-linear discretized strand $\mathbf{S}$, the piecewise cubic can be written as $\vec{X}(t)$ in the canonical space.

Given this clean signal, we densely and uniformly sample $\vec{X}(t)$ along $t$ to create a piecewise linear curve. This dense sampling (super-sampling) is meant to properly capture the bulk of the problematic high frequencies. Next, we identify per-channel edge lengths (i.e., $\scriptstyle\sqrt{(\Delta x)^2 + (\Delta t)^2}$ or $\scriptstyle\sqrt{(\Delta y)^2 + (\Delta t)^2}$ or $\scriptstyle\sqrt{(\Delta z)^2 + (\Delta t)^2}$ ) with a length at least three standard deviations larger than the mean. Then, per-channel adaptive Gaussian smoothing with an edge-length correlated kernel is applied in regions containing an offending edge iteratively until all edges are within three standard deviations of the mean length. Finally, the smoothed piecewise linear curve is re-interpolated to create a new piecewise cubic spline, $\vec{\gamma}(t)$. Note that $\vec{\gamma}$ will not necessarily interpolate the vertices of $\mathbf{S}$ (as $\vec{X}$ does). 

Even though $\vec{X}$ could be analytically differentiated twice due to its $C^2$ continuity, sharp bends and twists in $S$ as well as issues with polynomial interpolation (e.g. overfitting) introduce high-variance into the derivatives. Sampling these high-variance signals leads to aliasing artifacts polluting the inputs to texture parameter inference. Thus, we prefer to analytically differentiate the smoothed $\vec{\gamma}$. After differentiating, the result is densely sampled, smoothed, and re-interpolated to obtain $\vec{\gamma}'$. Note that $\vec{\gamma}'$ is inconsistent with $\gamma$, just as $\vec{\gamma}$ is inconsistent with $\mathbf{S}$; however, it is important to remove any potentially-aliasing higher frequencies in $\vec{\gamma}'$ that were introduced by differentiating $\vec{\gamma}$. Although one might, alternatively, aim to smooth $\vec{\gamma}$ enough such that $\vec{\gamma}'$ does not contain potentially-aliasing higher frequencies, this incorrectly removes resolvable signals in $\gamma$. The process is repeated to obtain $\vec{\gamma}''$ and $\vec{\gamma}'''$. As a final note, one could replace reinterpolation plus analytic differentiation with numerical differentiation if desired, as long as the data has been smoothed enough for numerical differentiation to be viable.

 Given these smoothed $\vec{\gamma}$'s, we proceed as follows: The unit tangent can be computed as long as $||\vec{\gamma}'(t)||$ is large enough to avoid numerical overflow. When $||\vec{\gamma}'(t)|| < \epsilon$ for some small $\epsilon > 0$, we perturb the coordinate of $\vec{\gamma}'$ that has the least local variation to have a value of $\epsilon$. This perturbation to $\vec{\gamma}'$ is done after sampling the differentiated $\vec{\gamma}$, but before smoothing. The unit normal is computed by normalizing the portion of $\vec{\gamma}''$ orthogonal to $\vec{T}$.  When $||\vec{\gamma}''(t)|| < \epsilon$, it is perturbed in the exact same manner as $\vec{\gamma}'$ was. When $\vec{\gamma}''$ is too parallel to $\vec{T}$ for the orthogonal component to be meaningful, we linearly interpolate the direction of $\vec{N}$ from the closest well-defined normals in each direction (i.e., interpolate along the angle between the vectors). Constant extrapolation is used when only one direction has a well-defined normal. When there are no normals on the entire strand (e.g. a straight line), an arbitrarily-chosen single value for $\vec{N}$ is used along the entire strand. After computing $\vec{T}$ and $\vec{N}$ with robust denominators, $\vec{B}$, $\kappa$, $\tau$, and $v$ are straightforward. When $\vec{N}$ is defined via either interpolation or extrapolation, $\vec{N}'$ is computed (in the definition of $\tau$) by differentiating through the interpolation/extrapolation formulas. The resulting discontinuity of $\vec{N}'$ at the boundary creates a discontinuity in $\tau$ signifying a twist. This is crucial for hair, which can contain moments of twist even when curvature vanishes.

\begin{figure*}
  \includegraphics[width=\textwidth]{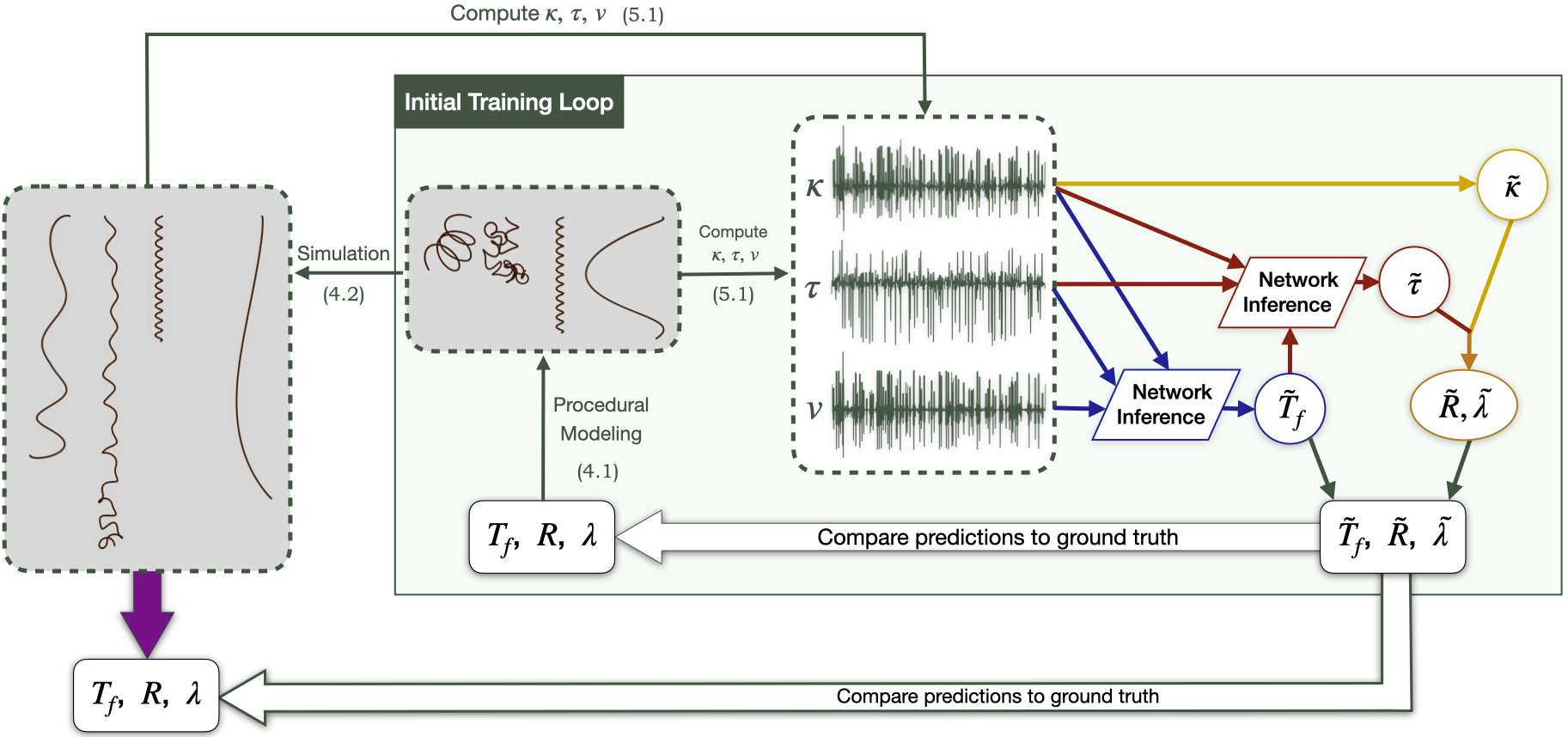}
  \caption{\footnotesize Inferencing strand texture parameters: To create a dataset for training, we construct strands following Section~\ref{sec:constructing_strands}. Then, the process described in Section~\ref{sec:ktv_representation} is used to compute curvature, torsion, and speed. A network is trained to inference $\tilde{T}_f$ directly from the computed values of curvature, torsion, and speed. A second network inferences $\tilde{\tau}$ from $\tilde{T}_f$ and the computed values of curvature and torsion. $\tilde{\kappa}$ is determined from the low-frequency components of the computed curvature and combined with $\tilde{\tau}$ to obtain $\tilde{R}$ and $\tilde{\lambda}$. Ground truth values for $R$ and $\lambda$ are compared to $\tilde{R}$ and $\tilde{\lambda}$ in order to train the second network (for inferencing $\tilde{\tau}$). The second network is first trained on a large number of unsimulated strands (that do not require hand labeling), before it is finetuned on a relatively low number of hand-labeled simulated strands. The large purple arrow highlights where some hand-labeling is required.}\
  \Description{A flowchart illustrating the process of learning to get inferenced parameters back from existing strands. }
  \label{fig:draft_diagram}
\end{figure*}

\subsection{Inferencing texture parameters}
\label{sec:inferencing_parameters}

Although one could create vast quantities of synthetic strands, label them all by hand, and train a neural network to correctly inference those labels, this quickly becomes prohibitive if not intractible. Instead,
we bootstrap the process by first training with a large number (approximately 50k) of procedurally modeled strands that are unsimulated (consistent with the parameters used to construct them) and thus do not require any hand labeling. Afterwards, a relatively low number (approximately 1k) of simulated strands, which require some hand labeling, are used for finetuning. This distribution shift minimizes the amount of hand-labeled data required to overcome the domain gap between procedurally modeled and physically realistic strands. Figure~\ref{fig:draft_diagram} separates the input strands into two groups (i.e. two gray boxes) in order to stress the difference in treatment required for procedurally modeled (according to Section~\ref{sec:strand_modeling}) versus subsequentially simulated (according to Section~\ref{sec:simulation}) strands.

We begin by computing curvature, torsion, and speed following Section~\ref{sec:ktv_representation}. Although the smoothing proposed in Section~\ref{sec:ktv_representation} removes extreme spikes in the (especially higher) derivatives as well as other high frequencies that cause aliasing when resampling, it is not intended to produce smooth outputs. The goal is merely to omit spurious information caused by aliasing; in fact, after resampling $\gamma$ and its derivatives, the computed quantities ($\vec{T}$, $\vec{N}$, $\vec{B}$, $\kappa$, $\tau$, $v$) may contain very high frequencies since their denominators are only protected against division by zero or overflow. We emphasize this by including an especially high frequency example in the graphs in Figure~\ref{fig:draft_diagram}. Given curvature, torsion, and speed computed along the strand, we then proceed to obtain user-consistent, interpretable labels for radius $\tilde{R}$, wavelength $\tilde{\lambda}$, and twist prevalence $\tilde{T}_f$. This is accomplished with the aid of a number of procedural methods along with two neural networks. $\tilde{T_f}$ is inferenced first, and then the result is used to infererence a characteristic torsion $\tilde{\tau}$. A procedural approach is used to determine a characteristic curvature $\tilde{\kappa}$, and then 
\begin{equation*}
  \tilde{R} = \frac{\tilde{\kappa}}{\tilde{\kappa}^2 + \tilde{\tau}^2} \quad \text{and} \quad \tilde{\lambda} = \frac{2 \pi \tilde{\tau}}{\tilde{\kappa}^2 + \tilde{\tau}^2}
  \end{equation*}
are computed via the helix equations. See Figure~\ref{fig:draft_diagram}. 

\subsubsection{Inferencing twist prevalence: } The first neural network (shown in blue in Figure~\ref{fig:draft_diagram}) infers $\tilde{T}_f$ from the computed values of curvature, torsion, and speed. We employ a bi-directional LSTM~\cite{Hochreiter1997} to capture the temporal dependencies and geometric correlations along the strand aiming to recognize the simultaneous data patterns induced by twist into all three channels. Attention \cite{Bahdanau2016} is used to compute adaptive weights along the signals, enabling the prioritization of geometry that reflects signal patterns indicative of twist. A single-valued final prediction is determined via an MLP containing three fully connected layers (256→128→64→1) with ReLU activations and dropout regularization. The final layer applies a sigmoid activation to produce a $\tilde{T}_f \in [0,1]$. Since $T_f$ represents physical kinks in the strand, it doesn't change during simulation and the ground truth value is known for both procedurally modeled and subsequently simulated strands. Note that user labeling is guided to be consistent with those ground truth values via a series of labeled visualizations (see Figure~\ref{fig:twist_prevalence}); thus, it is straightforward to obtain a plethora of ground truth data.

\newpage
\subsubsection{Inferencing characteristic torsion: } Since real-world strands lack a uniform radius and wavelength (due to gravity, contact forces, etc.), we rely on a social understanding of how people describe hair to inform our construction of $\tilde{\kappa}$ and $\tilde{\tau}$. Specifically, we prioritize the more visually prominent and typically more highly textured lower sections of strands. The second neural network (shown in red in Figure~\ref{fig:draft_diagram}) determines a single torsion value $\tilde{\tau}$ to characterize the lower section of the strand. In order to avoid twist-induced noise, we prioritize lower twist regions of the strand with the aid of five input channels: In addition to $\kappa$, we also include the low-frequency values of $\kappa$ to provide overall geometric context. Note that $\kappa$ being approximately equal to its low frequency version indicates that a region is less affected by twist. In addition to $\tau$, we also include a weighted version of $\tau$ computed using the inverse of the normalized Laplacian kernel of the difference between $\kappa$ and the low-frequency version of $\kappa$. This identifies and prioritizes (by using larger weights in regions where the difference is smaller) regions where $\tau$ is less affected by twist. $\tilde{T}_f$ is included to provide guidance on how much noise should be expected. The model utilizes a transformer-encoder architecture with multihead attention \cite{Vaswani2023} in order to capture long-range and multi-scale dependencies. Pre-layer normalization \cite{Xiong2020} increases gradient stability, and mean pooling smooths twist-induced spikes. A final MLP containing three fully connected layers (256→256→128→1) with ReLU activations and dropout regularization outputs out final $\tilde{\tau}$.

\subsubsection{Determining characteristic curvature: } Fortuitously, we found frequency decomposition to be an effective tool for separating twist-induced noise from the computed curvature information, removing the need for a neural network or other data driven approaches. For simulated strands, where radius and wavelength vary along the strand yielding no single obvious ground truth value for curvature, the low frequency component also varies smoothly along the strand. Thus, it is robust to approximate the characteristic curvature as the average of the low frequency components over the lower sections of the strand.

\subsubsection{Training}

The second neural network (for inferencing $\tilde{\tau}$) is first trained on large number (approximately 50k) of unsimulated strands using automatically-obtained ground truth labels for $R$ and $\lambda$. This is done after training the first neural network (for inferencing $\tilde{T}_f$) on the same unsimulated data; alternatively, the two networks could be trained together. After the second network is trained on the unsimulated data, the first network is retrained using simulated data. Then, the second network is finetuned using a relatively low number of hand-labeled simulated strands. We hand-labeled a set of one thousand hair strands across hair types in order to supervise finetuning. Using a ADA6000 GPU, the first and second neural networks only took a few hours each to train on 50k unsimulated strands and less than an hour each to finetune on 10k and 1k simulated strands respectively.

\section{Generative Modeling}\label{sec:generative_modeling}

Although procedural modeling plus simulation can produce a large variety of strand geometries, it is difficult to determine the pre-simulation parameters that produce the desired post-simulation strand types. Instead of aiming to solve this ill-posed inverse problem, we train a diffusion-based model to generate strands consistent with any desired texture parameters. In order to increase the efficacy of a diffusion-based approach, we disentangle texture from geometry by generating strands in the canonical space. This reduces the degrees of freedom, which reduces the amount of training data required and also improves the robustness of supervision. Via experimentation, we obtained the best results by training the diffusion-based model to output a curvature-torsion-speed representation (see Section~\ref{sec:ktv_representation}), which can be used to construct strand geometry with Frenet-Serret integration~\cite{Frenet}. This representation is more robust to perturbation than a position-based approach. For example, a local perturbation changes all downstream position-based information, while leaving downstream curvature, torsion, and speed intact. In addition, two strands with similar texture parameters have similar curvature, torsion, and speed, even though they may have vastly different position-based information. 
  
\begin{figure*}
  \includegraphics[width=.9\textwidth]{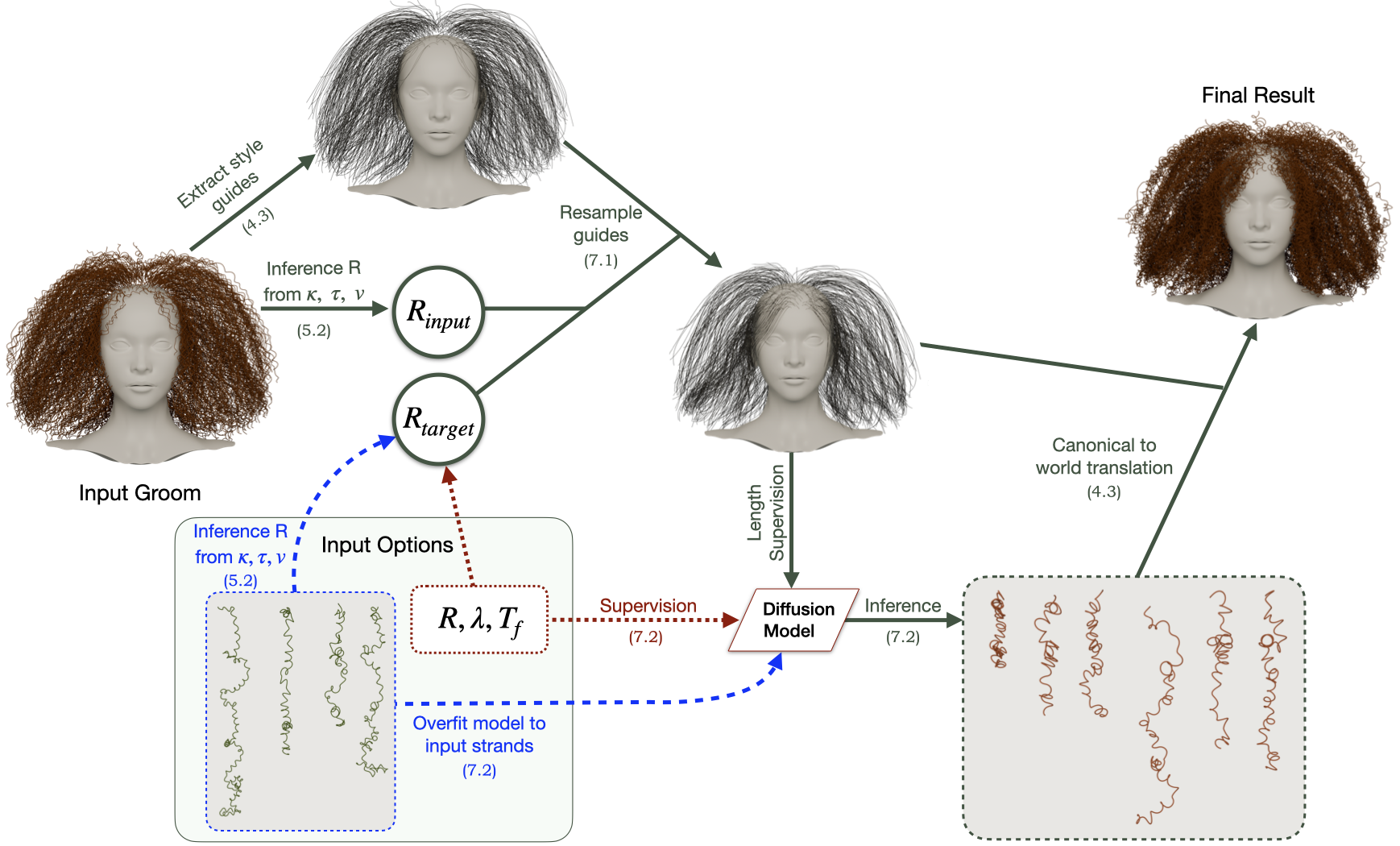}
  \caption{\footnotesize Texture editing pipeline. The new texture can be specified either via a small set of target strands or via texture parameters (blue and red arrows illustrate the only differences between the two approaches). After extracting style guides from the input groom (top left), the radii of the input and target textures are used to determine the amount of style guide resampling necessary to adequately support the new texture (see Section~\ref{sec:style_guide_resampling}). Once the style guides are resampled, their lengths are used as input to the diffusion model.  When the target texture is specified via a small number of strands, the diffusion model is overfit to those strands (blue arrow); otherwise, the pretrained model (from Section~\ref{sec:training}) is used (red arrow) as discussed in Section~\ref{sec:diffusion_integration}. The inferenced strands are lengthened/shortened if neccessary (see Section~\ref{sec:populating_style_guides}) before being mapped to the world space style guides (as detailed in Section~\ref{sec:canonical}).}
  \label{fig:full_pipeline}
  \Description{A flowchart illustrating the full pipeline for user-generated hair from either sample texture strands or target texture parameters. }
\end{figure*}

  \subsection{Data Generation and Curation}
  \label{sec:data_generation}  
 Aiming to represent the wide breadth of admissible texture parameter combinations while also prioritizing common parameter correlations, we collated a number of scientific studies containing empirical data on hair across different buckets. We begin with a 2007 L'Oreal report~\cite{Mettrie2007} that clusters samples from almost 1,500 subjects into eight descriptive buckets based on curve diameter, curl index, and wavelength. Using their groupings as a basis, incorporating measured data from other studies~\cite{Blume-Peytavi2008, Bernstein1927, hairProperties, hairScans, Loussouarn2001, hairDifferences}, and interpolating gaps in the data, we constructed mean and standard deviation values for radius, wavelength, twist prevalence, and length for each of the eight buckets. We artificially inflate the standard deviation to ensure the inclusion of less common hair types (that do not fit neatly into the eight catagories) when sampling from these distributions.

Given an $R$, $\lambda$, $T_f$ sample, we can procedurally create (Section~\ref{sec:strand_modeling}) and optionally subsequently simulate (Section~\ref{sec:simulation}) a strand before mapping it to a canonical space (Section~\ref{sec:canonical}) in order to compute $\kappa$, $\tau$, and $v$ (Section~\ref{sec:ktv_representation}) as well as length target values for training. Since simulation causes a shift away from the intended domain of a sample, we vary simulation parameters (primarily stiffness and weight) to create a number of strands for each sample. In this sense, the eight inflated buckets act more as a starting point than as strict constraints. Note that the methods proposed in Section~\ref{sec:constructing_strands} and Section~\ref{sec:ktv_representation} are only used for preprocessing before training and therefore do not require differentiability.

Shortcomings in existing software-based methods for generating afro-textured hair increase the risk of misrepresenting the community that the dataset aims to portray. Thus, in addition to procedurally generated and simulated strands, we also utilize expert artists who have experience working around these shortcomings. In particular, artists commissioned by this project were selected based on their experience and expertise creating curly and afro-textured grooms. Additionally, and importantly, the artist-created grooms were largely based on images of community members who provided iterative feedback in satisfactorily completing the groom.

\subsection{Architecture}
  \label{sec:architecture}
  We implement our generative model via diffusion where the denoiser is a time-conditioned 1D U-Net encoder/decoder, suitable for strand sequences~\cite{U-Net}. The network consists of four downsampling stages, each composed of residual blocks with stride-two convolution downsampling after each stage, along with mirrored upsampling. The numbers of channels after each downsampling stage are 64, 128, 256, and 256. We use the Swish activation function~\cite{Ramachandran2017} for nonlinearity and employ Group Normalization~\cite{wuGroupNormalization2018} to stabilize training. Self-attention~\cite{Vaswani2023} is inserted at the third level of both the downward and upward passes as well as in the bottleneck. The scalar time parameter is embedded into a 256-dimensional vector using the standard sinusoidal embedding~\cite{Vaswani2023}; subsequently, the result is injected into the network's residual blocks using a FiLM-style mechanism~\cite{FiLM}. Inspired by~\cite{Chen2023}, we condition the U-Net with the log-SNR parameter of the noise schedule instead of the raw diffusion time index. Since distance from the scalp can inform the strand texture, we augment the raw input of the network with absolute positional features: a linear ramp from root to tip as well as four sinusoidal embeddings of that ramp.

\subsection{Training}
\label{sec:training}
  We formulate our generative model as a variance-preserving denoising diffusion process parameterized to predict a clean signal under a cosine noise schedule~\cite{Nichol2021}. The denoiser is optimized to minimize the error between the (whitened) predicted and ground-truth curvature-torsion-speed representations. The optimization is carried out using AdamW~\cite{Loshchilov2019} with standard learning rate scheduling, gradient clipping, and mixed-precision acceleration. Using an RTX 4090 GPU, training on 50k simulated strands takes approximately 24 hours including strand generation and simulation. Training on a selection of existing strands takes less than a minute on a 4090 GPU and three minutes on an Apple M2 pro.

  \subsection{Inference}
  \label{sec:inference}
  We employ the multistep DPM-Solver~\cite{DPM-Solver++}, which provides high-quality reconstructions efficiently in only 20 denoising iterations. Inspired by Diffusion Posterior Sampling~\cite{Chung2024}, we use a measurement-driven guidance mechanism to enable controllable synthesis. At each denoising step, we utilize the (differentiable) method proposed in Section~\ref{sec:inferencing_parameters} to label the current predicted signal and compare it to a target value; then, backpropagation can be used to steer the signal towards a desirable outcome. 

\subsection{Strand Reconstruction}
\label{sec:strand_reconstruction}
Given the final predicted signal, i.e. $\kappa$, $\tau$, and $v$ (predicted at or interpolated to whatever resolution is desired), the Frenet-Serret equations~\cite{Frenet} 

\begin{align*}
\frac{d}{dt}\begin{bmatrix} \vec{T} \\ \vec{N} \\ \vec{B}\end{bmatrix} &= v \begin{bmatrix}
0 & \kappa & 0 \\
-\kappa & 0 & \tau \\
0 & -\tau & 0
\end{bmatrix}
\begin{bmatrix} \vec{T} \\ \vec{N} \\ \vec{B}\end{bmatrix} 
\end{align*}
can be integrated using fourth-order Runge-Kutta~\cite{Kutta1901} with an axis-aligned initial condition for $\vec{T}$, $\vec{N}$, and $\vec{B}$. At each step, we re-orthonormalize $\vec{T}$, $\vec{N}$, and $\vec{B}$ to control drift. Note that the extra factor of $v$ is due to the chain rule, incurred when differentiating by the parameter $t$ instead of arc length. Afterwards, we obtain position information by integrating $\frac{d\vec{X}}{dt} = v T$ using fourth-order Runge-Kutta with an $\vec{X}_0 = \vec{0}$ initial condition. $\vec{X}$ is then sampled/interpolated  in order to obtain a discretized strand $\mathbf{S}$, which is subsequently transformed into the canonical space.

\begin{figure}[ht]
  \centering
  \includegraphics[width=0.3\textwidth]{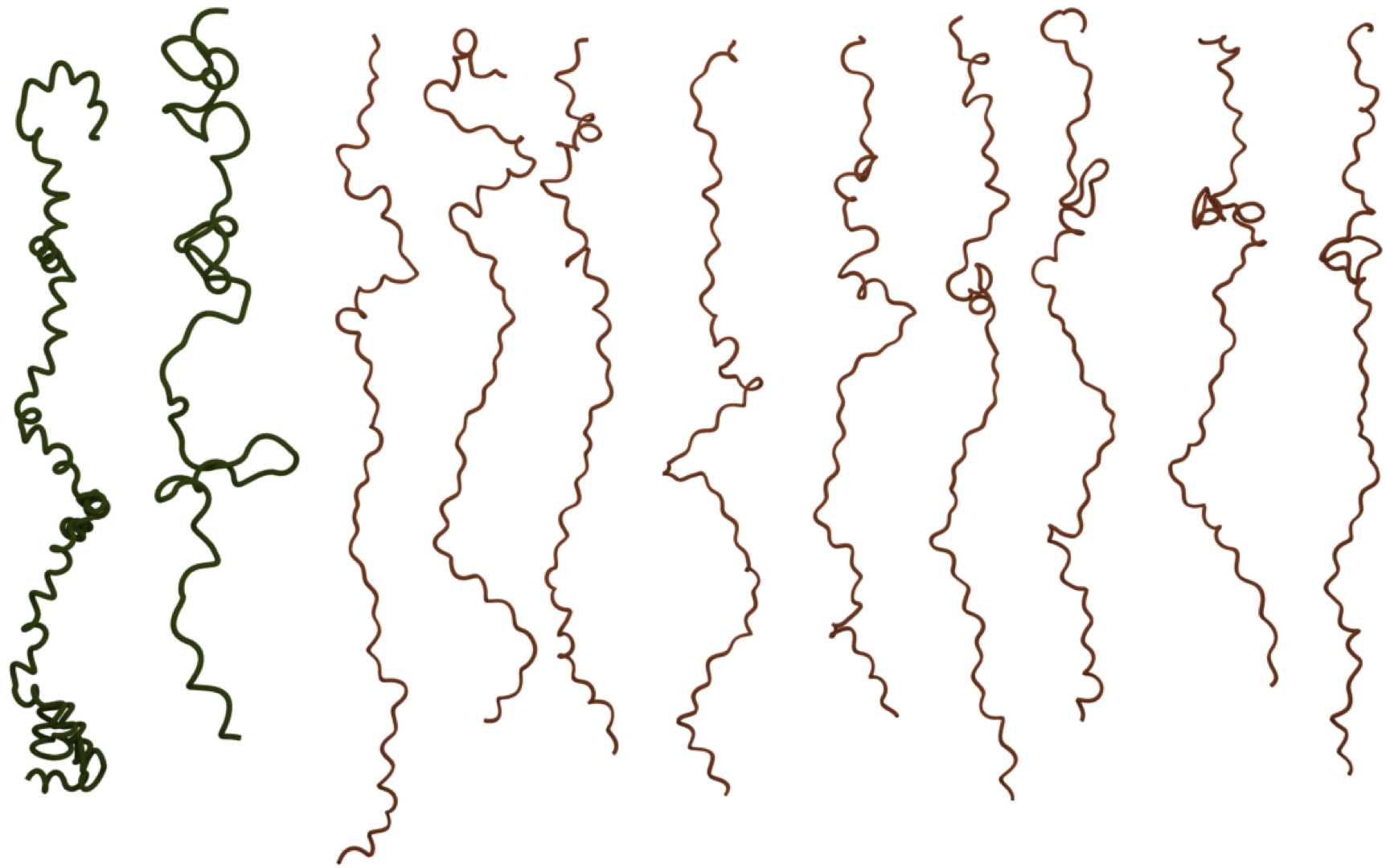}
  \caption{\footnotesize  The diffusion model is able to generate a diverse set of texture-consistent strands (red) even when it is trained on only two input strands (green). Note how various features (twisted bundles of curls, sections that jut to the left and right, and a hook-like curl at the root) of the input strands manifest differently in each generated strand while still preserving the overall look and feel of the texture.}
  \Description{A set of generated strands that are consistent with the texture of the input strands. }\label{fig:small_input}
\end{figure}

\section{Texture Editing} \label{sec:full_pipeline}{}{}

In order to demonstrate the efficacy of our approach we illustrate how to entirely replace the texture of an existing groom with a new desired texture. This requires minimal user effort, for texture specification, and minimal computational resources. We present two options for specifying a desired new texture in order to accomodate users ranging from novice to expert. See Figure~\ref{fig:full_pipeline}. For those that prefer to specify target parameters, the discussion in Section~\ref{sec:strand_texture} illustrates how target radius, wavelength, and twist prevalence can be obtained. The system accepts either single values representing an average or arrays representing change along the strand. For those with a technical background or who have specific strands that they would like to emulate, the system allows for texture specification via a small set of target strands. These target strands can be created from a 3D modeling tool (Blender~\cite{blender}, Houdini~\cite{houdini}, etc.), the process described in Sections~\ref{sec:strand_modeling} and \ref{sec:simulation}, or any other method; alternatively, they can be obtained from an existing groom. As demonstrated in Figure~\ref{fig:small_input}, the diffusion model does not require a high number of input strands. This makes the approach suitable for those who prefer to spend their time creatively modeling just a few strands representative of the desired texture instead of aiming to create a full groom with procedural methods that typically lack precision.  

\begin{figure}[ht]
  \centering
  \includegraphics[width=0.45\textwidth]{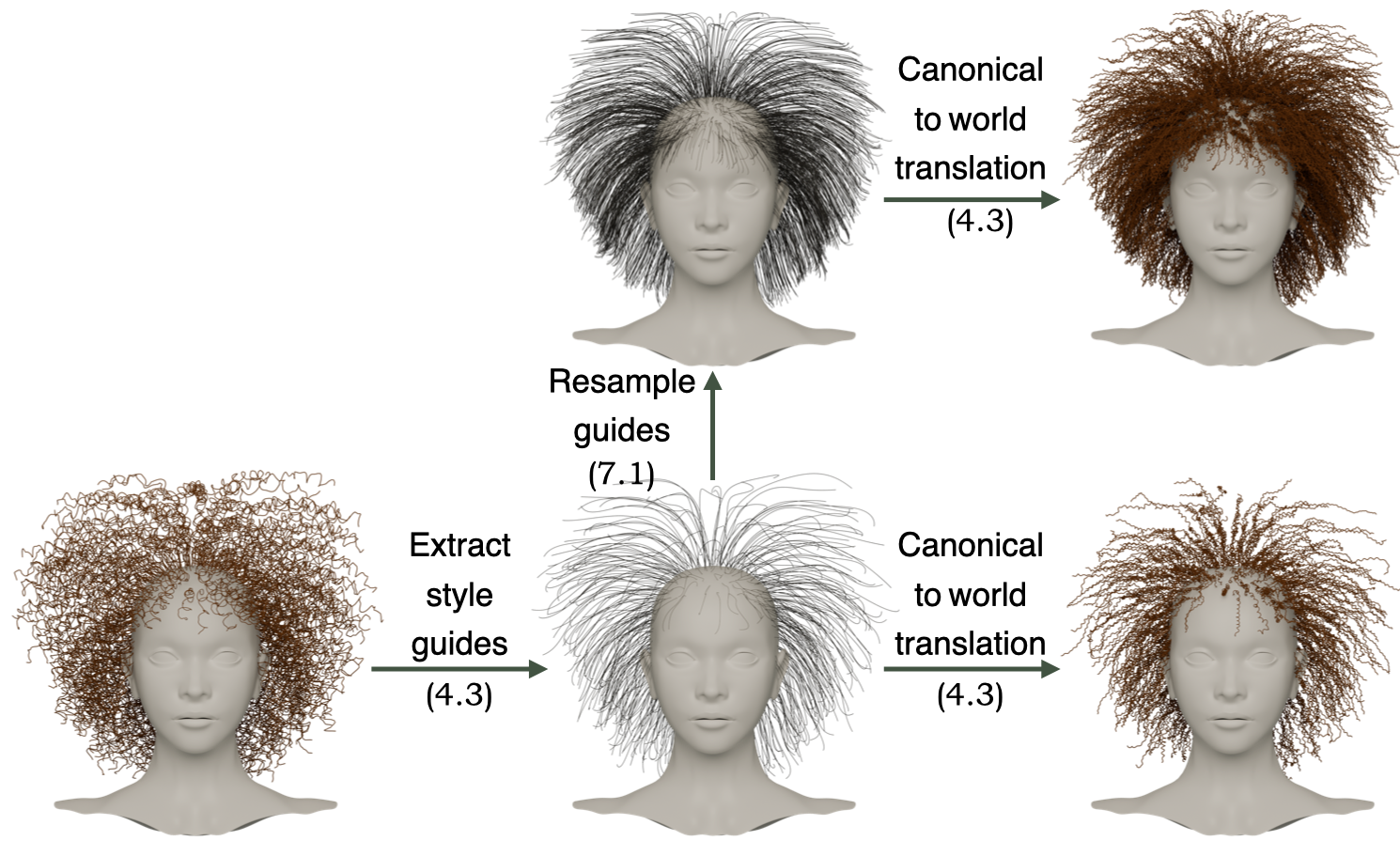}
  \caption{\footnotesize Modifying the texture of an input groom (bottom left) without changing the density of the style guides (bottom middle) can lead to unrealistic results (bottom right). In this specific case where the texture radius decreases, denser style guides (top middle) are required to obtain more realistic results (top right). }
  \Description{An input groom, it's extracted style guides, and result with texture applied (bottom). Resampled style guides and the result of applying textured strands to the resampled style guides (top). }\label{fig:style_guide_resampling}
\end{figure}

\subsection{Style guide resampling}
\label{sec:style_guide_resampling}
It is typically necessary to adjust the style guide density in order to support the new texture. For example, if the input groom has wide curls with each style guide representing a single curl, then the style guides will be sparse and far apart in order to avoid collisions between neighboring curls. Replacing this texture with tighter curls on the same style guides leads to unrealistic gaps between neighboring strands. This can be remedied with a denser set of style guides, as shown in Figure~\ref{fig:style_guide_resampling}. To determine the amount of resampling required, we compare the radii of the input and target textures in the canonical space. The radii of input and/or target strands can be determined via the mean distance of each point from the $z$-axis. For robustness, we remove distance values that fall more than three standard deviations from the average. The square of the ratio of the radii (cross-sectional area for circle-packing) indicates the required increase or decrease in style guide density. Using the new density, new strand roots are sampled across the UV texture of the scalp; then, new style guides are interpolated from the original style guides. Importantly, we forgo including roots where neighboring strands diverge significantly enough to hinder accurate interpolation.

\begin{figure}[ht]
  \centering
  \includegraphics[width=0.45\textwidth]{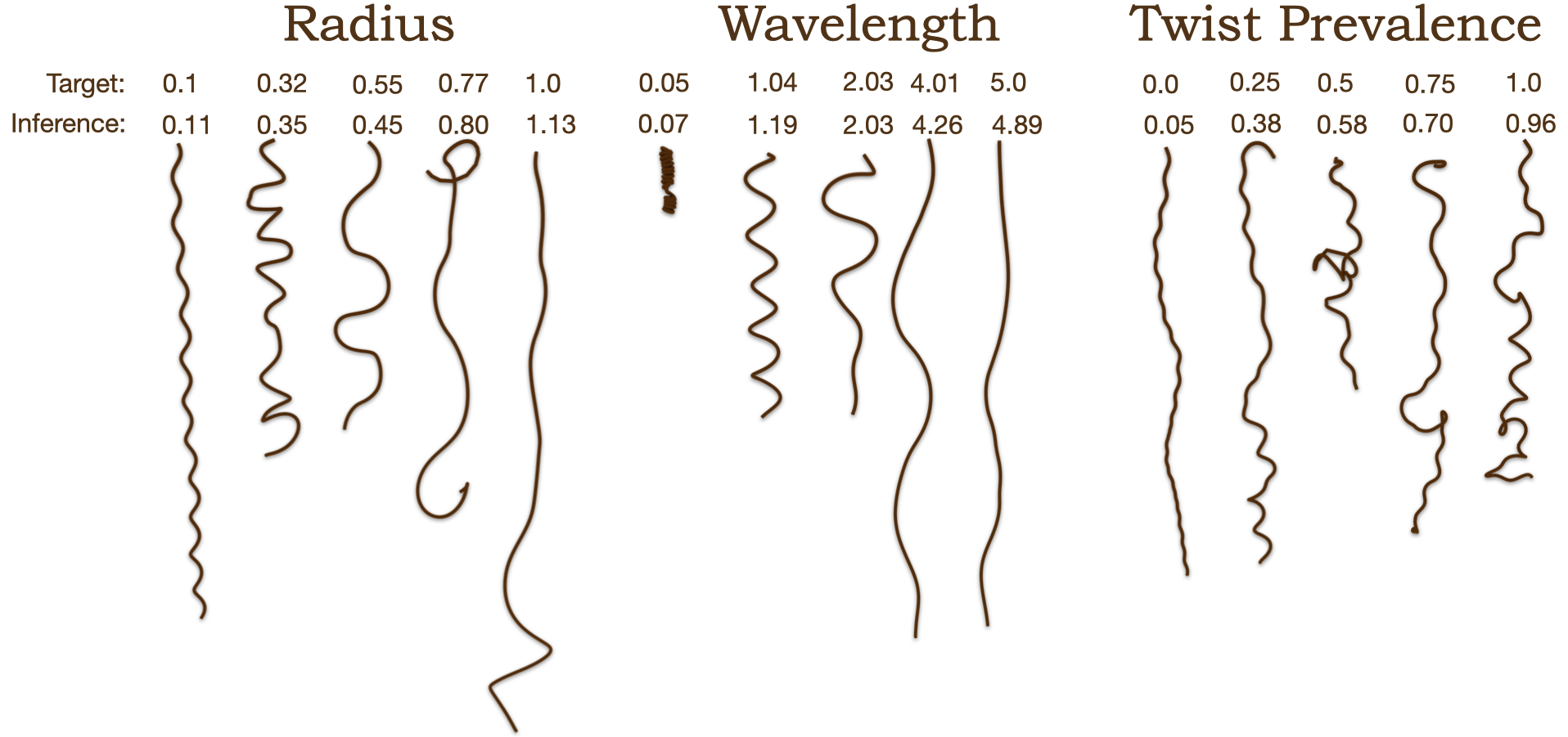}
  \caption{\footnotesize This figure illustrates the efficacy of the guidance mechanism discussed in Section~\ref{sec:diffusion_integration}, when it is used on the pre-trained model from Section~\ref{sec:training}. For each parameter, we increase the specified target value (from left to right) and show the resulting inferenced strand along with its labeled (via Section~\ref{sec:inferencing_parameters}) parameter value.}
  \Description{Strands varying in radius, wavelength, and twist prevalence consistent with model inputs.  }
  \label{fig:supervision}
\end{figure}

\subsection{Strand Synthesis}
\label{sec:diffusion_integration}
When the user specifies a target texture via a small number of strands, the diffusion model is trained on those strands. This takes just a few minutes with a modern GPU and is also accessible via the CPU for those with limited computational resources. After training, the overfit model inferences using only the lengths of the target style guides as guidance. When the user specifies target parameters rather than target strands, we utilize the pre-trained model from Section~\ref{sec:training}. Since the pretrained model has not been overfit to a target texture, a combined guidance mechanism is used to ensure that the output strands match each of the specified parameters in addition to the lengths of the target style guides. Figure~\ref{fig:supervision} illustrates the efficacy of this approach.

\subsection{Adjusting Strand Length}
\label{sec:populating_style_guides}
\begin{figure}[ht]
  \centering
  \includegraphics[width=0.3\textwidth]{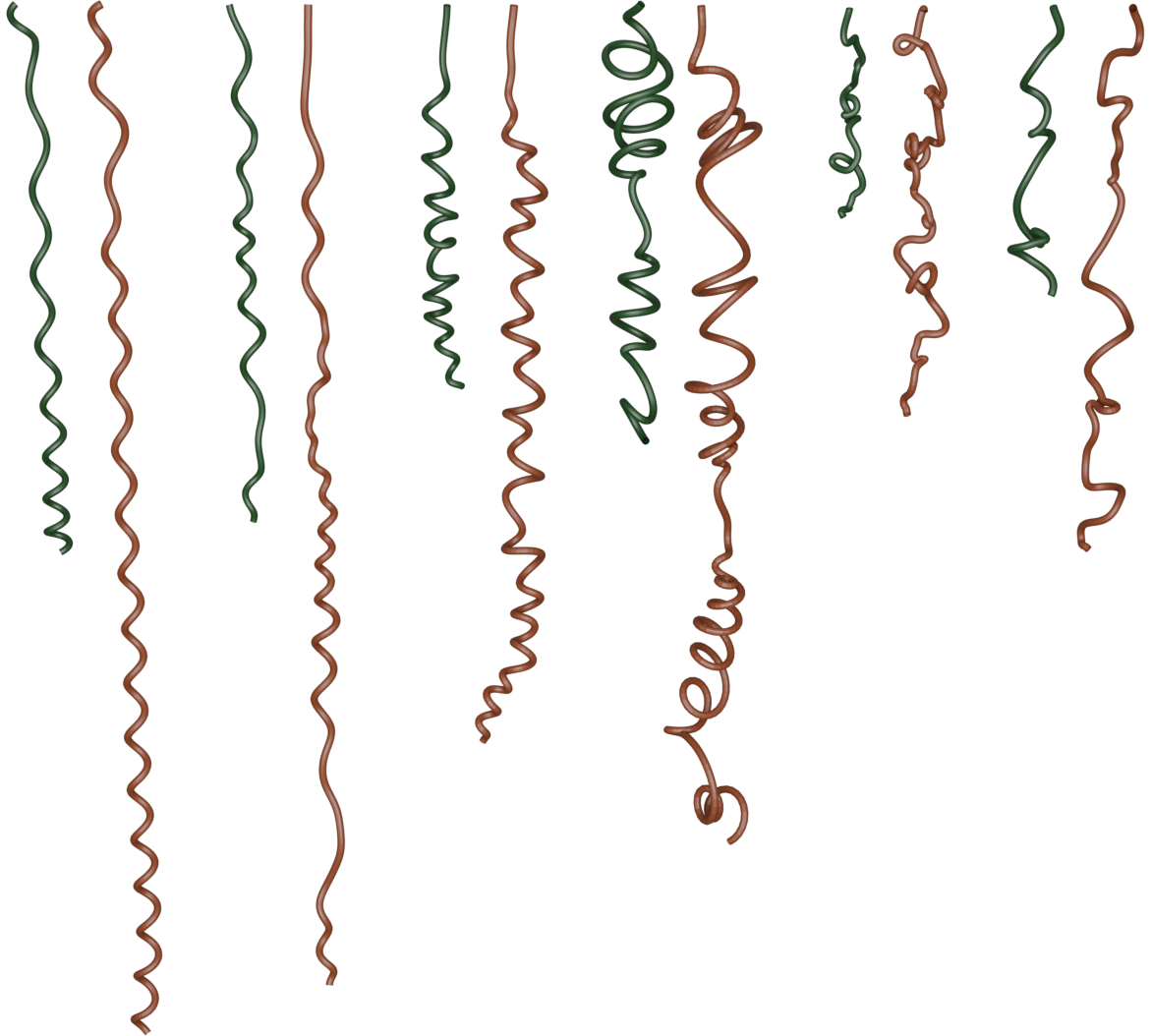}
  \caption{\footnotesize Strands of different textures (green) next to their elongated versions (red), after scaling the speed channel by a factor of 2. }
  \Description{Strands of different textures (green) next to their corresponding strands (red), after scaling the speed channel by a factor of 2. }
\label{fig:resizing_strands}
\end{figure}

Although length is supervised, the lengths of generated strands may differ from the target lengths. In order to preserve the texture while modifying the length, we scale the final predicted speed signal by a constant factor before reconstructing the strand via the method discussed in Section~\ref{sec:strand_reconstruction}. Figure \ref{fig:resizing_strands} illustrates how well this approach preserves strand texture. 

\subsection{Treating Interpenetrations}
After the inferenced strands are lengthened/shortened (if neccessary), they are mapped to the world space style guides following Section~\ref{sec:canonical}. Even though interpolated style guides should not interpenetrate the head, ears, or neck (and are deleted if they do), their world space strands may. In fact, due to changes in texture, world space strands added to original non-interpolated style guides, may also have interpenetrations. We alleviate interpenetrations by rotating style guides radially around their roots to remove penetrations if/when possible. Alternatively, offending hairs could be deleted or simulated. 

\begin{figure}[ht]
  \centering
  \includegraphics[width=.4\textwidth]{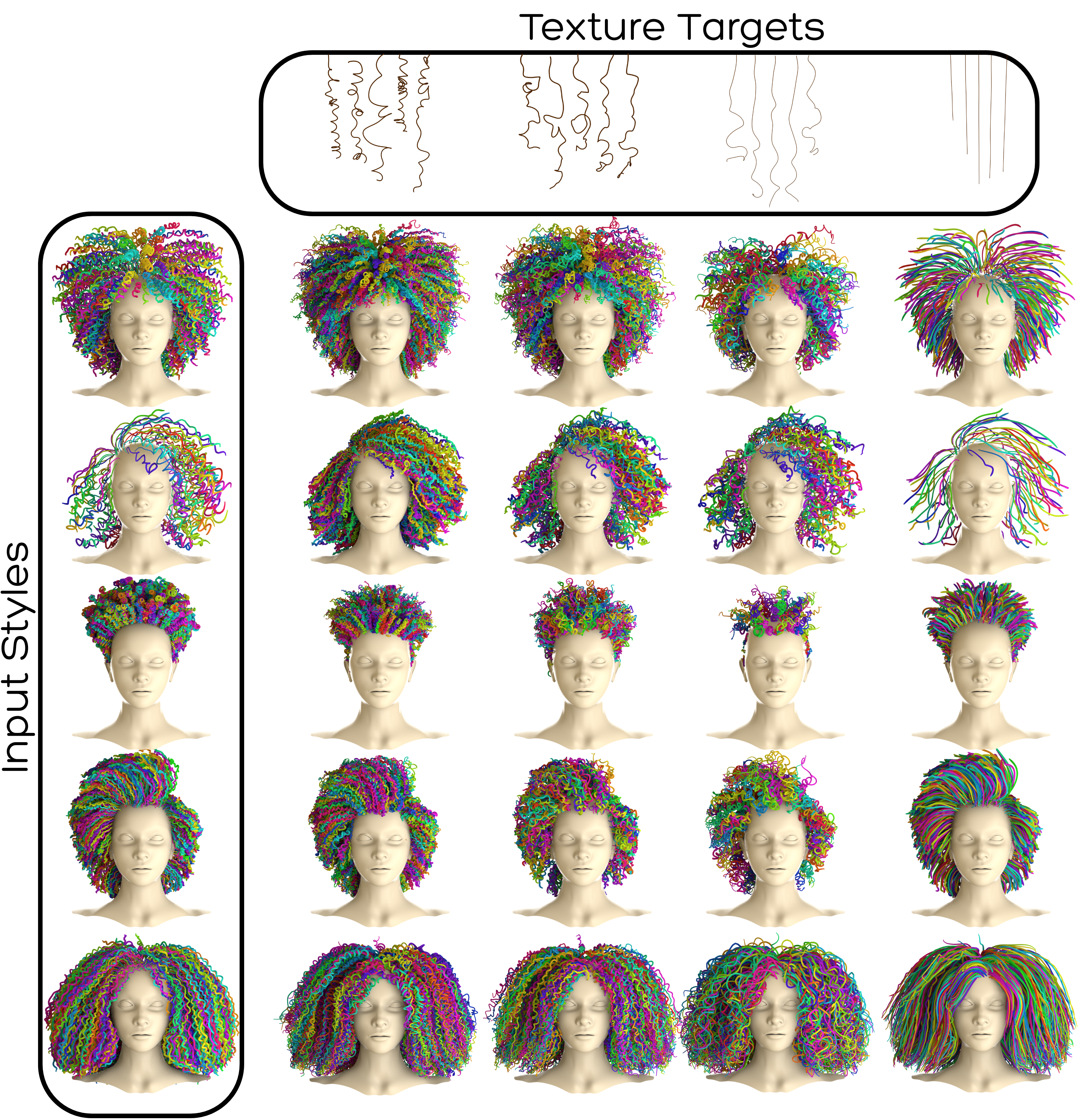}
  \caption{\footnotesize In order to demonstrate disentanglement between style and texture, we illustrate the results obtained by using our method on four disparate styles (left) and four disparate textures (top). See Figure~\ref{fig:results_grid_rendered}.}
  \Description{A set of styles (left) and the corresponding styles generated from our pipeline (bottom). }\label{fig:results_grid}
  \vspace*{\fill}
\end{figure}

\begin{figure}[ht]
  \centering
  \includegraphics[width=.4\textwidth]{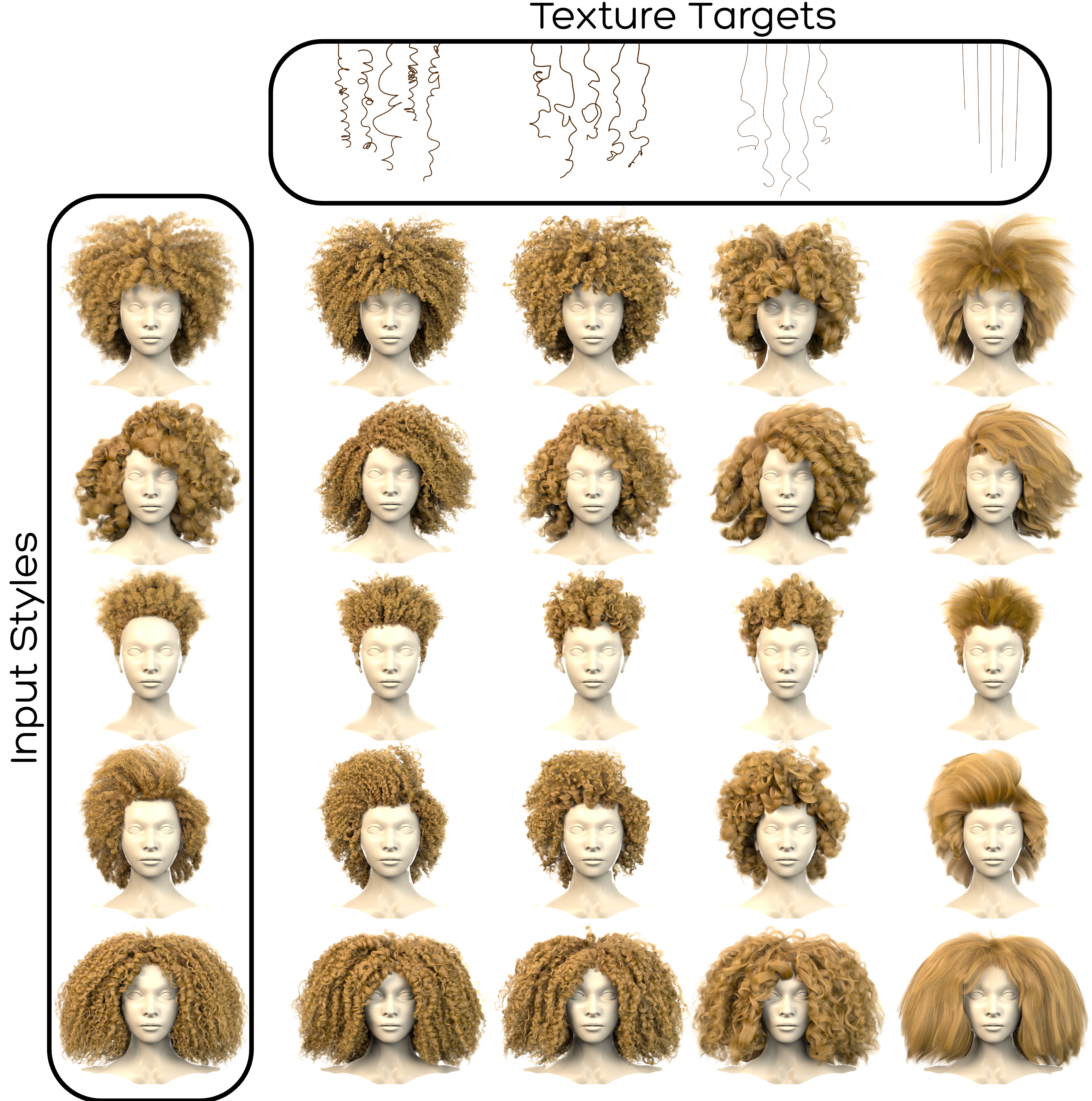}
  \caption{\footnotesize Same as Figure~\ref{fig:results_grid}, except that render hairs (from the guide strands in Figure~\ref{fig:results_grid}) have been added. The render hairs were added in Houdini Apprentice (the non-commercial version of Houdini) with the use of a hair gen node and two clumping nodes to add realism. See Appendix~\ref{sec:render_hairs_houdini}.}
  \Description{A set of styles (left) and the corresponding styles generated from our pipeline (bottom). }\label{fig:results_grid_rendered}
  \vspace*{\fill}
\end{figure}

\newpage
\subsection{Results}\label{sec:results}
\begin{figure}[hb]
  \centering
  \includegraphics[width=0.4\textwidth]{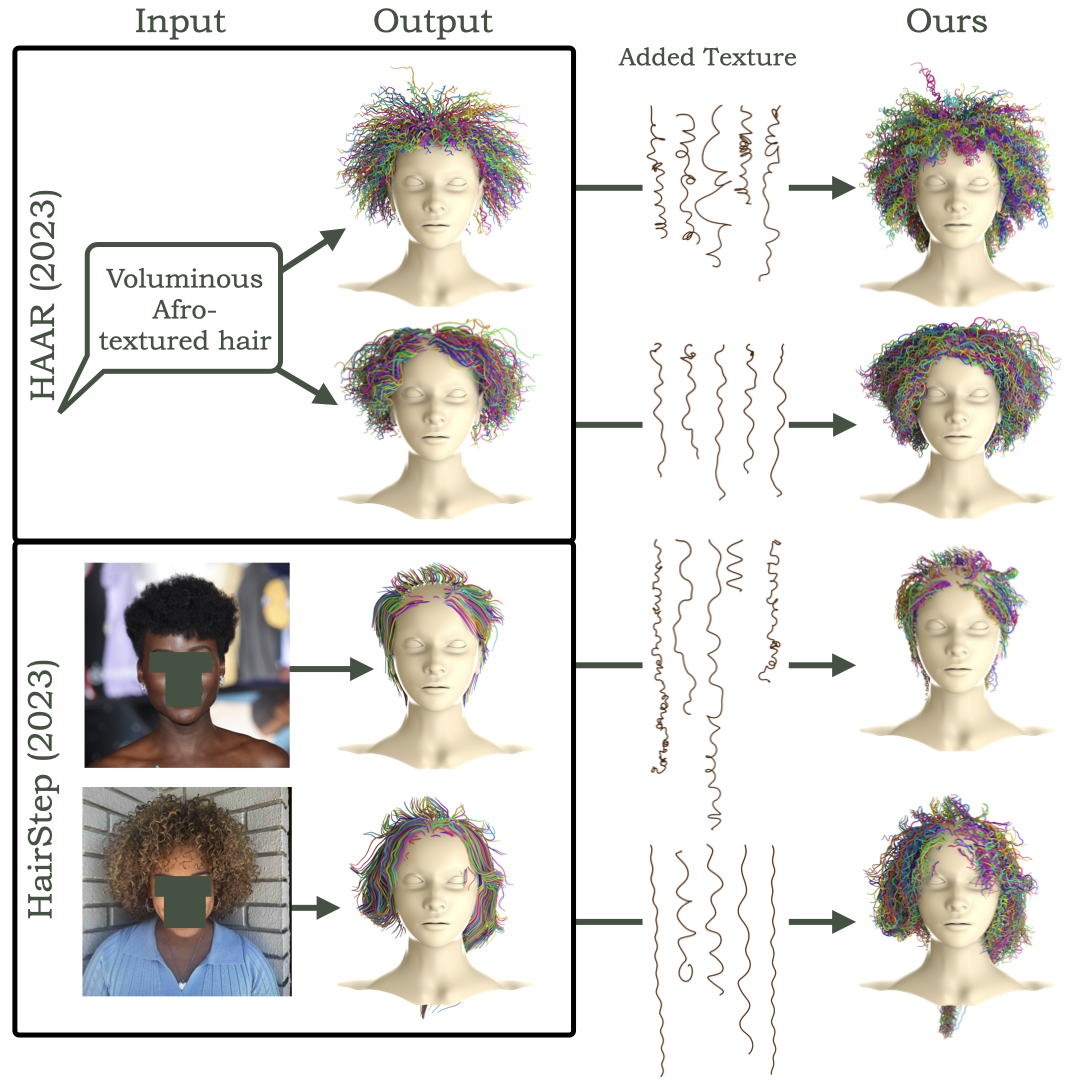}
  \caption{\footnotesize We struggled to generate curly and afro-textured hair using the methods proposed in~\cite{HAAR, hairStep}; However, we were able to improve upon the results by using our method to disentangle geometry and texture and subsequently swap in a better texture. Note that we did not change their generated style (i.e. the centerline geometry), which can also be problematic. See Figure~{\ref{fig:prev_work_rendered}}.}
  \Description{Left column: set of inputs to HAAR and Hairstep and their respective outputs. Middle column: canonical space texture strands we choose to apply. Right column: our results}
  \label{fig:prev_work_guides}
\end{figure}

\begin{figure}[ht]
  \centering
  \includegraphics[width=.4\textwidth]{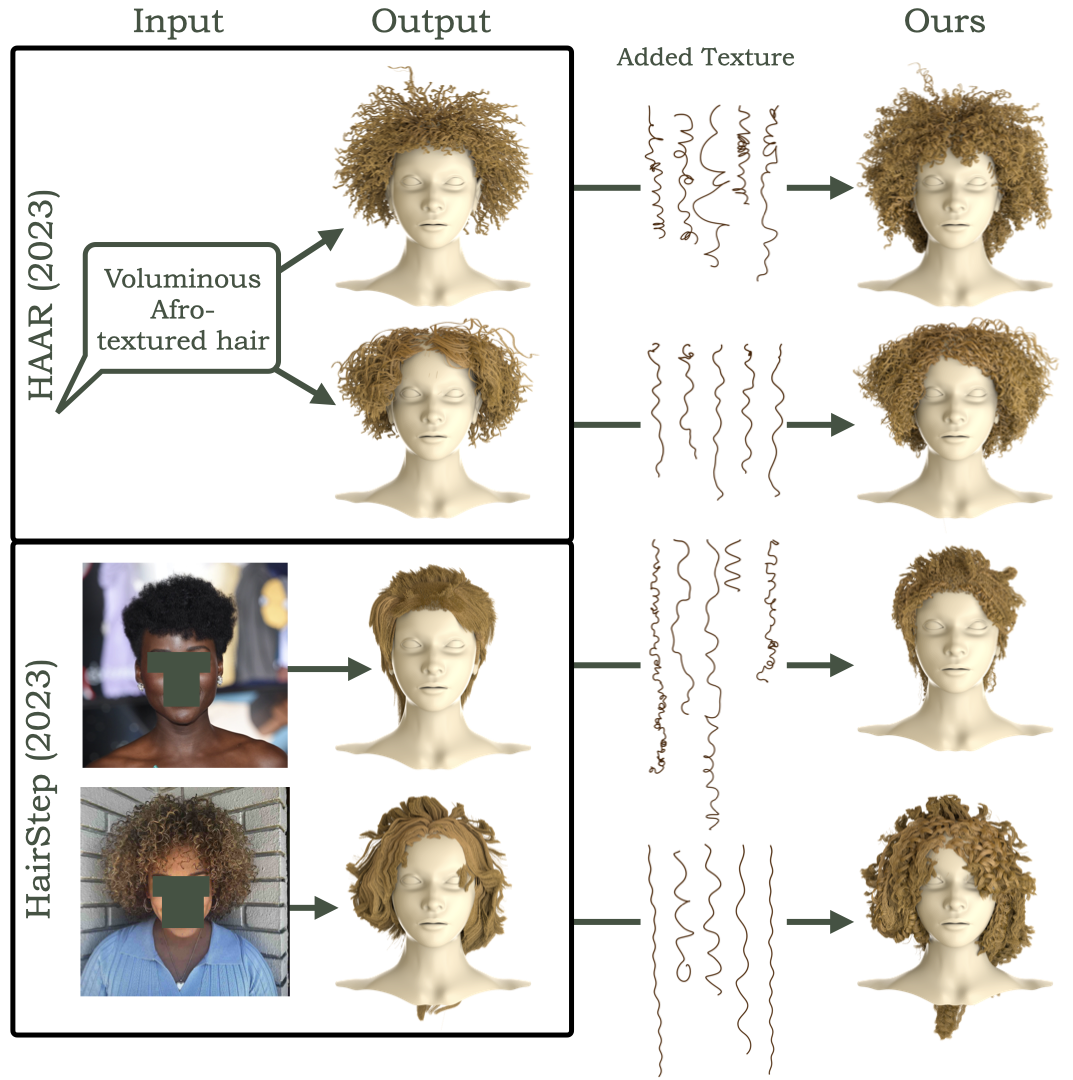}
  \caption{\footnotesize Same as Figure~\ref{fig:prev_work_guides}, except that render hairs (from the guide strands in Figure~\ref{fig:prev_work_guides}) have been added to both the before and after results. The render hairs were added in Houdini Apprentice (the non-commercial version of Houdini) with the use of a hair gen node, a clumping node, and a frizz node to add realism. See Appendix~\ref{sec:render_hairs_houdini}. Note that the same exact process was used on both the before and after results.} 
  \Description{A set of styles (left) and the corresponding styles generated from our pipeline (bottom). }
\label{fig:prev_work_rendered}
  \vspace*{\fill}
\end{figure}

\begin{figure*}[htp!]
  \centering
  \includegraphics[width=.9\textwidth]{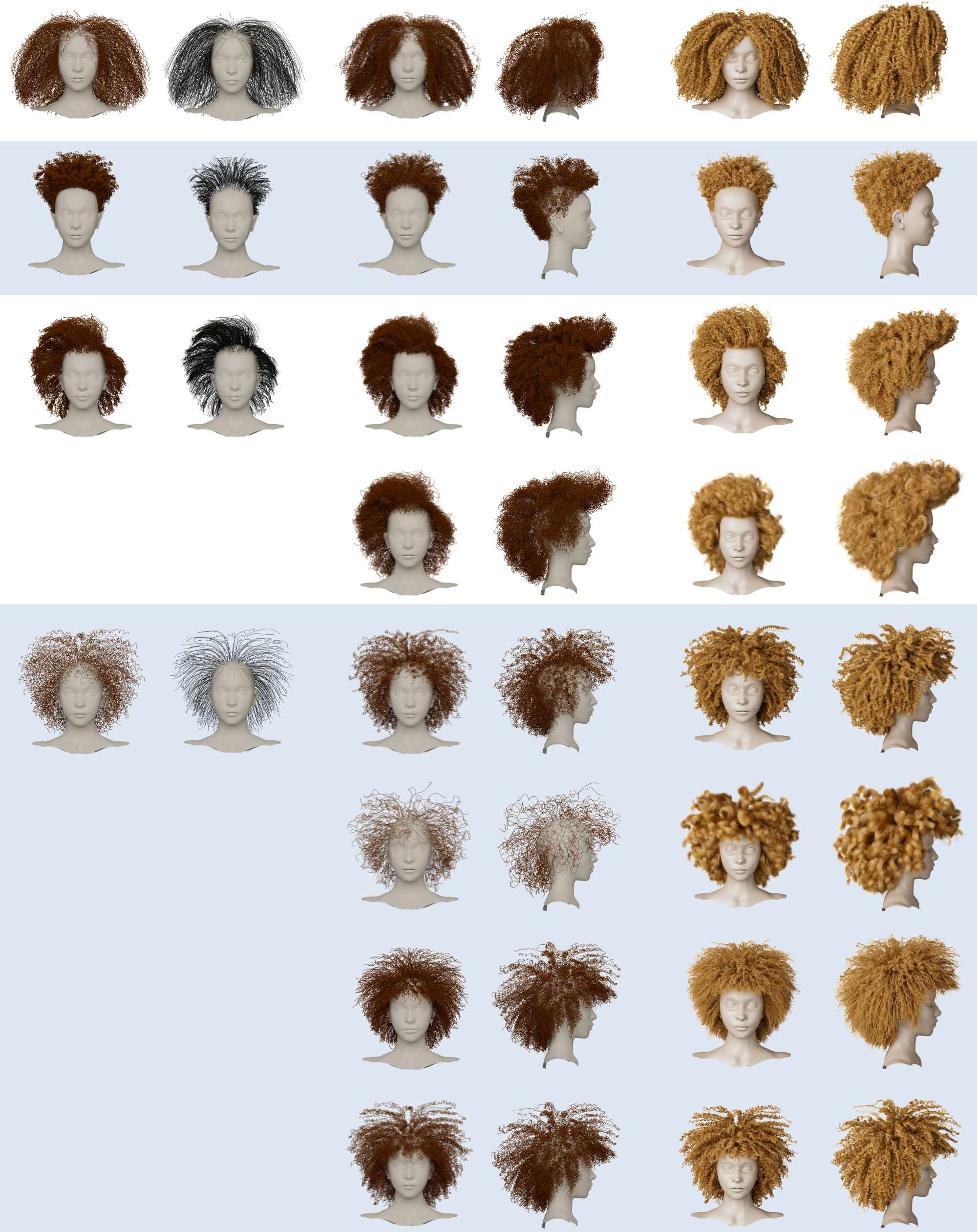}
  \caption{\footnotesize Leftmost two columns: input groom and extracted style guides. Middle two columns: outputs from our texture editing pipeline. Rightmost two columns: final results, using artist created render hairs (from the guide strands in the middle two columns). The diverse results illustrated in the last four rows (last two columns) emphasize the impact of texture variations, especially when noting that all four rows utilize the same input groom and style guides.}
  \Description{Our grooms with render hairs included to demonstrate the output a user would expect from using the guide strands from this pipeline.}\label{fig:rendered_results}
\end{figure*}

Figures~\ref{fig:results_grid} and~\ref{fig:results_grid_rendered} demonstrate the capabilities of our pipeline and highlight disentanglement between style and texture. In Figures~\ref{fig:prev_work_guides} and~\ref{fig:prev_work_rendered}, we demonstrate that our approach can be used to improve failure cases from prior works that struggle to capture curly and afro-textured hair. We show more examples in Figure~\ref{fig:rendered_results} to demonstrate the efficacy of our approach. 

\section{Discussion}
After numerous discussions with artists, scientists, and hair stylists as well as a review of the relevant scientific literature, we settled on an accurate and accessible description of hair texture. Unfortunately, traditional computational approaches are not able to utilize this description. This is likely to be true for a number of interesting phenomena (not just for
 hair), highlighting the importance of generative methods. In order to increase the efficacy of the generative approach, we disentangled texture from geometry and utilized a canonical space. Isolating the phenomena of interest via disentanglement and using a canonical space to reduce the amount of training data and improve the robustness of supervision (for the generative model) are both good practice for any generative approach (not just for hair). The largest impediment to generative modeling consistent with accessible human-driven descriptions is \textit{the intractibility of user supervision} at every step of the generative process; thus, a non-manual labeling scheme is required for tractible supervision. We trained a neural network to assign labels. The architecture of our labeling network relies on classical differential geometry representations and various numerical analysis techniques to increase accuracy and robustness and decrease the dependance on difficult-to-obtain training data.

\section{Limitations and Future Work}
This work does not address ponytails, braids, locs, buns, hats, headphones and other constrained and/or partially occluded hairstyles. Although we address interpenetrations of newly added strands with the head, ears, and neck in a preliminary fashion, we minimize but do not entirely eliminate these artifacts. We leverage only the most basic techniques for adding render hairs and do not believe that this is sufficiently representative of the high variance one would expect in curly and afro-textured hair. In future work, we will explore the complexities of the inter-hair interactions of curly and afro-textured hair beyond the wisp level discussed in this paper. This should lead to more accurate render hairs. Rendered visualizations also demonstrate a need for new research on rendering techniques for hair with higher twist prevalence. Finally, since we worked with artists at every stage of this research, we did obtain positive artist feedback on our proposed approach and modified our method to best serve artists. Given the enthusiasm, we plan to construct a deployable tool (modified by user studies, etc.) as future work. 

\section{Conclusion}\label{sec:conclusions}  
Since media representation impacts personal value, inclusion, acceptance, etc., technical limitations that hinder the diversity of hair types on screen cause both individual and societal harm. This accentuates the importance of our work, which makes the representation of curly and afro-textured hair more accessible. We accomplish this by merging stylist, artist, and science-based classifications of hair texture with powerful generative AI techniques in order to define and generate strands that match a wide variety of textures. Our newly-proposed five-dimensional parameter space is accessible to experts and non-experts alike, and our generative AI pipeline allows for the creation of new strands either directly from texture parameters or from a few example strands. We demonstrated the efficacy of our approach via the process depicted in Figure~\ref{fig:full_pipeline}, which illustrates texture replacement on a pre-existing groom; furthermore, in combination with methods for generating style-guides, entire grooms can be created from scratch. 

\newpage
\bibliographystyle{ACM-Reference-Format}
\bibliography{zotero}

@misc{3DGH,
  title = {{{3DGH}}: {{3D Head Generation}} with {{Composable Hair}} and {{Face}}},
  shorttitle = {{{3DGH}}},
  author = {He, Chengan and Li, Junxuan and Sevastopolsky, Artem and Saito, Shunsuke and Tan, Qingyang and Romero, Javier and Cao, Chen and Rushmeier, Holly and Nam, Giljoo},
  year = 2025,
  month = jun,
  number = {arXiv:2506.20875},
  eprint = {2506.20875},
  primaryclass = {cs},
  publisher = {arXiv},
  doi = {10.48550/arXiv.2506.20875},
  urldate = {2025-08-04},
  archiveprefix = {arXiv}
}

@misc{AG3D,
  title = {{{AG3D}}: {{Learning}} to {{Generate 3D Avatars}} from {{2D Image Collections}}},
  shorttitle = {{{AG3D}}},
  author = {Dong, Zijian and Chen, Xu and Yang, Jinlong and Black, Michael J. and Hilliges, Otmar and Geiger, Andreas},
  year = 2023,
  month = may,
  number = {arXiv:2305.02312},
  eprint = {2305.02312},
  primaryclass = {cs},
  publisher = {arXiv},
  doi = {10.48550/arXiv.2305.02312},
  urldate = {2025-06-05},
  archiveprefix = {arXiv}
}

@article{ammosovGeneralizedMultiscaleFinite2024,
  title = {Generalized Multiscale Finite Element Method for a Nonlinear Elastic Strain-Limiting {{Cosserat}} Model},
  author = {Ammosov, Dmitry and Mai, Tina and Galvis, Juan},
  year = 2024,
  month = dec,
  journal = {Journal of Computational Physics},
  volume = {519},
  pages = {113428},
  issn = {0021-9991},
  doi = {10.1016/j.jcp.2024.113428},
  urldate = {2025-06-02}
}

@article{ANIME-Rod,
  title = {{{ANIME-Rod}}: {{Adjustable Nonlinear Isotropic Materials}} for {{Elastic Rods}}},
  shorttitle = {{{ANIME-Rod}}},
  author = {Chen, Huanyu and Wen, Jiahao and Barbi{\v c}, Jernej},
  year = 2025,
  month = jul,
  journal = {ACM Trans. Graph.},
  volume = {44},
  number = {4},
  pages = {87:1--87:23},
  issn = {0730-0301},
  doi = {10.1145/3731208},
  urldate = {2025-08-04}
}

@article{Anjyo1992,
  title = {A Simple Method for Extracting the Natural Beauty of Hair},
  author = {Anjyo, Ken-ichi and Usami, Yoshiaki and Kurihara, Tsuneya},
  year = 1992,
  month = jul,
  journal = {ACM SIGGRAPH Computer Graphics},
  volume = {26},
  number = {2},
  pages = {111--120},
  issn = {0097-8930},
  doi = {10.1145/142920.134021},
  urldate = {2025-08-21},
  langid = {english}
}

@misc{augMassSpring,
  title = {Augmented {{Mass-Spring}} Model for {{Real-Time Dense Hair Simulation}}},
  shorttitle = {{{augMassSpring}}},
  author = {Herrera, Jorge Alejandro Amador and Zhou, Yi and Sun, Xin and Shu, Zhixin and He, Chengan and Pirk, S{\"o}ren and Michels, Dominik L.},
  year = 2024,
  month = dec,
  number = {arXiv:2412.17144},
  eprint = {2412.17144},
  primaryclass = {cs},
  publisher = {arXiv},
  doi = {10.48550/arXiv.2412.17144},
  urldate = {2025-06-02},
  archiveprefix = {arXiv}
}

@article{B1988,
  title = {Human Hair Form. {{Morphology}} Revealed by Light and Scanning Electron Microscopy and Computer Aided Three-Dimensional Reconstruction},
  author = {B, Lindel{\"o}f and B, Forslind and Ma, Hedblad and U, Kaveus},
  year = 1988,
  month = sep,
  journal = {Archives of dermatology},
  volume = {124},
  number = {9},
  publisher = {Arch Dermatol},
  issn = {0003-987X},
  urldate = {2025-06-02},
  langid = {english},
  pmid = {3415278}
}

@misc{Bahdanau2016,
  title = {Neural {{Machine Translation}} by {{Jointly Learning}} to {{Align}} and {{Translate}}},
  author = {Bahdanau, Dzmitry and Cho, Kyunghyun and Bengio, Yoshua},
  year = 2016,
  month = may,
  number = {arXiv:1409.0473},
  eprint = {1409.0473},
  primaryclass = {cs},
  publisher = {arXiv},
  doi = {10.48550/arXiv.1409.0473},
  urldate = {2026-05-05},
  archiveprefix = {arXiv}
}

@inproceedings{Bergou2008,
  title = {Discrete Elastic Rods},
  booktitle = {{{ACM SIGGRAPH}} 2008 Papers},
  author = {Bergou, Mikl{\'o}s and Wardetzky, Max and Robinson, Stephen and Audoly, Basile and Grinspun, Eitan},
  year = 2008,
  month = aug,
  series = {{{SIGGRAPH}} '08},
  pages = {1--12},
  publisher = {Association for Computing Machinery},
  address = {New York, NY, USA},
  doi = {10.1145/1399504.1360662},
  urldate = {2025-06-06},
  isbn = {978-1-4503-0112-1}
}

@article{Bernstein1927,
  title = {Racial and Sexual Differences in Hair Weight},
  author = {Bernstein, Maurice and Robertson, Sylvan},
  year = 1927,
  journal = {American Journal of Physical Anthropology},
  volume = {10},
  pages = {379--385}
}

@article{bertailsSuperhelicesPredictingDynamics2006,
  title = {Super-Helices for Predicting the Dynamics of Natural Hair},
  author = {Bertails, Florence and Audoly, Basile and Cani, Marie-Paule and Querleux, Bernard and Leroy, Fr{\'e}d{\'e}ric and L{\'e}v{\^e}que, Jean-Luc},
  year = 2006,
  month = jul,
  journal = {ACM Transactions on Graphics},
  volume = {25},
  number = {3},
  pages = {1180--1187},
  issn = {0730-0301, 1557-7368},
  doi = {10.1145/1141911.1142012},
  urldate = {2025-06-02},
  langid = {english}
}

@misc{BeyondtheCurl,
  title = {Beyond the {{Curl}}: {{Unraveling}} the {{Diversity}} of {{Hair}}},
  shorttitle = {Beyond the {{Curl}}},
  author = {Media, Scientific American Custom},
  journal = {Scientific American},
  urldate = {2026-04-24},
  howpublished = {https://www.scientificamerican.com/custom-media/loreal/beyond-the-curl-unraveling-the-diversity-of-hair/},
  langid = {english}
}

@book{bezier,
  title = {A {{First Course}} in {{Applied Mathematics}}},
  author = {Rebaza, Jorge},
  year = 2012,
  month = apr,
  publisher = {John Wiley \& Sons},
  googlebooks = {lFwXglfyoIQC},
  isbn = {978-1-118-27715-7},
  langid = {english}
}

@misc{bigHairVogue,
  title = {Hair: {{Why}} Representation and Beauty Culture Matter},
  shorttitle = {{{bigHairVogue}}},
  author = {Alapati, Katlyn Sofaea Alo},
  year = 2022,
  month = may,
  journal = {The San Francisco Chronicle},
  urldate = {2025-06-02},
  howpublished = {https://www.sfchronicle.com/projects/2022/hair/representation/},
  langid = {english}
}

@article{Bishop1975,
  title = {There Is {{More}} than {{One Way}} to {{Frame}} a {{Curve}}},
  author = {Bishop, Richard L.},
  year = 1975,
  journal = {The American Mathematical Monthly},
  volume = {82},
  number = {3},
  eprint = {2319846},
  eprinttype = {jstor},
  pages = {246--251},
  publisher = {[Taylor \& Francis, Ltd., Mathematical Association of America]},
  issn = {0002-9890},
  doi = {10.2307/2319846},
  urldate = {2025-08-18}
}

@article{blackHairPerception,
  title = {How {{Media Influence}} about {{Hair Texture Impacts Internalized Racial Oppression}} and {{Why The Crown Act Simultaneously Promotes Necessary Change}} and {{Yet Familiar Defeat}}},
  shorttitle = {{{blackHairPerception}}},
  author = {LaMar, Kristy L and Rolle, Helen N},
  year = 2022,
  journal = {Journal of Psychology},
  volume = {10},
  number = {2},
  pages = {1--8}
}

@techreport{blackWomenInFilm,
  title = {Representations of {{Black Women}} in {{Hollywood}}},
  shorttitle = {{{blackWomenInFilm}}},
  author = {{Geena Davis Institute on Gender in Media}},
  year = 2021,
  institution = {Mount Saint Mary's University,}
}

@misc{blender,
  title = {Blender},
  shorttitle = {Blender},
  author = {{Blender Online Community}},
  year = 2026,
  address = {Amsterdam},
  howpublished = {Blender Foundation}
}

@misc{blowingHair,
  title = {Automatic {{Animation}} of {{Hair Blowing}} in {{Still Portrait Photos}}},
  shorttitle = {{{blowingHair}}},
  author = {Xiao, Wenpeng and Liu, Wentao and Wang, Yitong and Ghanem, Bernard and Li, Bing},
  year = 2023,
  month = sep,
  number = {arXiv:2309.14207},
  eprint = {2309.14207},
  primaryclass = {cs},
  publisher = {arXiv},
  doi = {10.48550/arXiv.2309.14207},
  urldate = {2025-06-05},
  archiveprefix = {arXiv}
}

@book{Blume-Peytavi2008,
  title = {Hair Growth and Disorders},
  author = {{Blume-Peytavi}, Ulrike},
  year = 2008,
  publisher = {Springer},
  address = {Berlin},
  isbn = {978-3-540-46911-7},
  langid = {english},
  lccn = {616.546}
}

@book{Brunner1974,
  title = {The {{Identification}} of {{Mammalian Hair}}},
  author = {Brunner, Hans and Coman, Brian J.},
  year = 1974,
  publisher = {Inkata Press},
  googlebooks = {CKiPQgAACAAJ},
  isbn = {978-0-909605-01-8},
  langid = {english}
}

@incollection{CatmullRom,
  title = {A {{CLASS OF LOCAL INTERPOLATING SPLINES}}},
  booktitle = {Computer {{Aided Geometric Design}}},
  author = {Catmull, Edwin and Rom, Raphael},
  year = 1974,
  month = jan,
  pages = {317--326},
  publisher = {Academic Press},
  doi = {10.1016/B978-0-12-079050-0.50020-5},
  urldate = {2026-04-16},
  langid = {american}
}

@article{chemicalBondsTypes,
  title = {Chemical Bonds and Hair Behaviour- {{A}} Review},
  shorttitle = {{{chemicalBondsTypes}}},
  author = {Breakspear, S. and N{\"o}cker, B. and Popescu, C.},
  year = 2024,
  journal = {International Journal of Cosmetic Science},
  volume = {46},
  number = {5},
  pages = {806--814},
  doi = {10.1111/ics.12967}
}

@misc{Chen2023,
  title = {On the {{Importance}} of {{Noise Scheduling}} for {{Diffusion Models}}},
  author = {Chen, Ting},
  year = 2023,
  month = may,
  number = {arXiv:2301.10972},
  eprint = {2301.10972},
  primaryclass = {cs},
  publisher = {arXiv},
  doi = {10.48550/arXiv.2301.10972},
  urldate = {2025-09-29},
  archiveprefix = {arXiv}
}

@misc{Chung2024,
  title = {Diffusion {{Posterior Sampling}} for {{General Noisy Inverse Problems}}},
  author = {Chung, Hyungjin and Kim, Jeongsol and Mccann, Michael T. and Klasky, Marc L. and Ye, Jong Chul},
  year = 2024,
  month = may,
  number = {arXiv:2209.14687},
  eprint = {2209.14687},
  primaryclass = {stat},
  publisher = {arXiv},
  doi = {10.48550/arXiv.2209.14687},
  urldate = {2025-09-18},
  archiveprefix = {arXiv}
}

@misc{clip,
  title = {Learning {{Transferable Visual Models From Natural Language Supervision}}},
  shorttitle = {Clip},
  author = {Radford, Alec and Kim, Jong Wook and Hallacy, Chris and Ramesh, Aditya and Goh, Gabriel and Agarwal, Sandhini and Sastry, Girish and Askell, Amanda and Mishkin, Pamela and Clark, Jack and Krueger, Gretchen and Sutskever, Ilya},
  year = 2021,
  month = feb,
  number = {arXiv:2103.00020},
  eprint = {2103.00020},
  primaryclass = {cs},
  publisher = {arXiv},
  doi = {10.48550/arXiv.2103.00020},
  urldate = {2025-06-10},
  archiveprefix = {arXiv}
}

@article{cloeteSystemsApproachHuman2019,
  title = {Systems {{Approach}} to {{Human Hair Fibers}}: {{Interdependence Between Physical}}, {{Mechanical}}, {{Biochemical}} and {{Geometric Properties}} of {{Natural Healthy Hair}}},
  shorttitle = {{{structuralTaxonomy}}},
  author = {Cloete, Elsabe and Khumalo, Nonhlanhla P. and Van Wyk, Jennifer C. and Ngoepe, Malebogo N.},
  year = 2019,
  journal = {Frontiers in Physiology},
  volume = {Volume 10 - 2019},
  issn = {1664-042X},
  doi = {10.3389/fphys.2019.00112}
}

@inproceedings{CodecAvatarStudio,
  title = {Codec {{Avatar Studio}}: {{Paired Human Captures}} for {{Complete}}, {{Driveable}}, and {{Generalizable Avatars}}},
  shorttitle = {Codec {{Avatar Studio}}},
  booktitle = {The {{Thirty-eight Conference}} on {{Neural Information Processing Systems Datasets}} and {{Benchmarks Track}}},
  author = {Martinez, Julieta and Kim, Emily and Romero, Javier and Bagautdinov, Timur and Saito, Shunsuke and Yu, Shoou-I. and Anderson, Stuart and Zollh{\"o}fer, Michael and Wang, Te-Li and Bai, Shaojie and Li, Chenghui and Wei, Shih-En and Joshi, Rohan and Borsos, Wyatt and Simon, Tomas and Saragih, Jason and Theodosis, Paul and Greene, Alexander and Josyula, Anjani and Maeta, Silvio Mano and Jewett, Andrew I. and Venshtain, Simion and Heilman, Christopher and Chen, Yueh-Tung and Fu, Sidi and Elshaer, Mohamed Ezzeldin A. and Du, Tingfang and Wu, Longhua and Chen, Shen-Chi and Kang, Kai and Wu, Michael and Emad, Youssef and Longay, Steven and Brewer, Ashley and Shah, Hitesh and Booth, James and Koska, Taylor and Haidle, Kayla and Andromalos, Matthew and Hsu, Joanna Ching-Hui and Dauer, Thomas and Selednik, Peter and Godisart, Tim and Ardisson, Scott and Cipperly, Matthew and Humberston, Ben and Farr, Lon and Hansen, Bob and Guo, Peihong and Braun, Dave and Krenn, Steven and Wen, He and Evans, Lucas and Fadeeva, Natalia and Stewart, Matthew and Schwartz, Gabriel and Gupta, Divam and Moon, Gyeongsik and Guo, Kaiwen and Dong, Yuan and Xu, Yichen and Shiratori, Takaaki and Nino, Fabian Andres Prada and Pires, Bernardo R. and Peng, Bo and Buffalini, Julia and Trimble, Autumn and McPhail, Kevyn Alex Anthony and Schoeller, Melissa Robinson and Sheikh, Yaser},
  year = 2024,
  month = nov,
  urldate = {2025-06-05},
  langid = {english}
}

@article{crossSectionRace,
  title = {Variation in Human Hair Ultrastructure among Three Biogeographic Populations},
  shorttitle = {{{crossSectionRace}}},
  author = {Koch, Sandra L. and Shriver, Mark D. and Jablonski, Nina G.},
  year = 2019,
  month = jan,
  journal = {Journal of Structural Biology},
  volume = {205},
  number = {1},
  pages = {60--66},
  issn = {1047-8477},
  doi = {10.1016/j.jsb.2018.11.008},
  urldate = {2025-08-19}
}

@article{CT2Hair,
  title = {{{CT2Hair}}: {{High-Fidelity 3D Hair Modeling}} Using {{Computed Tomography}}},
  shorttitle = {{{CT2Hair}}},
  author = {Shen, Yuefan and Saito, Shunsuke and Wang, Ziyan and Maury, Olivier and Wu, Chenglei and Hodgins, Jessica and Zheng, Youyi and Nam, Giljoo},
  year = 2023,
  journal = {ACM Transactions on Graphics},
  volume = {42},
  number = {4},
  pages = {1--13},
  publisher = {ACM New York, NY, USA}
}

@book{culturalHistoryOfHair,
  title = {A {{Cultural History}} of {{Hair}} in {{Antiquity}}},
  shorttitle = {{{culturalHistoryOfHair}}},
  author = {Harlow, Mary and {Biddle-Perry}, Geraldine and Snook, Edith and Milliken, Roberta and Powell, Margaret K. and Roach, Joseph R. and Heaton, Sarah},
  year = 2022,
  publisher = {Bloomsbury Academic},
  googlebooks = {vSiFEAAAQBAJ},
  isbn = {978-1-350-28532-3},
  langid = {english}
}

@inproceedings{Curly-Cue,
  title = {Curly-{{Cue}}: {{Geometric Methods}} for {{Highly Coiled Hair}}},
  shorttitle = {Curly-{{Cue}}},
  booktitle = {{{SIGGRAPH Asia}} 2024 {{Conference Papers}}},
  author = {Wu, Haomiao and Shi, Alvin and Darke, A.M. and Kim, Theodore},
  year = 2024,
  month = dec,
  pages = {1--11},
  publisher = {ACM},
  address = {Tokyo Japan},
  doi = {10.1145/3680528.3687641},
  urldate = {2025-06-02},
  isbn = {979-8-4007-1131-2},
  langid = {english}
}

@inproceedings{curvatureAnalysis,
  title = {Curvature {{Analysis}} of {{Sculpted Hair Meshes}} for {{Hair Guides Generation}}},
  shorttitle = {{{curvatureAnalysis}}},
  booktitle = {Advances in {{Computer Graphics}}},
  author = {Pellegrin, Florian and Beauchamp, Andre and Paquette, Eric},
  editor = {{Magnenat-Thalmann}, Nadia and Interrante, Victoria and Thalmann, Daniel and Papagiannakis, George and Sheng, Bin and Kim, Jinman and Gavrilova, Marina},
  year = 2021,
  pages = {378--397},
  publisher = {Springer International Publishing},
  address = {Cham},
  doi = {10.1007/978-3-030-89029-2_30},
  isbn = {978-3-030-89029-2},
  langid = {english}
}

@article{databaseCombinations,
  title = {Single-{{View Hair Modeling Using A Hairstyle Database}}},
  shorttitle = {{{databaseCombinations}}},
  author = {Hu, Liwen and Ma, Chongyang and Luo, Linjie and Li, Hao},
  year = 2015,
  month = jul,
  journal = {ACM Transactions on Graphics (Proceedings SIGGRAPH 2015)},
  volume = {34},
  number = {4},
  publisher = {ACM}
}

@inproceedings{Daviet2023,
  title = {Interactive {{Hair Simulation}} on the {{GPU}} Using {{ADMM}}},
  booktitle = {Special {{Interest Group}} on {{Computer Graphics}} and {{Interactive Techniques Conference Conference Proceedings}}},
  author = {Daviet, Gilles},
  year = 2023,
  month = jul,
  pages = {1--11},
  publisher = {ACM},
  address = {Los Angeles CA USA},
  doi = {10.1145/3588432.3591551},
  urldate = {2025-06-02},
  isbn = {979-8-4007-0159-7},
  langid = {english}
}

@inproceedings{DCT,
  title = {Doubly {{Hierarchical Geometric Representations}} for {{Strand-based Human Hairstyle Generation}}},
  shorttitle = {{{DCT}}},
  booktitle = {Advances in {{Neural Information Processing Systems}}},
  author = {Chen, Yunlu and Carrasco, Francisco Vicente and H{\"a}ne, Christian and Nam, Giljoo and Bazin, Jean-Charles and {De la Torre}, Fernando},
  editor = {Globerson, A. and Mackey, L. and Belgrave, D. and Fan, A. and Paquet, U. and Tomczak, J. and Zhang, C.},
  year = 2024,
  volume = {37},
  pages = {89728--89751},
  publisher = {Curran Associates, Inc.},
  address = {Pittsburgh, PA, USA}
}

@article{DeformableNerfs,
  title = {Deformable {{Neural Radiance Fields}}},
  shorttitle = {{{DeformableNerfs}}},
  author = {Park, Keunhong and Sinha, Utkarsh and Barron, Jonathan T. and Bouaziz, Sofien and Goldman, Dan B. and Seitz, Steven M. and {Martin-Brualla}, Ricardo},
  year = 2020,
  journal = {CoRR},
  volume = {abs/2011.12948},
  eprint = {2011.12948},
  archiveprefix = {arXiv}
}

@inproceedings{DiffLocks,
  title = {{{DiffLocks}}: {{Generating 3D Hair}} from a {{Single Image}} Using {{Diffusion Models}}},
  shorttitle = {{{DiffLocks}}},
  booktitle = {Proceedings {{IEEE Conf}}. on {{Computer Vision}} and {{Pattern Recognition}} ({{CVPR}})},
  author = {Rosu, Radu Alexandru and Wu, Keyu and Feng, Yao and Zheng, Youyi and Black, Michael J.},
  year = 2025,
  publisher = {CVPR 2025},
  address = {Stuttgart Germany}
}

@misc{diffusion,
  title = {Deep {{Unsupervised Learning}} Using {{Nonequilibrium Thermodynamics}}},
  shorttitle = {Diffusion},
  author = {{Sohl-Dickstein}, Jascha and Weiss, Eric A. and Maheswaranathan, Niru and Ganguli, Surya},
  year = 2015,
  month = nov,
  number = {arXiv:1503.03585},
  eprint = {1503.03585},
  primaryclass = {cs},
  publisher = {arXiv},
  doi = {10.48550/arXiv.1503.03585},
  urldate = {2025-06-02},
  archiveprefix = {arXiv}
}

@misc{DPM-Solver++,
  title = {{{DPM-Solver}}++: {{Fast Solver}} for {{Guided Sampling}} of {{Diffusion Probabilistic Models}}},
  shorttitle = {{{DPM-Solver}}++},
  author = {Lu, Cheng and Zhou, Yuhao and Bao, Fan and Chen, Jianfei and Li, Chongxuan and Zhu, Jun},
  year = 2025,
  month = may,
  eprint = {2211.01095},
  primaryclass = {cs},
  doi = {10.1007/s11633-025-1562-4},
  urldate = {2025-06-10},
  archiveprefix = {arXiv}
}

@inproceedings{Dr.Hair,
  title = {Dr.{{Hair}}: {{Reconstructing Scalp-Connected Hair Strands}} without {{Pre-Training}} via {{Differentiable Rendering}} of {{Line Segments}}},
  shorttitle = {Dr.{{Hair}}},
  booktitle = {Proceedings of the {{IEEE}}/{{CVF Conference}} on {{Computer Vision}} and {{Pattern Recognition}}},
  author = {Takimoto, Yusuke and Takehara, Hikari and Sato, Hiroyuki and Zhu, Zihao and Zheng, Bo},
  year = 2024,
  pages = {20601--20611},
  urldate = {2025-06-05},
  langid = {english}
}

@misc{dynamicGaussians,
  title = {Dynamic {{3D Gaussians}}: {{Tracking}} by {{Persistent Dynamic View Synthesis}}},
  shorttitle = {{{dynamicGaussians}}},
  author = {Luiten, Jonathon and Kopanas, Georgios and Leibe, Bastian and Ramanan, Deva},
  year = 2023
}

@article{DynamicReconstruction,
  title = {Dynamic Hair Modeling from Monocular Videos Using Deep Neural Networks},
  shorttitle = {{{DynamicReconstruction}}},
  author = {Yang, Lingchen and Shi, Zefeng and Zheng, Youyi and Zhou, Kun},
  year = 2019,
  journal = {ACM Transactions on Graphics (TOG)},
  volume = {38},
  number = {6},
  pages = {1--12}
}

@article{Fei2019,
  title = {A Multi-Scale Model for Coupling Strands with Shear-Dependent Liquid},
  author = {Fei, Yun (Raymond) and Batty, Christopher and Grinspun, Eitan and Zheng, Changxi},
  year = 2019,
  month = dec,
  journal = {ACM Transactions on Graphics},
  volume = {38},
  number = {6},
  pages = {1--20},
  issn = {0730-0301, 1557-7368},
  doi = {10.1145/3355089.3356532},
  urldate = {2025-08-25},
  langid = {english}
}

@article{fiberShape,
  title = {Quantifying Variation in Human Scalp Hair Fiber Shape and Pigmentation},
  shorttitle = {{{fiberShape}}},
  author = {Lasisi, Tina and Ito, Shosuke and Wakamatsu, Kazumasa and Shaw, Colin N.},
  year = 2016,
  month = jun,
  journal = {American Journal of Physical Anthropology},
  volume = {160},
  number = {2},
  pages = {341--352},
  issn = {1096-8644},
  doi = {10.1002/ajpa.22971},
  langid = {english},
  pmid = {26955790}
}

@misc{FiLM,
  title = {{{FiLM}}: {{Visual Reasoning}} with a {{General Conditioning Layer}}},
  shorttitle = {{{FiLM}}},
  author = {Perez, Ethan and Strub, Florian and de Vries, Harm and Dumoulin, Vincent and Courville, Aaron},
  year = 2017,
  month = dec,
  number = {arXiv:1709.07871},
  eprint = {1709.07871},
  primaryclass = {cs},
  publisher = {arXiv},
  doi = {10.48550/arXiv.1709.07871},
  urldate = {2025-09-29},
  archiveprefix = {arXiv}
}

@inproceedings{FlashAvatar,
  title = {Flashavatar: {{High-fidelity}} Head Avatar with Efficient Gaussian Embedding},
  shorttitle = {{{FlashAvatar}}},
  booktitle = {Proceedings of the {{IEEE}}/{{CVF Conference}} on {{Computer Vision}} and {{Pattern Recognition}}},
  author = {Xiang, Jun and Gao, Xuan and Guo, Yudong and Zhang, Juyong},
  year = 2024,
  pages = {1802--1812}
}

@article{follicleEnigma,
  title = {The Hair Follicle Enigma},
  shorttitle = {{{follicleEnigma}}},
  author = {Bernard, Bruno A},
  year = 2017,
  journal = {Experimental dermatology},
  volume = {26},
  number = {6},
  pages = {472--477},
  publisher = {Wiley Online Library}
}

@book{Fractals,
  title = {The Fractal Geometry of Nature},
  shorttitle = {Fractals},
  author = {Mandelbrot, B. B.},
  year = 1983,
  edition = {3},
  publisher = {{W. H. Freeman and Comp.}},
  address = {New York}
}

@phdthesis{Frenet,
  title = {Sur La Th\'eorie Des Courbes \`a Double Courbure},
  author = {Frenet, Jean Fr{\'e}d{\'e}ric},
  year = 1850,
  school = {Universit\'e de Toulouse}
}

@misc{GAF,
  title = {{{GAF}}: {{Gaussian Avatar Reconstruction}} from {{Monocular Videos}} via {{Multi-view Diffusion}}},
  shorttitle = {{{GAF}}},
  author = {Tang, Jiapeng and Davoli, Davide and Kirschstein, Tobias and Schoneveld, Liam and Niessner, Matthias},
  year = 2025,
  month = apr,
  number = {arXiv:2412.10209},
  eprint = {2412.10209},
  primaryclass = {cs},
  publisher = {arXiv},
  doi = {10.48550/arXiv.2412.10209},
  urldate = {2025-07-21},
  archiveprefix = {arXiv}
}

@misc{gan,
  title = {Generative {{Adversarial Networks}}},
  shorttitle = {Gan},
  author = {Goodfellow, Ian J. and {Pouget-Abadie}, Jean and Mirza, Mehdi and Xu, Bing and {Warde-Farley}, David and Ozair, Sherjil and Courville, Aaron and Bengio, Yoshua},
  year = 2014,
  month = jun,
  number = {arXiv:1406.2661},
  eprint = {1406.2661},
  primaryclass = {stat},
  publisher = {arXiv},
  doi = {10.48550/arXiv.1406.2661},
  urldate = {2025-06-10},
  archiveprefix = {arXiv},
  langid = {english}
}

@misc{GaussianAvatar,
  title = {{{GaussianAvatar}}: {{Towards Realistic Human Avatar Modeling}} from a {{Single Video}} via {{Animatable 3D Gaussians}}},
  shorttitle = {{{GaussianAvatar}}},
  author = {Hu, Liangxiao and Zhang, Hongwen and Zhang, Yuxiang and Zhou, Boyao and Liu, Boning and Zhang, Shengping and Nie, Liqiang},
  year = 2024
}

@inproceedings{GaussianAvatars,
  title = {Gaussianavatars: {{Photorealistic}} Head Avatars with Rigged 3d Gaussians},
  shorttitle = {{{GaussianAvatars}}},
  booktitle = {Proceedings of the {{IEEE}}/{{CVF Conference}} on {{Computer Vision}} and {{Pattern Recognition}}},
  author = {Qian, Shenhan and Kirschstein, Tobias and Schoneveld, Liam and Davoli, Davide and Giebenhain, Simon and Nie{\ss}ner, Matthias},
  year = 2024,
  pages = {20299--20309}
}

@article{gaussianHair,
  title = {{{GaussianHair}}: {{Hair Modeling}} and {{Rendering}} with {{Light-aware Gaussians}}},
  shorttitle = {{{gaussianHair}}},
  author = {Luo, Haimin and Ouyang, Min and Zhao, Zijun and Jiang, Suyi and Zhang, Longwen and Zhang, Qixuan and Yang, Wei and Xu, Lan and Yu, Jingyi and Technology, Deemos and Technology, LumiAni},
  year = 2024,
  langid = {english}
}

@article{GaussianHead,
  title = {Gaussianhead: {{High-fidelity}} Head Avatars with Learnable Gaussian Derivation},
  shorttitle = {{{GaussianHead}}},
  author = {Wang, Jie and Xie, Jiu-Cheng and Li, Xianyan and Xu, Feng and Pun, Chi-Man and Gao, Hao},
  year = 2023,
  journal = {arXiv preprint arXiv:2312.01632},
  eprint = {2312.01632},
  archiveprefix = {arXiv}
}

@article{GaussianHeads,
  title = {{{GaussianHeads}}: {{End-to-End Learning}} of {{Drivable Gaussian Head Avatars}} from {{Coarse-to-fine Representations}}},
  shorttitle = {{{GaussianHeads}}},
  author = {Teotia, Kartik and Kim, Hyeongwoo and Garrido, Pablo and Habermann, Marc and Elgharib, Mohamed and Theobalt, Christian},
  year = 2024,
  journal = {ACM Transactions on Graphics (TOG)},
  volume = {43},
  number = {6},
  pages = {1--12},
  publisher = {ACM New York, NY, USA}
}

@misc{gaussiansplats,
  title = {{{3D Gaussian Splatting}} for {{Real-Time Radiance Field Rendering}}},
  shorttitle = {Gaussiansplats},
  author = {Kerbl, Bernhard and Kopanas, Georgios and Leimk{\~A}{\OE}hler, Thomas and Drettakis, George},
  year = 2023
}

@article{genGuides,
  title = {Transforming {{Unstructured Hair Strands}} into {{Procedural Hair Grooms}}},
  shorttitle = {{{genGuides}}},
  author = {Chang, Wesley and Russell, Andrew and Grabli, Stephane and Chiang, Matt Jen-Yuan and Hery, Christophe and Roble, Doug and Ramamoorthi, Ravi and Li, Tzu-Mao and Maury, Olivier},
  year = 2025,
  month = may,
  journal = {ACM Trans. Graph.},
  volume = {44},
  number = {August 2025},
  doi = {10.1145/3731168},
  urldate = {2025-08-04},
  chapter = {publications},
  langid = {english}
}

@misc{GeoWizard,
  title = {{{GeoWizard}}: {{Unleashing}} the {{Diffusion Priors}} for {{3D Geometry Estimation}} from a {{Single Image}}},
  shorttitle = {{{GeoWizard}}},
  author = {Fu, Xiao and Yin, Wei and Hu, Mu and Wang, Kaixuan and Ma, Yuexin and Tan, Ping and Shen, Shaojie and Lin, Dahua and Long, Xiaoxiao},
  year = 2024,
  month = mar,
  number = {arXiv:2403.12013},
  eprint = {2403.12013},
  primaryclass = {cs},
  publisher = {arXiv},
  doi = {10.48550/arXiv.2403.12013},
  urldate = {2025-06-04},
  archiveprefix = {arXiv}
}

@book{Gray2000,
  title = {Human {{Hair Diversity}}},
  author = {Gray, John},
  year = 2000,
  publisher = {Wiley-Blackwell},
  address = {Abingdon, Oxon},
  isbn = {978-0-632-05672-9},
  langid = {english}
}

@article{GroomCap,
  title = {{{GroomCap}}: {{High-Fidelity Prior-Free Hair Capture}}},
  shorttitle = {{{GroomCap}}},
  author = {Zhou, Yuxiao and Chai, Menglei and Wang, Daoye and Winberg, Sebastian and Wood, Erroll and Sarkar, Kripasindhu and Gross, Markus and Beeler, Thabo},
  year = 2024,
  month = dec,
  journal = {ACM Transactions on Graphics},
  volume = {43},
  number = {6},
  eprint = {2409.00831},
  primaryclass = {cs},
  pages = {1--15},
  issn = {0730-0301, 1557-7368},
  doi = {10.1145/3687768},
  urldate = {2025-06-05},
  archiveprefix = {arXiv}
}

@article{HAAR,
  title = {{{HAAR}}: {{Text-Conditioned Generative Model}} of {{3D Strand-based Human Hairstyles}}},
  shorttitle = {{{HAAR}}},
  author = {Sklyarova, Vanessa and Zakharov, Egor and Hilliges, Otmar and Black, Michael J and Thies, Justus},
  year = 2023,
  month = dec,
  journal = {ArXiv}
}

@misc{HairCLIPv2,
  title = {{{HairCLIPv2}}: {{Unifying Hair Editing}} via {{Proxy Feature Blending}}},
  shorttitle = {{{HairCLIPv2}}},
  author = {Wei, Tianyi and Chen, Dongdong and Zhou, Wenbo and Liao, Jing and Zhang, Weiming and Hua, Gang and Yu, Nenghai},
  year = 2023,
  month = oct,
  number = {arXiv:2310.10651},
  eprint = {2310.10651},
  primaryclass = {cs},
  publisher = {arXiv},
  doi = {10.48550/arXiv.2310.10651},
  urldate = {2025-06-05},
  archiveprefix = {arXiv}
}

@misc{HairCode,
  title = {Custom {{Haircare Backed}} by {{Science}} \textbar{} {{Hair Type Quiz}} \textbar{} {{HairCode}}},
  shorttitle = {{{HairCode}}},
  author = {{Proctor and Gamble}},
  year = 2026,
  urldate = {2026-01-22},
  howpublished = {https://haircode.com/},
  langid = {english}
}

@article{hairDifferences,
  title = {How Different Is Human Hair? {{A}} Critical Appraisal of the Reported Differences in Global Hair Fibre Characteristics and Properties towards Defining a More Relevant Framework for Hair Type Classification},
  shorttitle = {{{hairDifferences}}},
  author = {Daniels, Gabriela and Fraser, Ashiana and Westgate, Gillian E.},
  year = 2023,
  journal = {International Journal of Cosmetic Science},
  volume = {45},
  number = {1},
  pages = {50--61},
  issn = {1468-2494},
  doi = {10.1111/ics.12819},
  urldate = {2025-06-02},
  copyright = {\copyright{} 2022 Society of Cosmetic Scientists and Societe Francaise de Cosmetologie.},
  langid = {english}
}

@article{hairGAN,
  title = {Hair-{{GAN}}: {{Recovering 3D}} Hair Structure from a Single Image Using Generative Adversarial Networks},
  shorttitle = {{{hairGAN}}},
  author = {Zhang, Meng and Zheng, Youyi},
  year = 2019,
  journal = {Visual Informatics},
  volume = {3},
  number = {2},
  pages = {102--112},
  publisher = {Elsevier}
}

@inproceedings{Hairmony,
  title = {Hairmony: {{Fairness-aware}} Hairstyle Classification},
  shorttitle = {Hairmony},
  booktitle = {{{SIGGRAPH Asia}} 2024 {{Conference Papers}}},
  author = {Meishvili, Givi and Clemoes, James and Hewitt, Charlie and Hosenie, Zafiirah and Xiao, Xian and Gorce, Martin de La and Takacs, Tibor and Baltrusaitis, Tadas and Criminisi, Antonio and McRae, Chyna and Jablonski, Nina and Wilczkowiak, Marta},
  year = 2024,
  month = dec,
  eprint = {2410.11528},
  primaryclass = {cs},
  pages = {1--11},
  doi = {10.1145/3680528.3687582},
  urldate = {2025-06-05},
  archiveprefix = {arXiv}
}

@inproceedings{HairNeRF,
  title = {{{HairNeRF}}: {{Geometry-Aware Image Synthesis}} for {{Hairstyle Transfer}}},
  shorttitle = {{{HairNeRF}}},
  booktitle = {Proceedings of the {{IEEE}}/{{CVF International Conference}} on {{Computer Vision}}},
  author = {Chang, Seunggyu and Kim, Gihoon and Kim, Hayeon},
  year = 2023,
  pages = {2448--2458}
}

@inproceedings{HairNet,
  title = {{{HairNet}}: {{Hairstyle Transfer}} with~{{Pose Changes}}},
  shorttitle = {{{HairNet}}},
  booktitle = {Computer {{Vision}} -- {{ECCV}} 2022: 17th {{European Conference}}, {{Tel Aviv}}, {{Israel}}, {{October}} 23--27, 2022, {{Proceedings}}, {{Part XVI}}},
  author = {Zhu, Peihao and Abdal, Rameen and Femiani, John and Wonka, Peter},
  year = 2022,
  month = oct,
  pages = {651--667},
  publisher = {Springer-Verlag},
  address = {Berlin, Heidelberg},
  doi = {10.1007/978-3-031-19787-1_37},
  urldate = {2025-06-05},
  isbn = {978-3-031-19786-4}
}

@article{hairProperties,
  title = {Hair Fiber Characteristics and Methods to Evaluate Hair Physical and Mechanical Properties},
  shorttitle = {{{hairProperties}}},
  author = {Velasco, Maria Val{\'e}ria Robles and Dias, Tania Cristina de S{\'a} and de Freitas, Anderson Zanardi and J{\'u}nior, Nilson Dias Vieira and Pinto, Claudin{\'e}ia Aparecida Sales de Oliveira and Kaneko, Telma Mary and Baby, Andr{\'e} Rolim},
  year = 2009,
  journal = {Brazilian Journal of pharmaceutical sciences},
  volume = {45},
  pages = {153--162},
  publisher = {SciELO Brasil}
}

@article{hairScans,
  title = {Quantifying Whole Human Hair Scalp Fibres of Varying Curl: {{A}} Micro-computed Tomographic Study},
  shorttitle = {{{hairScans}}},
  author = {{van den Berg}, Claire and Khumalo, Nonhlanhla P. and Ngoepe, Malebogo N.},
  year = 2025,
  month = feb,
  journal = {Journal of Microscopy},
  volume = {297},
  number = {2},
  pages = {227--251},
  issn = {0022-2720},
  doi = {10.1111/jmi.13365},
  urldate = {2025-11-03},
  pmcid = {PMC11733847},
  pmid = {39564786}
}

@inproceedings{hairStep,
  title = {Hairstep: {{Transfer}} Synthetic to Real Using Strand and Depth Maps for Single-View 3d Hair Modeling},
  shorttitle = {{{hairStep}}},
  booktitle = {Proceedings of the {{IEEE}}/{{CVF Conference}} on {{Computer Vision}} and {{Pattern Recognition}}},
  author = {Zheng, Yujian and Jin, Zirong and Li, Moran and Huang, Haibin and Ma, Chongyang and Cui, Shuguang and Han, Xiaoguang},
  year = 2023,
  pages = {12726--12735}
}

@article{hairStructure,
  title = {The Structure of People's Hair},
  shorttitle = {{{hairStructure}}},
  author = {Yang, Fei-Chi and Zhang, Yuchen and Rheinst{\"a}dter, Maikel C},
  year = 2014,
  month = oct,
  journal = {PeerJ},
  volume = {2},
  pages = {e619},
  publisher = {PeerJ},
  langid = {english}
}

@misc{hairwater,
  title = {A {{Robust Volume Conserving Method}} for {{Character-Water Interaction}}},
  shorttitle = {Hairwater},
  author = {Lee, Minjae and Hyde, David and Li, Kevin and Fedkiw, Ronald},
  year = 2019
}

@misc{HeadGaS,
  title = {{{HeadGaS}}: {{Real-Time Animatable Head Avatars}} via {{3D Gaussian Splatting}}},
  shorttitle = {{{HeadGaS}}},
  author = {Dhamo, Helisa and Nie, Yinyu and Moreau, Arthur and Song, Jifei and Shaw, Richard and Zhou, Yiren and {P{\'e}rez-Pellitero}, Eduardo},
  year = 2024,
  month = aug,
  number = {arXiv:2312.02902},
  eprint = {2312.02902},
  primaryclass = {cs},
  publisher = {arXiv},
  doi = {10.48550/arXiv.2312.02902},
  urldate = {2025-06-04},
  archiveprefix = {arXiv}
}

@inproceedings{HifiGaussianHeadAvatar,
  title = {Gaussian Head Avatar: {{Ultra}} High-Fidelity Head Avatar via Dynamic Gaussians},
  shorttitle = {{{HifiGaussianHeadAvatar}}},
  booktitle = {Proceedings of the {{IEEE}}/{{CVF}} Conference on Computer Vision and Pattern Recognition},
  author = {Xu, Yuelang and Chen, Benwang and Li, Zhe and Zhang, Hongwen and Wang, Lizhen and Zheng, Zerong and Liu, Yebin},
  year = 2024,
  pages = {1931--1941}
}

@misc{Ho2020,
  title = {Denoising {{Diffusion Probabilistic Models}}},
  author = {Ho, Jonathan and Jain, Ajay and Abbeel, Pieter},
  year = 2020,
  month = dec,
  number = {arXiv:2006.11239},
  eprint = {2006.11239},
  primaryclass = {cs},
  publisher = {arXiv},
  doi = {10.48550/arXiv.2006.11239},
  urldate = {2025-06-10},
  archiveprefix = {arXiv}
}

@article{Hochreiter1997,
  title = {Long {{Short-Term Memory}}},
  author = {Hochreiter, Sepp and Schmidhuber, J{\"u}rgen},
  year = 1997,
  month = nov,
  journal = {Neural Computation},
  volume = {9},
  pages = {1735--1780},
  doi = {10.1162/neco.1997.9.8.1735}
}

@misc{houdini,
  title = {Houdini},
  shorttitle = {Houdini},
  author = {{SideFX}},
  year = 2026,
  howpublished = {Side Effects Software Inc.}
}

@article{hq3DAvatar,
  title = {Hq3davatar: {{High-quality}} Implicit 3d Head Avatar},
  shorttitle = {{{hq3DAvatar}}},
  author = {Teotia, Kartik and Pan, Xingang and Kim, Hyeongwoo and Garrido, Pablo and Elgharib, Mohamed and Theobalt, Christian},
  year = 2024,
  journal = {ACM Transactions on Graphics},
  volume = {43},
  number = {3},
  pages = {1--24},
  publisher = {ACM New York, NY}
}

@article{HSSAN,
  title = {{{HSSAN}}: Hair Synthesis with Style-Guided Spatially Adaptive Normalization on Generative Adversarial Network},
  shorttitle = {{{HSSAN}}},
  author = {Hu, Xinrong and Chang, Qing and Huang, Junjie and Luo, Ruiqi and Wang, Bangchao and Hu, Chang},
  year = 2023,
  month = aug,
  journal = {The Visual Computer},
  volume = {39},
  number = {8},
  pages = {3311--3318},
  issn = {1432-2315},
  doi = {10.1007/s00371-023-02998-5},
  urldate = {2025-06-05},
  langid = {english}
}

@article{Hsu2023,
  title = {Sag-{{Free Initialization}} for {{Strand-Based Hybrid Hair Simulation}}},
  author = {Hsu, Jerry and Wang, Tongtong and Pan, Zherong and Gao, Xifeng and Yuksel, Cem and Wu, Kui},
  year = 2023,
  month = aug,
  journal = {ACM Transactions on Graphics},
  volume = {42},
  number = {4},
  pages = {1--14},
  issn = {0730-0301, 1557-7368},
  doi = {10.1145/3592143},
  urldate = {2025-06-04},
  langid = {english}
}

@article{Hsu2024,
  title = {Real-Time {{Physically Guided Hair Interpolation}}},
  author = {Hsu, Jerry and Wang, Tongtong and Pan, Zherong and Gao, Xifeng and Yuksel, Cem and Wu, Kui},
  year = 2024,
  month = jul,
  journal = {ACM Transactions on Graphics},
  volume = {43},
  number = {4},
  pages = {1--11},
  issn = {0730-0301, 1557-7368},
  doi = {10.1145/3658176},
  urldate = {2025-06-04},
  langid = {english}
}

@inproceedings{Hsu2025,
  title = {Stable {{Cosserat Rods}}},
  booktitle = {Proceedings of the {{Special Interest Group}} on {{Computer Graphics}} and {{Interactive Techniques Conference Conference Papers}}},
  author = {Hsu, Jerry and Wang, Tongtong and Wu, Kui and Yuksel, Cem},
  year = 2025,
  month = aug,
  pages = {1--10},
  publisher = {ACM},
  address = {Vancouver BC Canada},
  doi = {10.1145/3721238.3730618},
  urldate = {2025-08-18},
  isbn = {979-8-4007-1540-2},
  langid = {english}
}

@misc{HumanRef,
  title = {{{HumanRef}}: {{Single Image}} to {{3D Human Generation}} via {{Reference-Guided Diffusion}}},
  shorttitle = {{{HumanRef}}},
  author = {Zhang, Jingbo and Li, Xiaoyu and Zhang, Qi and Cao, Yanpei and Shan, Ying and Liao, Jing},
  year = 2023,
  month = nov,
  number = {arXiv:2311.16961},
  eprint = {2311.16961},
  primaryclass = {cs},
  publisher = {arXiv},
  doi = {10.48550/arXiv.2311.16961},
  urldate = {2025-06-04},
  archiveprefix = {arXiv}
}

@misc{HVH,
  title = {{{HVH}}: {{Learning}} a {{Hybrid Neural Volumetric Representation}} for {{Dynamic Hair Performance Capture}}},
  shorttitle = {{{HVH}}},
  author = {Wang, Ziyan and Nam, Giljoo and Stuyck, Tuur and Lombardi, Stephen and Zollhoefer, Michael and Hodgins, Jessica and Lassner, Christoph},
  year = 2021,
  month = dec,
  number = {arXiv:2112.06904},
  eprint = {2112.06904},
  primaryclass = {cs},
  publisher = {arXiv},
  doi = {10.48550/arXiv.2112.06904},
  urldate = {2025-06-05},
  archiveprefix = {arXiv},
  langid = {english}
}

@article{HyperNeRF,
  title = {{{HyperNeRF}}: {{A Higher-Dimensional Representation}} for {{Topologically Varying Neural Radiance Fields}}},
  shorttitle = {{{HyperNeRF}}},
  author = {Park, Keunhong and Sinha, Utkarsh and Hedman, Peter and Barron, Jonathan T. and Bouaziz, Sofien and Goldman, Dan B. and {Martin-Brualla}, Ricardo and Seitz, Steven M.},
  year = 2021,
  journal = {CoRR},
  volume = {abs/2106.13228},
  eprint = {2106.13228},
  archiveprefix = {arXiv}
}

@article{IfBigNatural2015,
  title = {If {{Big}}, {{Natural Hair}} Is {{In}}, {{Why Don}}'t {{We See}} It {{On Television}}?},
  shorttitle = {{{naturalHairTV}}},
  author = {{NBC News}},
  year = 2015,
  month = jan,
  journal = {NBC News},
  urldate = {2025-06-02},
  langid = {english}
}

@article{ImpactOfFilms,
  title = {Impact of {{Films}}: {{Changes}} in {{Young People}}'s {{Attitudes}} after {{Watching}} a {{Movie}}},
  shorttitle = {{{ImpactOfFilms}}},
  author = {Kubrak, Tina},
  year = 2020,
  journal = {Behavioral Sciences},
  volume = {10},
  number = {5},
  issn = {2076-328X},
  doi = {10.3390/bs10050086}
}

@inproceedings{insta,
  title = {Instant Volumetric Head Avatars},
  shorttitle = {Insta},
  booktitle = {Proceedings of the {{IEEE}}/{{CVF}} Conference on Computer Vision and Pattern Recognition},
  author = {Zielonka, Wojciech and Bolkart, Timo and Thies, Justus},
  year = 2023,
  pages = {4574--4584}
}

@misc{IntrinsicAvatar,
  title = {{{IntrinsicAvatar}}: {{Physically Based Inverse Rendering}} of {{Dynamic Humans}} from {{Monocular Videos}} via {{Explicit Ray Tracing}}},
  shorttitle = {{{IntrinsicAvatar}}},
  author = {Wang, Shaofei and Anti{\'c}, Bo{\v z}idar and Geiger, Andreas and Tang, Siyu},
  year = 2024,
  month = jul,
  number = {arXiv:2312.05210},
  eprint = {2312.05210},
  primaryclass = {cs},
  publisher = {arXiv},
  doi = {10.48550/arXiv.2312.05210},
  urldate = {2025-06-04},
  archiveprefix = {arXiv}
}

@misc{Karras2022,
  title = {Elucidating the {{Design Space}} of {{Diffusion-Based Generative Models}}},
  author = {Karras, Tero and Aittala, Miika and Aila, Timo and Laine, Samuli},
  year = 2022,
  month = oct,
  number = {arXiv:2206.00364},
  eprint = {2206.00364},
  primaryclass = {cs},
  publisher = {arXiv},
  doi = {10.48550/arXiv.2206.00364},
  urldate = {2025-06-10},
  archiveprefix = {arXiv}
}

@article{kimPorousModelsEnhanced2025,
  title = {Porous {{Models}} for {{Enhanced Representation}} of {{Saturated Curly Hairs}}: {{Simulation}} and {{Learning}}},
  shorttitle = {{{wetCurls}}},
  author = {Kim, Jong-Hyun and Lee, Jung},
  year = 2025,
  journal = {IEEE Access},
  volume = {13},
  pages = {84517--84540},
  issn = {2169-3536},
  doi = {10.1109/ACCESS.2025.3569324},
  urldate = {2025-08-25}
}

@misc{kimRacistLegacy,
  title = {The {{Racist Legacy}} of {{Computer-Generated Humans}}},
  shorttitle = {{{kimRacistLegacy}}},
  author = {Kim, Ted},
  year = 2020,
  month = aug
}

@book{Kirchhoff1876,
  title = {Vorlesungen \"Uber {{Mathematische Physik}}: {{Mechanik}}},
  author = {Kirchhoff, Gustav Robert},
  year = 1876,
  edition = {1},
  publisher = {B. G. Teubner},
  address = {Leipzig}
}

@book{Kutta1901,
  title = {{Beitrag zur n\"aherungsweisen Integration totaler Differentialgleichungen}},
  author = {Kutta, Wilhelm},
  year = 1901,
  publisher = {Teubner},
  googlebooks = {Zc4TAQAAIAAJ},
  langid = {ngerman}
}

@article{Lespagnol2024,
  title = {A Mixed-Dimensional Formulation for the Simulation of Slender Structures Immersed in an Incompressible Flow},
  author = {Lespagnol, Fabien and Grandmont, C{\'e}line and Zunino, Paolo and Fern{\'a}ndez, Miguel A.},
  year = 2024,
  month = dec,
  journal = {Computer Methods in Applied Mechanics and Engineering},
  volume = {432},
  pages = {117316},
  issn = {0045-7825},
  doi = {10.1016/j.cma.2024.117316},
  urldate = {2025-06-02}
}

@article{lipidcontent,
  title = {A Systematic Review on the Lipid Composition of Human Hair},
  shorttitle = {Lipidcontent},
  author = {Csuka, David and Csuka, Ella and Juh{\~A}{\textexclamdown}sz, Margit and Sharma, Ajay and Mesinkovska, Natasha},
  year = 2022,
  month = feb,
  journal = {International Journal of Dermatology},
  volume = {62},
  doi = {10.1111/ijd.16109}
}

@misc{Loshchilov2019,
  title = {Decoupled {{Weight Decay Regularization}}},
  author = {Loshchilov, Ilya and Hutter, Frank},
  year = 2019,
  month = jan,
  number = {arXiv:1711.05101},
  eprint = {1711.05101},
  primaryclass = {cs},
  publisher = {arXiv},
  doi = {10.48550/arXiv.1711.05101},
  urldate = {2026-01-18},
  archiveprefix = {arXiv}
}

@article{Loussouarn2001,
  title = {African Hair Growth Parameters},
  author = {Loussouarn, G.},
  year = 2001,
  month = aug,
  journal = {British Journal of Dermatology},
  volume = {145},
  number = {2},
  pages = {294--297},
  issn = {0007-0963},
  doi = {10.1046/j.1365-2133.2001.04350.x},
  urldate = {2025-06-03}
}

@article{Loussouarn2007,
  title = {Worldwide Diversity of Hair Curliness: A New Method of Assessment},
  author = {Loussouarn, Genevi{\~A}{\v s}ve and Garcel, Anne-Lise and Lozano, Isabelle and Collaudin, Catherine and Porter, Crystal and Panhard, S{\~A}{\copyright}gol{\~A}{\v s}ne and {Saint-L{\~A}{\copyright}ger}, Didier and De La Mettrie, Roland},
  year = 2007,
  journal = {International Journal of Dermatology},
  volume = {46},
  number = {s1},
  pages = {2--6},
  doi = {10.1111/j.1365-4632.2007.03453.x}
}

@book{Lsystems,
  title = {The Algorithmic Beauty of Plants},
  shorttitle = {Lsystems},
  author = {Prusinkiewicz, P. and Lindenmayer, Aristid},
  year = 1990,
  publisher = {Springer-Verlag},
  address = {Berlin, Heidelberg},
  isbn = {0-387-97297-8}
}

@misc{LUCAS,
  title = {{{LUCAS}}: {{Layered Universal Codec Avatars}}},
  shorttitle = {{{LUCAS}}},
  author = {Liu, Di and Deng, Teng and Nam, Giljoo and Rong, Yu and Pidhorskyi, Stanislav and Li, Junxuan and Saragih, Jason and Metaxas, Dimitris N. and Cao, Chen},
  year = 2025,
  month = mar,
  number = {arXiv:2502.19739},
  eprint = {2502.19739},
  primaryclass = {cs},
  publisher = {arXiv},
  doi = {10.48550/arXiv.2502.19739},
  urldate = {2025-07-21},
  archiveprefix = {arXiv}
}

@article{macrofibrils,
  title = {Three-Dimensional Architecture of Macrofibrils in the Human Scalp Hair Cortex},
  shorttitle = {Macrofibrils},
  author = {Harland, Duane P and Walls, Richard J and Vernon, James A and Dyer, Jolon M and Woods, Joy L and Bell, Fraser},
  year = 2014,
  journal = {Journal of structural biology},
  volume = {185},
  number = {3},
  pages = {397--404},
  publisher = {Elsevier}
}

@misc{Maluleke2022,
  title = {Studying {{Bias}} in {{GANs}} through the {{Lens}} of {{Race}}},
  author = {Maluleke, Vongani H. and Thakkar, Neerja and Brooks, Tim and Weber, Ethan and Darrell, Trevor and Efros, Alexei A. and Kanazawa, Angjoo and Guillory, Devin},
  year = 2022,
  month = sep,
  number = {arXiv:2209.02836},
  eprint = {2209.02836},
  primaryclass = {cs},
  publisher = {arXiv},
  doi = {10.48550/arXiv.2209.02836},
  urldate = {2025-06-02},
  archiveprefix = {arXiv}
}

@misc{matrix,
  title = {Matrix {{Introduces Hair Diversity Matrix}} to {{Ensure All Its Products Work Across All Hair Types}}},
  shorttitle = {Matrix},
  author = {{allure.com}},
  year = 2021,
  month = mar
}

@misc{maya,
  title = {Maya},
  shorttitle = {Maya},
  author = {{Autodesk}},
  year = 2026,
  howpublished = {Autodesk, INC.}
}

@misc{measurePorosity,
  title = {{{HOW TO DETERMINE HAIR POROSITY AND WHAT IT MEANS FOR YOUR HAIR}}},
  shorttitle = {{{measurePorosity}}},
  author = {Curl, Clever},
  year = 2020,
  month = jul
}

@misc{mediaUseStats,
  title = {Media Usage in the {{United States}}},
  shorttitle = {{{mediaUseStats}}},
  author = {Guttmann, A},
  year = 2024,
  month = aug,
  journal = {Media usage in the U.S. - statistics \& facts},
  publisher = {Statista}
}

@article{MERCI,
  title = {{{MERCI}}: {{Mixed Curvature-Based Elements}} for {{Computing Equilibria}} of {{Thin Elastic Ribbons}}},
  shorttitle = {{{MERCI}}},
  author = {Charrondi{\`e}re, Rapha{\"e}l and Neukirch, S{\'e}bastien and {Bertails-Descoubes}, Florence},
  year = 2024,
  month = aug,
  journal = {ACM Trans. Graph.},
  volume = {43},
  number = {5},
  pages = {160:1--160:26},
  issn = {0730-0301},
  doi = {10.1145/3674502},
  urldate = {2025-06-02}
}

@article{Mettrie2007,
  title = {Shape {{Variability}} and {{Classification}} of {{Human Hair}}: {{A Worldwide Approach}}},
  author = {Mettrie, Roland De La and {Saint-L{\'e}ger}, Didier and Loussouarn, Geneviev{\`e}ve and Garcel, Annelise and Porter, Crystal and Langaney, Andr{\'e}},
  year = 2007,
  journal = {Human Biology},
  volume = {79},
  number = {3},
  pages = {265--281},
  publisher = {Wayne State University Press},
  doi = {10.1353/hub.2007.0045}
}

@article{mixedRaceStrands,
  title = {Unique {{Hair Properties}} That {{Emerge}} from {{Combinations}} of {{Multiple Races}}},
  shorttitle = {{{mixedRaceStrands}}},
  author = {Takahashi, Toshie},
  year = 2019,
  month = jun,
  journal = {Cosmetics},
  volume = {6},
  number = {2},
  pages = {36},
  publisher = {Multidisciplinary Digital Publishing Institute},
  issn = {2079-9284},
  doi = {10.3390/cosmetics6020036},
  urldate = {2025-06-03},
  copyright = {http://creativecommons.org/licenses/by/3.0/},
  langid = {english}
}

@inproceedings{monoGaussianAvatar,
  title = {Monogaussianavatar: {{Monocular}} Gaussian Point-Based Head Avatar},
  shorttitle = {{{monoGaussianAvatar}}},
  booktitle = {{{ACM SIGGRAPH}} 2024 {{Conference Papers}}},
  author = {Chen, Yufan and Wang, Lizhen and Li, Qijing and Xiao, Hongjiang and Zhang, Shengping and Yao, Hongxun and Liu, Yebin},
  year = 2024,
  pages = {1--9}
}

@misc{MonoHair,
  title = {{{MonoHair}}: {{High-Fidelity Hair Modeling}} from a {{Monocular Video}}},
  shorttitle = {{{MonoHair}}},
  author = {Wu, Keyu and Yang, Lingchen and Kuang, Zhiyi and Feng, Yao and Han, Xutao and Shen, Yuefan and Fu, Hongbo and Zhou, Kun and Zheng, Youyi},
  year = 2024,
  month = mar,
  number = {arXiv:2403.18356},
  eprint = {2403.18356},
  primaryclass = {cs},
  publisher = {arXiv},
  doi = {10.48550/arXiv.2403.18356},
  urldate = {2025-06-05},
  archiveprefix = {arXiv}
}

@misc{multiResPhysics,
  title = {Learning {{Controllable Adaptive Simulation}} for {{Multi-resolution Physics}}},
  shorttitle = {{{multiResPhysics}}},
  author = {Wu, Tailin and Maruyama, Takashi and Zhao, Qingqing and Wetzstein, Gordon and Leskovec, Jure},
  year = 2023,
  month = may,
  number = {arXiv:2305.01122},
  eprint = {2305.01122},
  primaryclass = {cs},
  publisher = {arXiv},
  doi = {10.48550/arXiv.2305.01122},
  urldate = {2025-06-04},
  archiveprefix = {arXiv}
}

@misc{MVHumanNet,
  title = {{{MVHumanNet}}: {{A Large-scale Dataset}} of {{Multi-view Daily Dressing Human Captures}}},
  shorttitle = {{{MVHumanNet}}},
  author = {Xiong, Zhangyang and Li, Chenghong and Liu, Kenkun and Liao, Hongjie and Hu, Jianqiao and Zhu, Junyi and Ning, Shuliang and Qiu, Lingteng and Wang, Chongjie and Wang, Shijie and Cui, Shuguang and Han, Xiaoguang},
  year = 2023,
  month = dec,
  number = {arXiv:2312.02963},
  eprint = {2312.02963},
  primaryclass = {cs},
  publisher = {arXiv},
  doi = {10.48550/arXiv.2312.02963},
  urldate = {2025-06-05},
  archiveprefix = {arXiv}
}

@book{Natural,
  title = {Natural: {{Black Beauty}} and the {{Politics}} of {{Hair}}},
  shorttitle = {Natural},
  author = {Johnson, Chelsea Mary Elise},
  year = 2024,
  eprint = {jj.28248903},
  eprinttype = {jstor},
  publisher = {NYU Press},
  urldate = {2026-01-22},
  isbn = {978-1-4798-1473-2}
}

@article{NeRF,
  title = {{{NeRF}}: {{Representing Scenes}} as {{Neural Radiance Fields}} for {{View Synthesis}}},
  shorttitle = {{{NeRF}}},
  author = {Mildenhall, Ben and Srinivasan, Pratul P. and Tancik, Matthew and Barron, Jonathan T. and Ramamoorthi, Ravi and Ng, Ren},
  year = 2020,
  journal = {CoRR},
  volume = {abs/2003.08934},
  eprint = {2003.08934},
  archiveprefix = {arXiv}
}

@inproceedings{neuralHaircut,
  title = {Neural {{Haircut}}: {{Prior-Guided Strand-Based Hair Reconstruction}}},
  shorttitle = {{{neuralHaircut}}},
  booktitle = {Proceedings of {{IEEE International Conference}} on {{Computer Vision}} ({{ICCV}})},
  author = {Sklyarova, Vanessa and Chelishev, Jenya and Dogaru, Andreea and Medvedev, Igor and Lempitsky, Victor and Zakharov, Egor},
  year = 2023
}

@inproceedings{neuralHDHair,
  title = {{{NeuralHDHair}}: {{Automatic High-fidelity Hair Modeling}} from a {{Single Image Using Implicit Neural Representations}}},
  shorttitle = {{{neuralHDHair}}},
  booktitle = {Proceedings of the {{IEEE}}/{{CVF Conference}} on {{Computer Vision}} and {{Pattern Recognition}}},
  author = {Wu, Keyu and Ye, Yifan and Yang, Lingchen and Fu, Hongbo and Zhou, Kun and Zheng, Youyi},
  year = 2022,
  pages = {1526--1535}
}

@article{NeuralStrands,
  title = {Neural {{Strands}}: {{Learning Hair Geometry}} and {{Appearance}} from {{Multi-View Images}}},
  shorttitle = {{{NeuralStrands}}},
  author = {Rosu, Radu Alexandru and Saito, Shunsuke and Wang, Ziyan and Wu, Chenglei and Behnke, Sven and Nam, Giljoo},
  year = 2022,
  journal = {ECCV}
}

@misc{Nichol2021,
  title = {Improved {{Denoising Diffusion Probabilistic Models}}},
  author = {Nichol, Alex and Dhariwal, Prafulla},
  year = 2021,
  month = feb,
  number = {arXiv:2102.09672},
  eprint = {2102.09672},
  primaryclass = {cs},
  publisher = {arXiv},
  doi = {10.48550/arXiv.2102.09672},
  urldate = {2025-09-29},
  archiveprefix = {arXiv}
}

@article{nissimovHairCurvatureNatural2014,
  title = {Hair Curvature: A Natural Dialectic and Review},
  shorttitle = {{{TwistAndCurvature}}},
  author = {Nissimov, Joseph N. and Das Chaudhuri, Asit Baran},
  year = 2014,
  journal = {Biological Reviews},
  volume = {89},
  number = {3},
  pages = {723--766},
  doi = {10.1111/brv.12081}
}

@misc{Palette,
  title = {Palette: {{Image-to-Image Diffusion Models}}},
  shorttitle = {Palette},
  author = {Saharia, Chitwan and Chan, William and Chang, Huiwen and Lee, Chris A. and Ho, Jonathan and Salimans, Tim and Fleet, David J. and Norouzi, Mohammad},
  year = 2022,
  month = may,
  number = {arXiv:2111.05826},
  eprint = {2111.05826},
  primaryclass = {cs},
  publisher = {arXiv},
  doi = {10.48550/arXiv.2111.05826},
  urldate = {2025-06-10},
  archiveprefix = {arXiv}
}

@misc{PanoHead,
  title = {{{PanoHead}}: {{Geometry-Aware 3D Full-Head Synthesis}} in 360\$\textasciicircum\textbraceleft\textbackslash circ\textbraceright\$},
  shorttitle = {{{PanoHead}}},
  author = {An, Sizhe and Xu, Hongyi and Shi, Yichun and Song, Guoxian and Ogras, Umit and Luo, Linjie},
  year = 2023,
  month = mar,
  number = {arXiv:2303.13071},
  eprint = {2303.13071},
  primaryclass = {cs},
  publisher = {arXiv},
  doi = {10.48550/arXiv.2303.13071},
  urldate = {2025-06-05},
  archiveprefix = {arXiv}
}

@article{perlinnoise,
  title = {An Image Synthesizer},
  shorttitle = {Perlinnoise},
  author = {Perlin, Ken},
  year = 1985,
  month = jul,
  journal = {SIGGRAPH Comput. Graph.},
  volume = {19},
  number = {3},
  pages = {287--296},
  publisher = {Association for Computing Machinery},
  address = {New York, NY, USA},
  issn = {0097-8930},
  doi = {10.1145/325165.325247}
}

@misc{Perm,
  title = {Perm: {{A Parametric Representation}} for {{Multi-Style 3D Hair Modeling}}},
  shorttitle = {Perm},
  author = {He, Chengan and Sun, Xin and Shu, Zhixin and Luan, Fujun and Pirk, S{\"o}ren and Herrera, Jorge Alejandro Amador and Michels, Dominik L. and Wang, Tuanfeng Y. and Zhang, Meng and Rushmeier, Holly and Zhou, Yi},
  year = 2025,
  month = may,
  number = {arXiv:2407.19451},
  eprint = {2407.19451},
  primaryclass = {cs},
  publisher = {arXiv},
  doi = {10.48550/arXiv.2407.19451},
  urldate = {2025-06-05},
  archiveprefix = {arXiv}
}

@misc{Phase2vec,
  title = {Phase2vec: {{Dynamical}} Systems Embedding with a Physics-Informed Convolutional Network},
  shorttitle = {Phase2vec},
  author = {Ricci, Matthew and Moriel, Noa and Piran, Zoe and Nitzan, Mor},
  year = 2023,
  month = feb,
  number = {arXiv:2212.03857},
  eprint = {2212.03857},
  primaryclass = {cs},
  publisher = {arXiv},
  doi = {10.48550/arXiv.2212.03857},
  urldate = {2025-06-04},
  archiveprefix = {arXiv}
}

@misc{PointAvatar,
  title = {{{PointAvatar}}: {{Deformable Point-based Head Avatars}} from {{Videos}}},
  shorttitle = {{{PointAvatar}}},
  author = {Zheng, Yufeng and Yifan, Wang and Wetzstein, Gordon and Black, Michael J. and Hilliges, Otmar},
  year = 2023,
  month = feb,
  number = {arXiv:2212.08377},
  eprint = {2212.08377},
  primaryclass = {cs},
  publisher = {arXiv},
  doi = {10.48550/arXiv.2212.08377},
  urldate = {2025-06-05},
  archiveprefix = {arXiv}
}

@article{porosityEffect,
  title = {Benefit of Coconut-Based Hair Oil via Hair Porosity Quantification},
  shorttitle = {{{porosityEffect}}},
  author = {Kaushik, Vaibhav and Kumar, Ajeet and Gosvami, Nitya Nand and Gode, Vaishali and Mhaskar, Sudhakar and Kamath, Yash},
  year = 2022,
  journal = {International Journal of Cosmetic Science},
  volume = {44},
  number = {3},
  pages = {289--298},
  doi = {10.1111/ics.12774}
}

@misc{Portrait4D,
  title = {{{Portrait4D}}: {{Learning One-Shot 4D Head Avatar Synthesis}} Using {{Synthetic Data}}},
  shorttitle = {{{Portrait4D}}},
  author = {Deng, Yu and Wang, Duomin and Ren, Xiaohang and Chen, Xingyu and Wang, Baoyuan},
  year = 2024,
  month = jun,
  number = {arXiv:2311.18729},
  eprint = {2311.18729},
  primaryclass = {cs},
  publisher = {arXiv},
  doi = {10.48550/arXiv.2311.18729},
  urldate = {2025-06-05},
  archiveprefix = {arXiv}
}

@article{psavatar,
  title = {Psavatar: A Point-Based Morphable Shape Model for Real-Time Head Avatar Creation with {{3D}} Gaussian Splatting},
  shorttitle = {Psavatar},
  author = {Zhao, Zhongyuan and Bao, Zhenyu and Li, Qing and Qiu, Guoping and Liu, Kanglin},
  year = 2024,
  journal = {arXiv e-prints},
  pages = {arXiv--2401}
}

@inproceedings{pv3d,
  title = {{{PV3D}}: {{A 3D Generative Model}} for {{Portrait Video Generation}}},
  shorttitle = {Pv3d},
  booktitle = {The {{Tenth International Conference}} on {{Learning Representations}}},
  author = {Xu, Zhongcong and Zhang, Jianfeng and Liew, Junhao and Zhang, Wenqing and Bai, Song and Feng, Jiashi and Shou, Mike Zheng},
  year = 2023
}

@misc{Quaffure,
  title = {Quaffure: {{Real-Time Quasi-Static Neural Hair Simulation}}},
  shorttitle = {Quaffure},
  author = {Stuyck, Tuur and Lin, Gene Wei-Chin and Larionov, Egor and Chen, Hsiao-yu and Bozic, Aljaz and Sarafianos, Nikolaos and Roble, Doug},
  year = 2025,
  month = apr,
  number = {arXiv:2412.10061},
  eprint = {2412.10061},
  primaryclass = {cs},
  publisher = {arXiv},
  doi = {10.48550/arXiv.2412.10061},
  urldate = {2025-07-21},
  archiveprefix = {arXiv}
}

@misc{Ramachandran2017,
  title = {Searching for {{Activation Functions}}},
  author = {Ramachandran, Prajit and Zoph, Barret and Le, Quoc V.},
  year = 2017,
  month = oct,
  number = {arXiv:1710.05941},
  eprint = {1710.05941},
  primaryclass = {cs},
  publisher = {arXiv},
  doi = {10.48550/arXiv.1710.05941},
  urldate = {2025-09-29},
  archiveprefix = {arXiv}
}

@article{realtimeNeuralSimulation,
  title = {Real-{{Time Hair Simulation With Neural Interpolation}}},
  shorttitle = {{{realtimeNeuralSimulation}}},
  author = {Lyu, Qing and Chai, Menglei and Chen, Xiang and Zhou, Kun},
  year = 2022,
  month = apr,
  journal = {IEEE Transactions on Visualization and Computer Graphics},
  volume = {28},
  number = {4},
  pages = {1894--1905},
  issn = {1941-0506},
  doi = {10.1109/TVCG.2020.3029823},
  urldate = {2025-06-02}
}

@misc{ReStyle,
  title = {{{ReStyle}}: {{A Residual-Based StyleGAN Encoder}} via {{Iterative Refinement}}},
  shorttitle = {{{ReStyle}}},
  author = {Alaluf, Yuval and Patashnik, Or and {Cohen-Or}, Daniel},
  year = 2021,
  month = aug,
  number = {arXiv:2104.02699},
  eprint = {2104.02699},
  primaryclass = {cs},
  publisher = {arXiv},
  doi = {10.48550/arXiv.2104.02699},
  urldate = {2025-06-04},
  archiveprefix = {arXiv}
}

@article{Romanya-Serrasolsas2025,
  title = {Painless {{Differentiable Rotation Dynamics}}},
  author = {{Romany{\`a}-Serrasolsas}, Mag{\'i} and Casafranca, Juan J. and Otaduy, Miguel A.},
  year = 2025,
  month = aug,
  journal = {ACM Transactions on Graphics},
  volume = {44},
  number = {4},
  pages = {1--13},
  issn = {0730-0301, 1557-7368},
  doi = {10.1145/3730944},
  urldate = {2025-08-04},
  langid = {english}
}

@misc{Sapiens,
  title = {Sapiens: {{Foundation}} for {{Human Vision Models}}},
  shorttitle = {Sapiens},
  author = {Khirodkar, Rawal and Bagautdinov, Timur and Martinez, Julieta and Zhaoen, Su and James, Austin and Selednik, Peter and Anderson, Stuart and Saito, Shunsuke},
  year = 2024,
  month = aug,
  number = {arXiv:2408.12569},
  eprint = {2408.12569},
  primaryclass = {cs},
  publisher = {arXiv},
  doi = {10.48550/arXiv.2408.12569},
  urldate = {2025-07-21},
  archiveprefix = {arXiv}
}

@article{Selle2008,
  title = {A Mass Spring Model for Hair Simulation},
  author = {Selle, Andrew and Lentine, Michael and Fedkiw, Ronald},
  year = 2008,
  month = aug,
  journal = {ACM Transactions on Graphics},
  volume = {27},
  number = {3},
  pages = {1--11},
  issn = {0730-0301, 1557-7368},
  doi = {10.1145/1360612.1360663},
  urldate = {2025-08-21},
  langid = {english}
}

@misc{shapeAdaptation,
  title = {Shape {{Adaptation}} for {{3D Hairstyle Retargeting}}},
  shorttitle = {{{shapeAdaptation}}},
  author = {Yu, Lu and Ren, Zhong and Zheng, Youyi and Chen, Xiang and Zhou, Kun},
  year = 2025,
  month = jul,
  number = {arXiv:2507.12168},
  eprint = {2507.12168},
  primaryclass = {cs},
  publisher = {arXiv},
  doi = {10.48550/arXiv.2507.12168},
  urldate = {2025-08-04},
  archiveprefix = {arXiv}
}

@article{shiLiftedCurlsModel2023,
  title = {Lifted {{Curls}}: {{A Model}} for {{Tightly Coiled Hair Simulation}}},
  shorttitle = {Lifted {{Curls}}},
  author = {Shi, Alvin and Wu, Haomiao and Parr, Jarred and Darke, A. M. and Kim, Theodore},
  year = 2023,
  month = aug,
  journal = {Proceedings of the ACM on Computer Graphics and Interactive Techniques},
  volume = {6},
  number = {3},
  pages = {1--19},
  issn = {2577-6193},
  doi = {10.1145/3606920},
  urldate = {2025-06-05},
  langid = {english}
}

@misc{SimAvatar,
  title = {{{SimAvatar}}: {{Simulation-Ready Avatars}} with {{Layered Hair}} and {{Clothing}}},
  shorttitle = {{{SimAvatar}}},
  author = {Li, Xueting and Yuan, Ye and Mello, Shalini De and Daviet, Gilles and Leaf, Jonathan and Macklin, Miles and Kautz, Jan and Iqbal, Umar},
  year = 2024,
  month = dec,
  number = {arXiv:2412.09545},
  eprint = {2412.09545},
  primaryclass = {cs},
  publisher = {arXiv},
  doi = {10.48550/arXiv.2412.09545},
  urldate = {2025-07-21},
  archiveprefix = {arXiv}
}

@misc{singleViewImplicit,
  title = {Towards {{Unified 3D Hair Reconstruction}} from {{Single-View Portraits}}},
  shorttitle = {{{singleViewImplicit}}},
  author = {Zheng, Yujian and Qiu, Yuda and Jin, Leyang and Ma, Chongyang and Huang, Haibin and Zhang, Di and Wan, Pengfei and Han, Xiaoguang},
  year = 2024,
  month = sep,
  number = {arXiv:2409.16863},
  eprint = {2409.16863},
  primaryclass = {cs},
  publisher = {arXiv},
  doi = {10.48550/arXiv.2409.16863},
  urldate = {2025-06-04},
  archiveprefix = {arXiv}
}

@inproceedings{splattingAvatar,
  title = {Splattingavatar: {{Realistic}} Real-Time Human Avatars with Mesh-Embedded Gaussian Splatting},
  shorttitle = {{{splattingAvatar}}},
  booktitle = {Proceedings of the {{IEEE}}/{{CVF Conference}} on {{Computer Vision}} and {{Pattern Recognition}}},
  author = {Shao, Zhijing and Wang, Zhaolong and Li, Zhuang and Wang, Duotun and Lin, Xiangru and Zhang, Yu and Fan, Mingming and Wang, Zeyu},
  year = 2024,
  pages = {1606--1616}
}

@inproceedings{strandAccurateMVS,
  title = {Strand-{{Accurate Multi-View Hair Capture}}},
  shorttitle = {{{strandAccurateMVS}}},
  author = {Nam, Giljoo and Wu, Chenglei and Kim, Min and Sheikh, Yaser},
  year = 2019,
  month = jun,
  pages = {155--164},
  doi = {10.1109/CVPR.2019.00024}
}

@article{strandLipids,
  title = {The Influence of Hair Lipids in Ethnic Hair Properties},
  shorttitle = {{{strandLipids}}},
  author = {Mart{\'i}, M and Barba, C and Manich, {\relax AM} and Rubio, L and Alonso, C and Coderch, L},
  year = 2016,
  journal = {International journal of cosmetic science},
  volume = {38},
  number = {1},
  pages = {77--84},
  publisher = {Wiley Online Library}
}

@article{Struct2Hair,
  title = {{{Struct2Hair}}: {{A}} Hair Shape Descriptor for Hairstyle Modeling},
  shorttitle = {{{Struct2Hair}}},
  author = {Zhang, Wenshu and Nie, Yinyu and Guo, Shihui and Chang, Jian and Zhang, Jianjun and Tong, Ruofeng},
  year = 2023,
  journal = {Computer Animation and Virtual Worlds},
  volume = {34},
  number = {5},
  pages = {e2128},
  issn = {1546-427X},
  doi = {10.1002/cav.2128},
  urldate = {2025-06-03},
  copyright = {\copyright{} 2022 John Wiley \& Sons Ltd.},
  langid = {english}
}

@misc{StyleCLIP,
  title = {{{StyleCLIP}}: {{Text-Driven Manipulation}} of {{StyleGAN Imagery}}},
  shorttitle = {{{StyleCLIP}}},
  author = {Patashnik, Or and Wu, Zongze and Shechtman, Eli and {Cohen-Or}, Daniel and Lischinski, Dani},
  year = 2021,
  month = mar,
  number = {arXiv:2103.17249},
  eprint = {2103.17249},
  primaryclass = {cs},
  publisher = {arXiv},
  doi = {10.48550/arXiv.2103.17249},
  urldate = {2025-06-04},
  archiveprefix = {arXiv}
}

@misc{stylegan,
  title = {A {{Style-Based Generator Architecture}} for {{Generative Adversarial Networks}}},
  shorttitle = {Stylegan},
  author = {Karras, Tero and Laine, Samuli and Aila, Timo},
  year = 2019,
  month = mar,
  number = {arXiv:1812.04948},
  eprint = {1812.04948},
  primaryclass = {cs},
  publisher = {arXiv},
  doi = {10.48550/arXiv.1812.04948},
  urldate = {2025-06-10},
  archiveprefix = {arXiv}
}

@inproceedings{styleGANSalon,
  title = {{{StyleGAN Salon}}: {{Multi-View Latent Optimization}} for {{Pose-Invariant Hairstyle Transfer}}},
  shorttitle = {{{styleGANSalon}}},
  booktitle = {{{IEEE Conference}} on {{Computer Vision}} and {{Pattern Recognition}} ({{CVPR}})},
  author = {Khwanmuang, Sasikarn and Phongthawee, Pakkapon and Sangkloy, Patsorn and Suwajanakorn, Supasorn},
  year = 2023
}

@misc{StyleYourHair,
  title = {Style {{Your Hair}}: {{Latent Optimization}} for {{Pose-Invariant Hairstyle Transfer}} via {{Local-Style-Aware Hair Alignment}}},
  shorttitle = {{{StyleYourHair}}},
  author = {Kim, Taewoo and Chung, Chaeyeon and Kim, Yoonseo and Park, Sunghyun and Kim, Kangyeol and Choo, Jaegul},
  year = 2022,
  month = aug,
  number = {arXiv:2208.07765},
  eprint = {2208.07765},
  primaryclass = {cs},
  publisher = {arXiv},
  doi = {10.48550/arXiv.2208.07765},
  urldate = {2025-06-03},
  archiveprefix = {arXiv}
}

@misc{Takahashi2025,
  title = {Rest {{Shape Optimization}} for {{Sag-Free Discrete Elastic Rods}}},
  shorttitle = {Takahashi2025},
  author = {Takahashi, Tetsuya and Batty, Christopher},
  year = 2025,
  month = feb,
  number = {arXiv:2409.12362},
  eprint = {2409.12362},
  primaryclass = {cs},
  publisher = {arXiv},
  doi = {10.48550/arXiv.2409.12362},
  urldate = {2025-06-02},
  archiveprefix = {arXiv}
}

@article{Tencers,
  title = {Tencers: {{Tension-Constrained Elastic Rods}}},
  shorttitle = {Tencers},
  author = {Dandy, Liliane-Joy and Vidulis, Michele and Ren, Yingying and Pauly, Mark},
  year = 2024,
  month = dec,
  journal = {ACM Transactions on Graphics},
  volume = {43},
  number = {6},
  pages = {1--13},
  issn = {0730-0301, 1557-7368},
  doi = {10.1145/3687967},
  urldate = {2025-06-02},
  langid = {english}
}

@misc{textTo3DAware,
  title = {Fast {{Text-to-3D-Aware Face Generation}} and {{Manipulation}} via {{Direct Cross-modal Mapping}} and {{Geometric Regularization}}},
  shorttitle = {{{textTo3DAware}}},
  author = {Zhang, Jinlu and Zhou, Yiyi and Zheng, Qiancheng and Du, Xiaoxiong and Luo, Gen and Peng, Jun and Sun, Xiaoshuai and Ji, Rongrong},
  year = 2024,
  month = aug,
  number = {arXiv:2403.06702},
  eprint = {2403.06702},
  primaryclass = {cs},
  publisher = {arXiv},
  doi = {10.48550/arXiv.2403.06702},
  urldate = {2025-06-03},
  archiveprefix = {arXiv}
}

@techreport{TexturedTrends,
  title = {Textured {{Trends Report}}},
  shorttitle = {{{texturedTrends}}},
  author = {Spiewak, Anna},
  year = 2018,
  institution = {L'Oreal}
}

@article{Thibaut2007,
  title = {Human Hair Keratin Network and Curvature},
  shorttitle = {Thibaut2007},
  author = {Thibaut, Sebastien and Barbarat, Philippe and Leroy, Frederic and Bernard, Bruno A.},
  year = 2007,
  journal = {International Journal of Dermatology},
  volume = {46},
  number = {s1},
  pages = {7--10},
  issn = {1365-4632},
  doi = {10.1111/j.1365-4632.2007.03454.x},
  urldate = {2025-06-02},
  langid = {english}
}

@article{TowardsRealtime,
  title = {Towards {{Realtime}}: {{A Hybrid Physics-based Method}} for {{Hair Animation}} on {{GPU}}},
  shorttitle = {Towards {{Realtime}}},
  author = {Huang, Li and Yang, Fan and Wei, Chendi and Chen, Yu Ju (Edwin) and Yuan, Chun and Gao, Ming},
  year = 2023,
  month = aug,
  journal = {Proceedings of the ACM on Computer Graphics and Interactive Techniques},
  volume = {6},
  number = {3},
  pages = {1--18},
  issn = {2577-6193},
  doi = {10.1145/3606937},
  urldate = {2025-06-02},
  langid = {english}
}

@misc{Tri2plane,
  title = {Tri\$\textasciicircum\textbraceleft 2\textbraceright\$-Plane: {{Thinking Head Avatar}} via {{Feature Pyramid}}},
  shorttitle = {Tri2plane},
  author = {Song, Luchuan and Liu, Pinxin and Chen, Lele and Yin, Guojun and Xu, Chenliang},
  year = 2024,
  month = jul,
  number = {arXiv:2401.09386},
  eprint = {2401.09386},
  primaryclass = {cs},
  publisher = {arXiv},
  doi = {10.48550/arXiv.2401.09386},
  urldate = {2025-06-04},
  archiveprefix = {arXiv}
}

@inproceedings{typeFromImage,
  title = {Towards Creation of a Curl Pattern Recognition System},
  shorttitle = {{{typeFromImage}}},
  booktitle = {Proceedings of the {{International Conference}} on {{Image Processing}}, {{Computer Vision}}, and {{Pattern Recognition}} ({{IPCV}})},
  author = {Kymberlee, Hill and Adesola, Abimbola and Prajjwhal, Dangal and Washington, Gloria and Legand, Burge III},
  year = 2018,
  pages = {30--33},
  publisher = {The Steering Committee of The World Congress in Computer Science, Computer \dots}
}

@misc{U-Net,
  title = {U-{{Net}}: {{Convolutional Networks}} for {{Biomedical Image Segmentation}}},
  shorttitle = {U-{{Net}}},
  author = {Ronneberger, Olaf and Fischer, Philipp and Brox, Thomas},
  year = 2015,
  month = may,
  number = {arXiv:1505.04597},
  eprint = {1505.04597},
  primaryclass = {cs},
  publisher = {arXiv},
  doi = {10.48550/arXiv.1505.04597},
  urldate = {2025-09-29},
  archiveprefix = {arXiv}
}

@misc{Vaswani2023,
  title = {Attention {{Is All You Need}}},
  author = {Vaswani, Ashish and Shazeer, Noam and Parmar, Niki and Uszkoreit, Jakob and Jones, Llion and Gomez, Aidan N. and Kaiser, Lukasz and Polosukhin, Illia},
  year = 2023,
  month = aug,
  number = {arXiv:1706.03762},
  eprint = {1706.03762},
  primaryclass = {cs},
  publisher = {arXiv},
  doi = {10.48550/arXiv.1706.03762},
  urldate = {2025-06-10},
  archiveprefix = {arXiv}
}

@misc{VenusDeWillendorf,
  title = {How {{This}} 30,000-{{Year-Old Figurine Continues}} to {{Captivate Today}}},
  shorttitle = {{{VenusDeWillendorf}}},
  author = {{Richman-Abdou}, Kelly},
  year = 2019
}

@misc{venusOfBrassempouy,
  title = {The {{Venus}} of {{Brassempouy}}, {{One}} of the {{Earliest Known Realistic Representations}} of a {{Human Face}}},
  shorttitle = {{{venusOfBrassempouy}}},
  author = {Norman, Jeremy M.},
  year = 2004,
  publisher = {Jeremy Norman's History of Information}
}

@misc{VoxGRAF,
  title = {{{VoxGRAF}}: {{Fast 3D-Aware Image Synthesis}} with {{Sparse Voxel Grids}}},
  shorttitle = {{{VoxGRAF}}},
  author = {Schwarz, Katja and Sauer, Axel and Niemeyer, Michael and Liao, Yiyi and Geiger, Andreas},
  year = 2022,
  month = nov,
  number = {arXiv:2206.07695},
  eprint = {2206.07695},
  primaryclass = {cs},
  publisher = {arXiv},
  doi = {10.48550/arXiv.2206.07695},
  urldate = {2025-06-03},
  archiveprefix = {arXiv}
}

@misc{wuGroupNormalization2018,
  title = {Group {{Normalization}}},
  author = {Wu, Yuxin and He, Kaiming},
  year = 2018,
  month = jun,
  number = {arXiv:1803.08494},
  eprint = {1803.08494},
  primaryclass = {cs},
  publisher = {arXiv},
  doi = {10.48550/arXiv.1803.08494},
  urldate = {2025-09-29},
  archiveprefix = {arXiv}
}

@misc{Xiong2020,
  title = {On {{Layer Normalization}} in the {{Transformer Architecture}}},
  author = {Xiong, Ruibin and Yang, Yunchang and He, Di and Zheng, Kai and Zheng, Shuxin and Xing, Chen and Zhang, Huishuai and Lan, Yanyan and Wang, Liwei and Liu, Tie-Yan},
  year = 2020,
  month = jun,
  number = {arXiv:2002.04745},
  eprint = {2002.04745},
  primaryclass = {cs},
  publisher = {arXiv},
  doi = {10.48550/arXiv.2002.04745},
  urldate = {2026-05-05},
  archiveprefix = {arXiv}
}

@article{xuPersonalizedHairstyleHair2024,
  title = {Personalized Hairstyle and Hair Color Editing Based on Multi-Feature Fusion},
  shorttitle = {{{hairEditingImage}}},
  author = {Xu, Jiayi and Zhang, Chenming and Zhu, Weikang and Zhang, Hongbin and Li, Li and Mao, Xiaoyang},
  year = 2024,
  month = jul,
  journal = {The Visual Computer},
  volume = {40},
  number = {7},
  pages = {4751--4763},
  issn = {1432-2315},
  doi = {10.1007/s00371-024-03468-2},
  urldate = {2025-06-05},
  langid = {english}
}

@misc{Zakharov2024,
  title = {Human {{Hair Reconstruction}} with {{Strand-Aligned 3D Gaussians}}},
  author = {Zakharov, Egor and Sklyarova, Vanessa and Black, Michael and Nam, Giljoo and Thies, Justus and Hilliges, Otmar},
  year = 2024,
  month = sep,
  number = {arXiv:2409.14778},
  eprint = {2409.14778},
  primaryclass = {cs},
  publisher = {arXiv},
  doi = {10.48550/arXiv.2409.14778},
  urldate = {2025-06-05},
  archiveprefix = {arXiv}
}

@inproceedings{zhangEnergyHairSketchBasedInteractive2022,
  title = {{{EnergyHair}}: {{Sketch-Based Interactive Guide Hair Design Using Physics-Inspired Energy}}},
  shorttitle = {{{EnergyHair}}},
  booktitle = {Graphics {{Interface}} 2022},
  author = {Zhang, Yuanwei and Kinuwaki, Shinichi and Umetani, Nobuyuki},
  year = 2022,
  month = mar,
  urldate = {2025-06-05},
  langid = {english}
}

@article{Zhou2023,
  title = {{{GroomGen}}: {{A High-Quality Generative Hair Model Using Hierarchical Latent Representations}}},
  author = {Zhou, Yuxiao and Chai, Menglei and Pepe, Alessandro and Gross, Markus and Beeler, Thabo},
  year = 2023,
  month = dec,
  journal = {ACM Transactions on Graphics},
  volume = {42},
  number = {6},
  pages = {1--16},
  publisher = {Association for Computing Machinery (ACM)},
  issn = {1557-7368},
  doi = {10.1145/3618309}
}

\newpage
\appendix

\section{Render Hairs Implementation}\label{sec:render_hairs_houdini}  
\begin{figure}[htbp]
    \centering
    \includegraphics[width=0.14\textwidth]{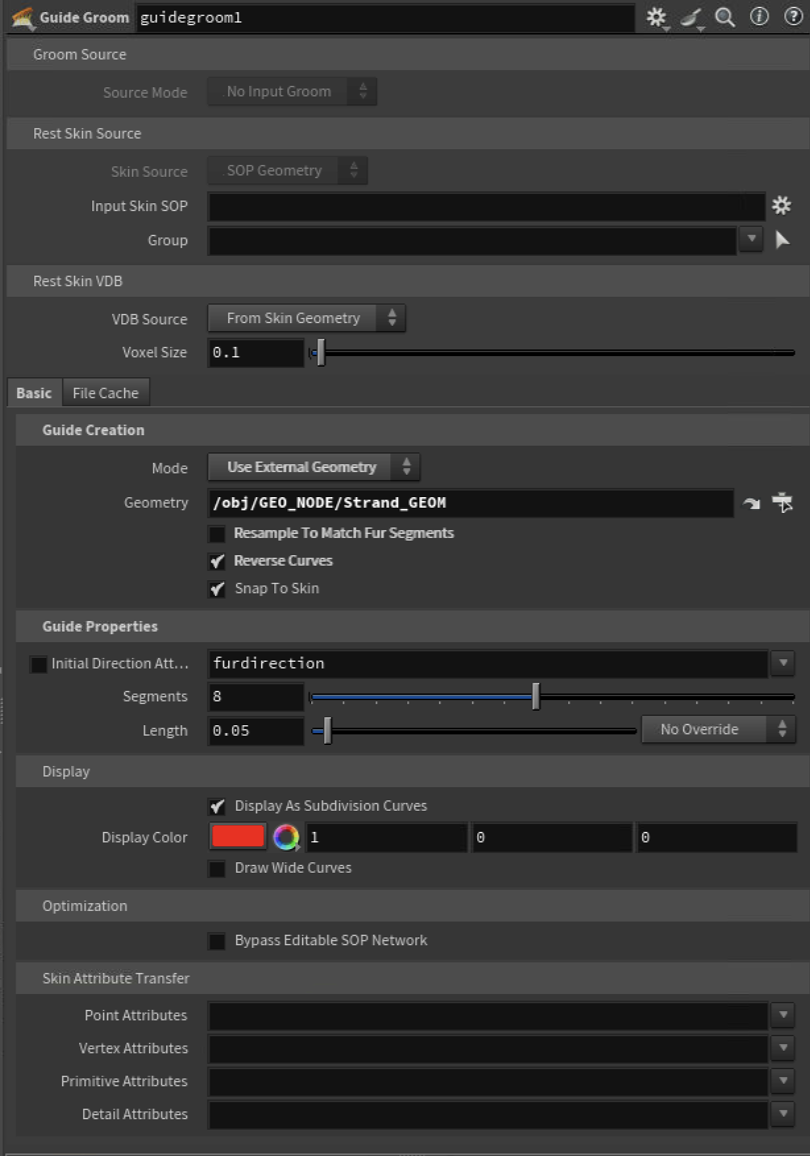}
    \hspace{.2cm} 
    \includegraphics[width=0.14\textwidth]{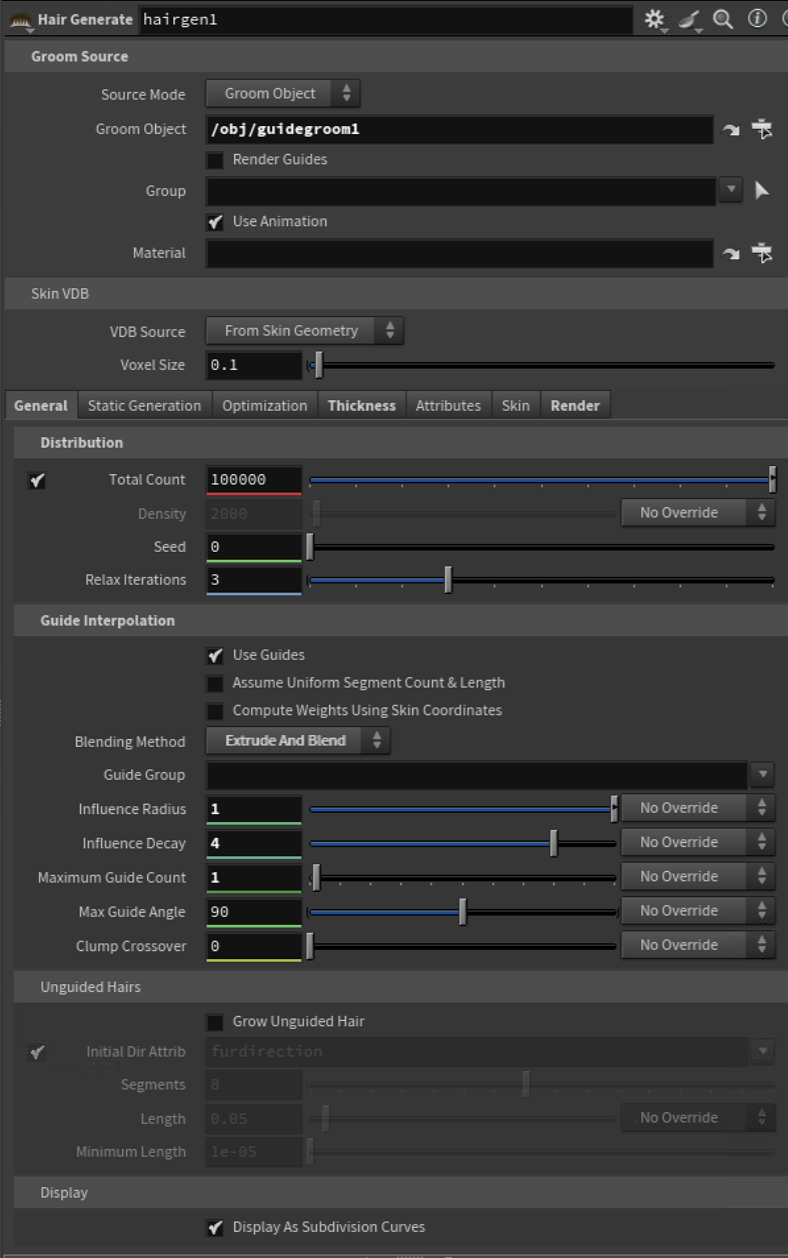}
    \hspace{0.01cm} 
    \includegraphics[width=0.14\textwidth]{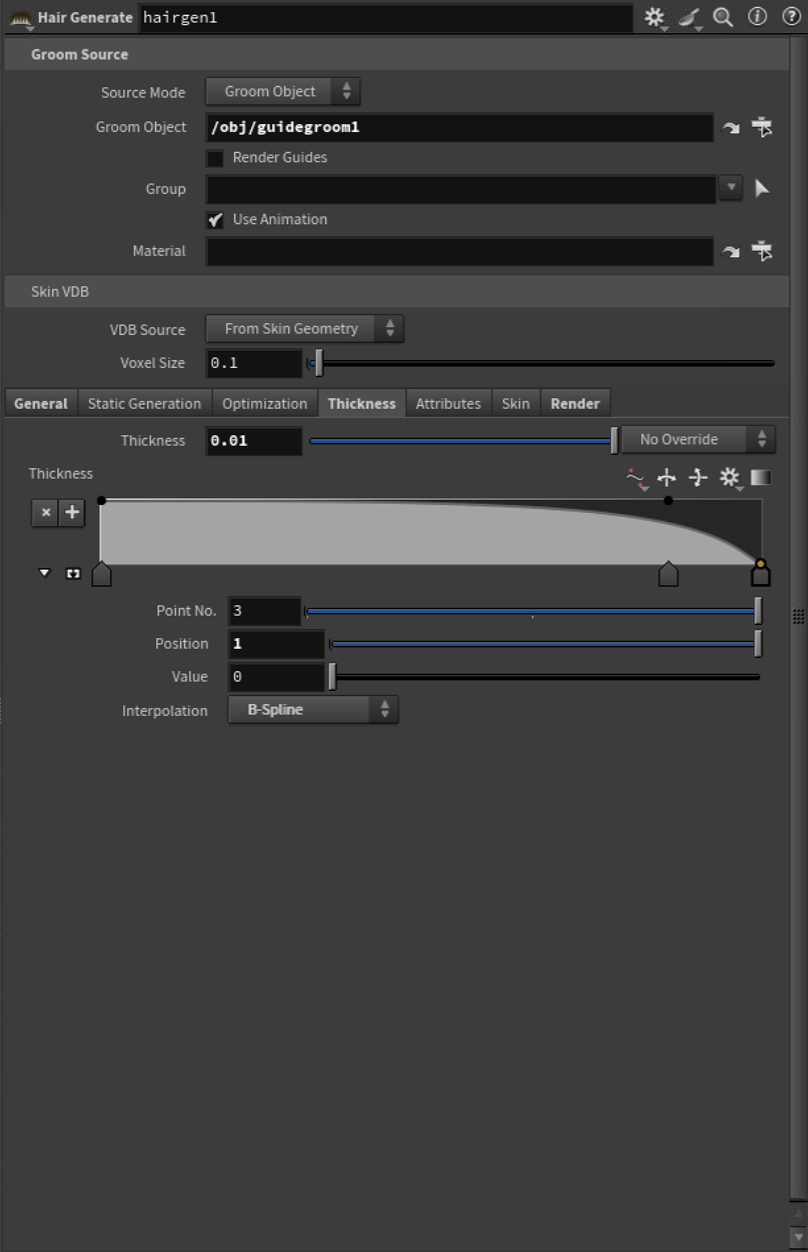}
    \caption{\footnotesize The parameter entries in the \textnormal{Guide Groom} node (left) and the \textnormal{Hair Generate} node (center and right) represent the entirety of settings we change from the defaults when generating render hairs. Specifically, in the \textnormal{Guide Groom} node under ``Guide Creation'', we change the \textit{Mode} and \textit{Geometry}, select \textit{Snap to Skin}, and deselect \textit{Resample To Match Fur Segments} . In the \textnormal{Hair Generate} node, in the ``General'' tab (center), we increase the \textit{Total Count} to 100k and set the \textit{Influence Radius} to 1. In the absence of clump nodes (Figure~\ref{fig:clump_variation}), \textit{Clump Crossover} should be set to 0. In the ``Thickness'' tab of the \textnormal{Hair Generate} node, we set \textit{Thickness} to the desired amount.}
    \Description{Screen captures of the relevant Houdini nodes to reproduce our configuration.}
    \label{fig:houdini_nodes}
\end{figure}

\begin{figure}[htbp]
    \centering
    \includegraphics[width=0.14\textwidth]{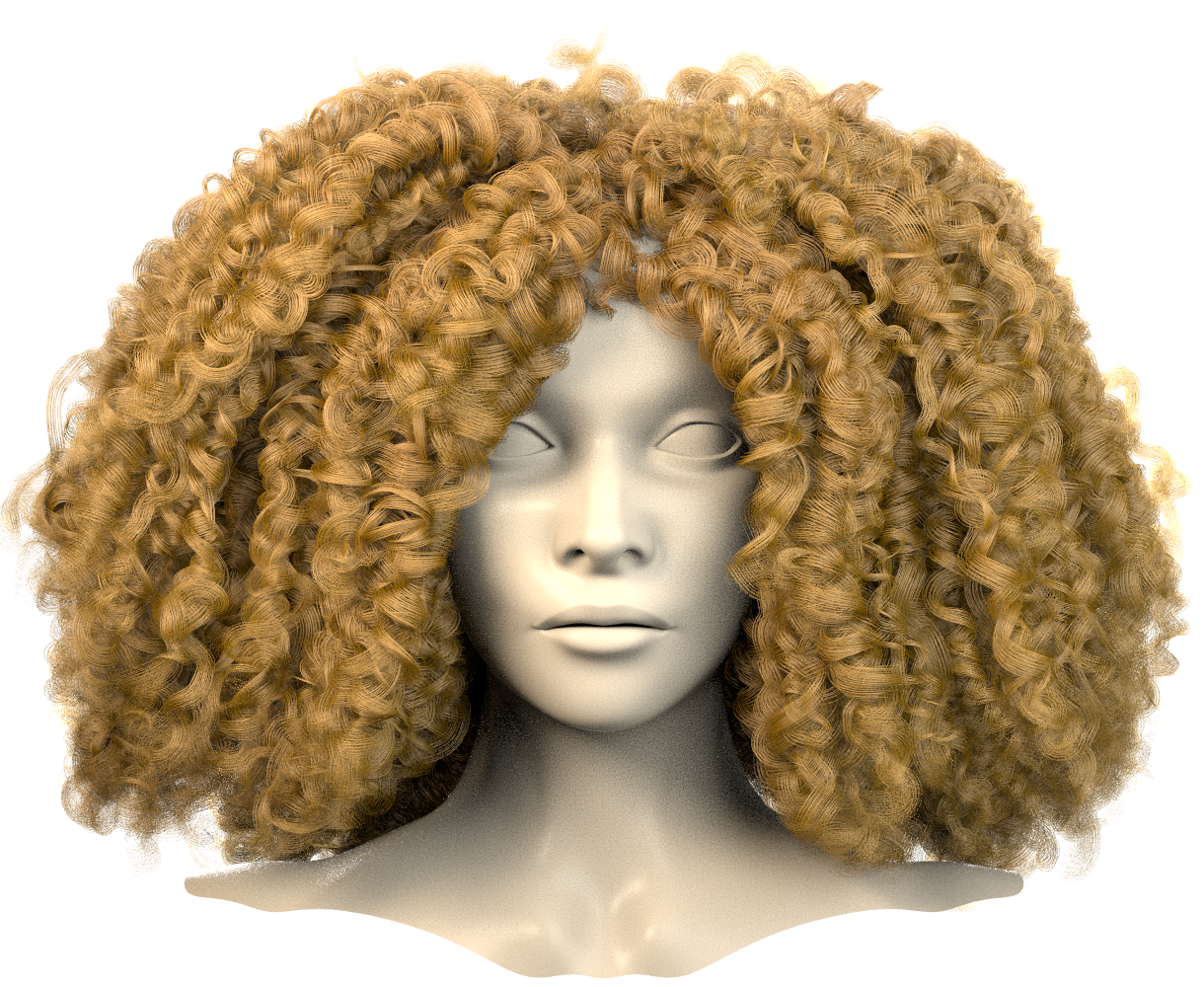}
    \hspace{.1cm} 
    \includegraphics[width=0.14\textwidth]{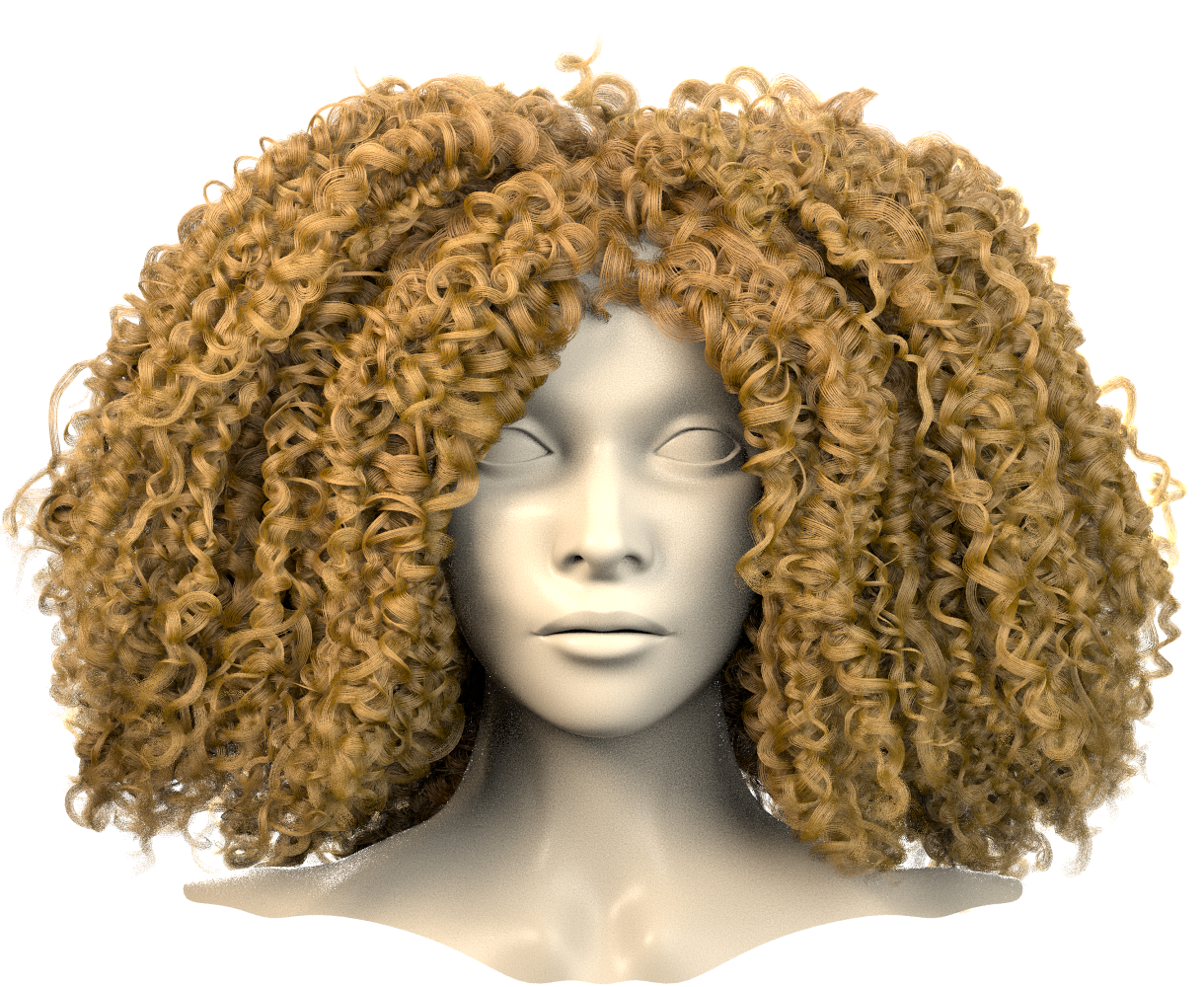}
    \hspace{0.1cm} 
    \includegraphics[width=0.14\textwidth]{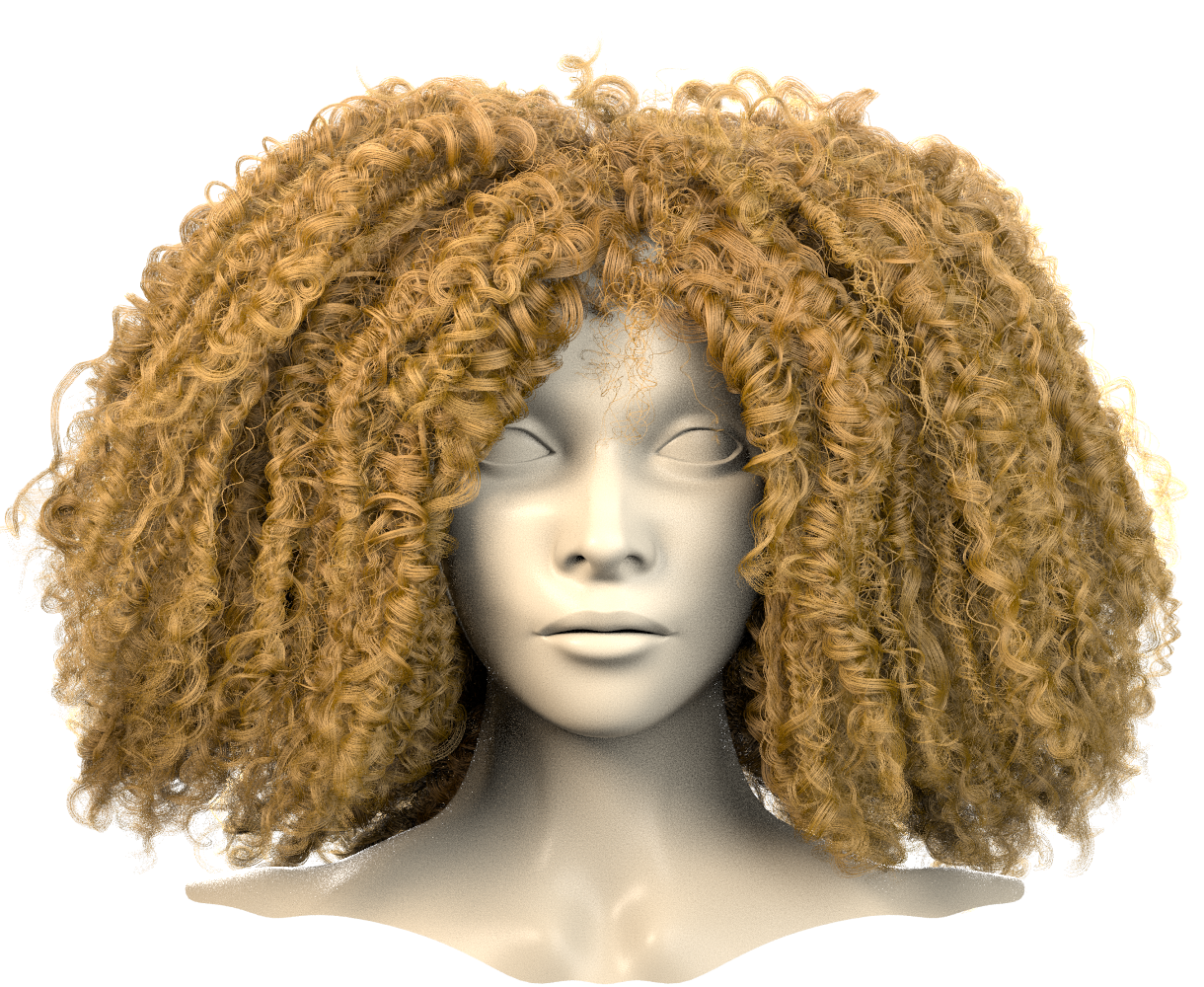}
    \caption{\footnotesize With the simple setup outlined in Figure~\ref{fig:houdini_nodes}, hair grooms might lack definition (left). To improve definition, we add one \textnormal{Hair Clump} node within the \textnormal{Guide Groom} node (center). In this node, we adjust \textit{Blend}, \textit{Clump Size}, \textit{Tightness}, and \textit{Clump Profile} to the given hairstyle. For further realism, we add another \textnormal{Hair Clump} node within the \textnormal{Hair Generate} node (right). In this node, we only manipulate the four values in the ``Fractal Clumping'' tab.}
    \Description{Screen captures of the relevant Houdini nodes to reproduce our configuration.}\label{fig:clump_variation}
\end{figure}

In order to better evaluate our results, we generate render hairs using an extremely simple node configuration in Houdini's free software, Houdini Apprentice \cite{houdini}. We begin by importing the strand geometry and head geometry via two \textbf{Geometry} nodes. We then add one \textbf{Guide Groom} node and one \textbf{Hair Generate} node. We connect the head \textbf{Geometry} node output to the second input of the \textbf{Guide Groom} node, also ensuring that the guide creation mode in the \textbf{Guide Groom} node is \textit{Use External Geometry} with the strand \textbf{Geometry} node selected. In the \textbf{Hair Generate} node, we select the \textbf{Guide Groom} node as the \textit{Groom Object} and adjust the parameters as desired. See Figure \ref{fig:houdini_nodes} for the specifications of our parameters. Optionally, for additional realism, we add \textbf{Hair Clump} nodes in both the \textbf{Guide Groom} and \textbf{Hair Generate} nodes. Clump in the \textbf{Guide Groom} node increases curl resolution from root to tip. Clump in the \textbf{Hair Generate} node introduces recursive clumping, which increases variation along the strand. See Figure~\ref{fig:clump_variation}. 

\end{document}